\documentclass[10pt,twocolumn,letterpaper]{article}
\PassOptionsToPackage{table,dvipsnames}{xcolor}
\usepackage[pagenumbers]{cvpr} %

\usepackage[utf8]{inputenc}
\usepackage[T1]{fontenc}
\usepackage{amsmath,amssymb,amsfonts}
\usepackage{nicefrac}
\usepackage{microtype}
\usepackage{textcomp}
\usepackage{booktabs}
\usepackage{graphicx}
\usepackage{pifont}
\usepackage{fontawesome5}
\usepackage{soul}
\usepackage[normalem]{ulem}
\usepackage{wrapfig}
\usepackage{multirow}
\usepackage{tabularx}
\usepackage{stackengine}
\usepackage{float}
\usepackage{array}
\usepackage[outline]{contour}
\usepackage{makecell}
\usepackage{pgfplots}
\usepackage{tabularray}
\usepackage{fancyvrb}
\usepackage{placeins}
\usetikzlibrary{calc, shapes, patterns, shadings, positioning, arrows.meta,3d, intersections}
\usepackage{styles}
\usepgfplotslibrary{fillbetween,colormaps}
\pgfplotsset{compat=1.17}
\usepackage{paralist}
\usetikzlibrary{tikzmark,patterns,shapes.misc,shapes.geometric,positioning,calc}

\definecolor{cvprblue}{rgb}{0.21,0.49,0.74}
\usepackage[pagebackref,breaklinks,colorlinks,allcolors=cvprblue]{hyperref}
\hypersetup{pdftitle={Less Is More: Semantic Curation of Earth Observation Data for Self-Supervised Learning},
  pdfauthor={Guillaume Astruc, Nicolas Gonthier, Cl\'ement Mallet, Loic Landrieu}}

\def\confName{CVPR}
\def\confYear{2027}

\AtEndPreamble{%
  \crefname{table}{Table}{Tables}
  \crefname{figure}{Figure}{Figures}
  \Crefname{figure}{Figure}{Figures}
  \crefname{appendix}{App.}{Apps.}
  \Crefname{appendix}{Appendix}{Appendices}
  \crefname{subappendix}{App.}{Apps.}
  \Crefname{subappendix}{Appendix}{Appendices}
}

\usepackage{xspace}
\newcommand{\ours}{\textup{\textsc{Geo\-Select}}\xspace} %

\newcommand{\papertitle}{%
Less Is More: Semantic Curation of Earth Observation Data\\ for Self-Supervised Learning%
}

\title{\papertitle}

\author{
Guillaume Astruc\textsuperscript{1,2,4} \quad
Nicolas Gonthier\textsuperscript{2,3} \quad
Clément Mallet\textsuperscript{2} \quad
Loic Landrieu\textsuperscript{1,2}\\[0.2cm]
{\normalsize \textsuperscript{1}LIGM, Ecole Nationale des Ponts et Chaussées, IP Paris, Univ Gustave Eiffel, CNRS}\\
{\normalsize \textsuperscript{2}LASTIG, Univ Gustave Eiffel, IGN, ENSG \quad \textsuperscript{3}IGN \quad \textsuperscript{4}CNES}
}

\begin{document}
\setlength{\leftmargini}{2em}

\maketitle

\begin{abstract}
Earth-observation (EO) foundation models are pretrained on large satellite archives with substantial spatial redundancy.
While these archives are often sampled using hand-designed geographic and metadata-based rules, they are rarely curated semantically, despite the success of this approach in computer vision.
We investigate the impact of data curation on self-supervised EO pretraining and introduce \ours{}, a simple, training-free semantic curation method tailored to EO.
{In a single pass over the features of a frozen EO foundation model}, \ours{} selects a fixed-size subset that promotes both global diversity, across observations, and local diversity, within each observation, measured from its patch features.
To evaluate curation independently of the pretrained representation used for selection, we establish a common protocol in which a new model is pretrained from scratch on each selected subset and evaluated by linear probing across an array of classification and segmentation benchmarks.
On the global Sentinel-2 Major-TOM archive, pretraining on only $1\%$ of the data improves mean performance by $+2.4$ points in classification and $+2.2$ mIoU in segmentation,  outperforming geographic, metadata-based, and standard semantic curation baselines.
These gains persist across curation budgets and SSL objectives, and extend across sensing modalities.
\ours{} also generalizes to the multi-temporal archive S2-NAIP, where it also retains its performance with only $1\%$ of the data.
Our results show that semantic curation can improve EO representation quality while reducing the pretraining archive by two orders of magnitude.
The curation pipeline and curated subsets are available at \href{https://github.com/gastruc/GeoSelect}{\nolinkurl{github.com/gastruc/GeoSelect}}.
\end{abstract}

\setcounter{topnumber}{1}\setcounter{dbltopnumber}{1}
\section{Introduction}
\label{sec:intro}
\begin{figure}[t]
    \centering
    \input{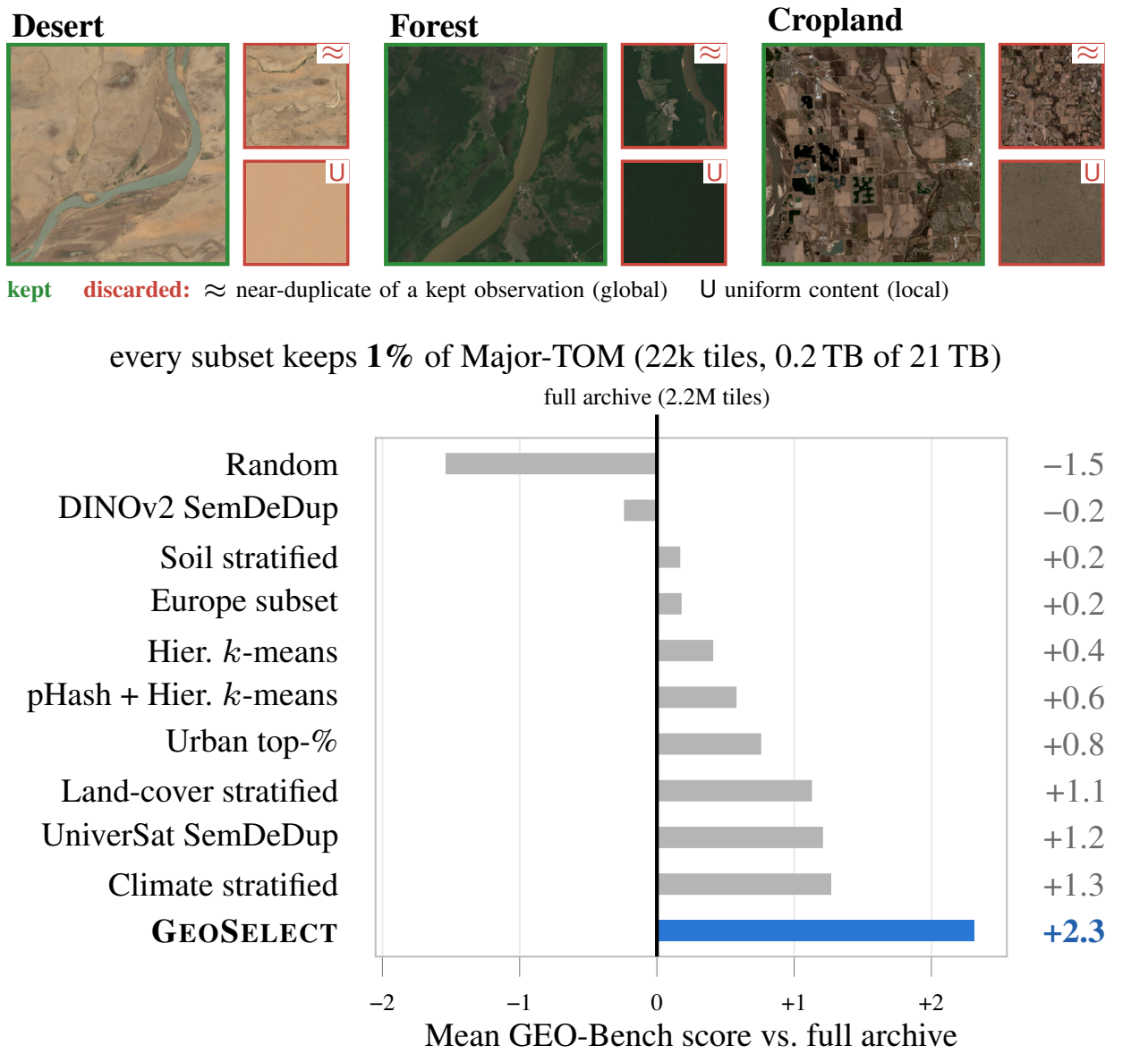}
    \vspace{-2mm}
\caption{\textbf{Better representations with $100\times$ less data.}
Top: \ours{} promotes global diversity by avoiding redundant observations while favoring observations with richer local content over internally uniform ones (\textsf{U}). Bottom: at the same $1\%$ budget, \ours{} outperforms both full-archive pretraining and alternative curation strategies, improving mean GEO-Bench linear-probe performance by $+2.3$ points.}
    \label{fig:teaser}
\end{figure}

The quality of self-supervised representations depends not only on model scale, compute, and dataset size \citep{chinchilla}, but also on the composition of the pretraining data \citep{caron}. This has motivated large-scale filtering and data selection in language modeling \citep{brown2020language,xie2023data}, and feature-based deduplication and diversification in vision \citep{abbas2023semdedup,oquab2024dinov2,simeoni2025dinov3}. Data composition is particularly important in Earth observation (EO)~\citep{roscher2024better}, where spatial autocorrelation and vast homogeneous regions create substantial redundancy. Yet EO pretraining datasets are still primarily constructed using hand-designed geographic, climatic, or land-cover sampling rules \citep{manas2021seasonal,schmitt2019sen12ms,wang2023ssl4eo,blumenstiel2025terramesh,zhu2026foundations}. Recent feature-based EO curation \citep{barseghyan2025less,kerdreux2025efficient,wei2025rsprune} still describes each observation by a single embedding. EO observations are also often internally homogeneous: their spatial resolution can be coarse relative to fine-scale texture, so large forests, deserts, or agricultural regions may appear nearly uniform. Such observations provide a weaker SSL signal: different crops contain little content variation for view-based objectives, while masked regions are easier to reconstruct for MAE-like objectives~\citep{he2022mae}. Effective curation should therefore promote diversity at two complementary levels: \emph{global diversity} across the selected observations, and \emph{local diversity} within each observation.

To address this, we introduce \ours{}, a simple feature-based curation method that explicitly promotes both global and local diversity~(\cref{fig:teaser}). It partitions the archive into semantic clusters and favors observations that are distinctive from others in their cluster, promoting global diversity. It further favors observations with heterogeneous patch-level features over internally uniform observations, promoting local diversity. Together, these criteria select observations that are both complementary to the rest of the subset and rich in local content. It needs no training, labels, or reference data, and runs once on frozen features.

\begin{figure*}[t]
    \centering
    \input{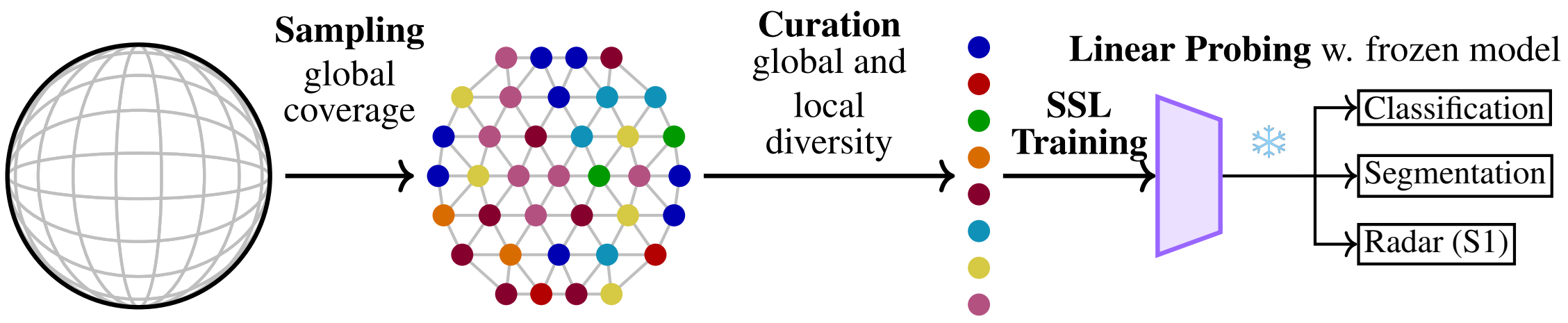}
    \caption{\textbf{Evaluation pipeline.}
    We curate a global EO archive using frozen UniverSat embeddings~\citep{perron2025universat}, pretrain an SSL model from scratch on the selected subset, and evaluate its frozen representation by linear probing on downstream tasks.}
    \label{fig:pipeline}
\end{figure*}

We evaluate \ours{} on Major-TOM~\citep{francis2024majortom}, a large, minimally curated Sentinel-2 archive covering the Earth's landmasses and coastlines. We establish a controlled protocol comparing geographic and metadata-based heuristics with feature-based methods such as SemDeDup~\citep{abbas2023semdedup} and hierarchical $k$-means~\citep{vo2024automatic}. Each method selects a fixed-size subset from the same archive; we then pretrain a ViT-B from scratch under matched optimization budgets and evaluate its frozen representations by linear probing on GEO-Bench (\cref{fig:pipeline}). We repeat this protocol across several SSL objectives to test whether the effect of curation depends on the pretraining strategy.

Our experiments reveal substantial redundancy in large EO archives: subsets containing only $1\%$ of Major-TOM can outperform full-archive pretraining across downstream tasks. Geographic and metadata-based heuristics already provide strong baselines, but local diversity, which single-embedding methods cannot capture, accounts for most of the gain of \ours{} over SemDeDup. These benefits persist across curation budgets, SSL objectives, and sensing modalities.

\vspace{1mm}
\noindent{Our main contributions are as follows:}
\begin{compactenum}
    \item We systematically compare geographic, metadata-based, and feature-based curation strategies for self-supervised EO pretraining.
    \item We introduce \ours{}, a {simple, training-free} semantic curation method that combines diversity across observations with diversity within each observation{; to our knowledge, it is the first EO curation method to measure the latter from patch features}.
    \item We show that curated subsets using only $1\%$ of a large EO archive can outperform full-data pretraining, with gains extending across SSL objectives, data budgets, and sensing modalities.
\end{compactenum}

\section{Related work}
\label{sec:related}

\paragraph{Data curation for self-supervised pretraining.}
Curating large pretraining corpora can preserve or improve downstream performance while substantially reducing the amount of data used for training \citep{sorscher2022beyond,abbas2023semdedup,gadre2023datacomp}. In language modeling, curation ranges from manually assembled mixtures of high-quality sources
\citep{gao2020pile,laurenccon2022bigscience}
to automatic quality filtering and deduplication
\citep{brown2020language,penedo2023refinedweb},
and distribution-aware data selection
\citep{xie2023data}.
In vision, SemDeDup removes semantic near-duplicates in embedding space
\citep{abbas2023semdedup}, while other approaches cluster pretrained representations to promote diversity
\citep{sorscher2022beyond,tirumala2023d4,vo2024automatic}.
{Density-based pruning~\citep{abbas2024dbp} allocates more of the budget to clusters that are spread out and far from the others, and D$^2$-Pruning~\citep{maharana2024d2} balances diversity with how hard each example is for a supervised model to learn.}
At web scale, DINOv2 combines embedding-based deduplication with retrieval from manually curated reference datasets
\citep{oquab2024dinov2}, whereas \citet{vo2024automatic} replace these references with hierarchical $k$-means and balanced sampling. {All these methods describe each image by a single embedding, and thus promote diversity across images but not within them.}

\paragraph{Data curation for Earth observation.}
Recent studies further show that pretraining performance depends strongly on which EO observations are retained, not only on geographic coverage
\citep{purohit2025spatial,plekhanova2025ssl4eco,kaur2026pretrainwhere}.
{Pretraining set size also matters~\citep{burgert2026georank}, and EO models themselves are redundant~\citep{hackel2026slimmability}.}
EO pretraining datasets are typically curated using hand-designed geographic or metadata-based rules, including sampling around populated areas \citep{manas2021seasonal,wang2023ssl4eo,stewart2023ssl4eol,astruc2026unigeoclip}, geographically stratified sampling \citep{schmitt2019sen12ms}, balancing across land-cover classes or eco-regions \citep{blumenstiel2025terramesh,jakubik2025terramind,szwarcman2024prithvi,gaydon2024fractal}, and aggregating existing task-specific datasets into heterogeneous collections \citep{mendieta2023gfm,velazquez2025earthview,astruc2024anysat}{, possibly spanning several sensors~\citep{xiong2024dofa}}. Major-TOM \citep{francis2024majortom} instead provides a deliberately uncurated, near-contiguous archive {of Sentinel-2 and Sentinel-1 images} covering most of the Earth's landmasses, together with a smaller official subset. Beyond these construction heuristics, \citet{jiang2024less} explore active-learning-inspired subset selection, while \citet{perera2026clustering} cluster hand-crafted spectral, texture, and acquisition descriptors.

More recently, feature-based curation has also received increasing attention \citep{redundancyaware2025,wei2025rsprune}. {RS-Prune~\citep{wei2025rsprune} discards images of low pixel entropy before clustering.} Hierarchical clustering of learned representations has been explored by \citet{vo2024automatic} and combined with semantic deduplication by \citet{barseghyan2025less}, while \citet{kerdreux2025efficient} dynamically recluster and prune the training archive during SSL pretraining. In the vision-language setting, \citet{muhtar2026quality} instead {train a reward model on preference data to select remote-sensing image-text pairs}. In contrast, \ours{} performs a single offline curation pass using frozen EO embeddings, {scoring each observation both by its redundancy with the rest of the archive and by the diversity of its own patch tokens.}

\section{Method}
\label{sec:method}
The goal of \ours{} is to select a fixed-size subset from a large, unfiltered EO archive that is diverse at two complementary levels. Globally, across observations, the subset should cover as much of the archive's semantic variability as possible. Locally, each retained observation should contain rich and varied content, providing a stronger source of supervision for self-supervised learning. The selected subset is then used for pretraining.

We first represent each observation with a pretrained EO foundation model and use its global and patch-level features to measure global and local diversity, respectively (\cref{sec:method:diversity}). We then combine these criteria under a fixed curation budget (\cref{sec:method:prune}).

\subsection{Measuring local and global diversity}
\label{sec:method:diversity}

Let $\cX=\{x_i\}_{i=1}^{N}$ denote an unfiltered archive of $N$ EO observations. We embed each observation with a pretrained EO foundation model $\phi^\text{FM}$, which maps $x_i$ to $T$ patch tokens $t_{i,1},\dots,t_{i,T}\in\mathbb{R}^D$. Their average defines the global embedding
\begin{align}
    f_i
    &=
    \frac{1}{T}
    \sum_{k=1}^{T}
    t_{i,k}.
    \label{eq:embed}
\end{align}
We $\ell_2$-normalize $f_i$, so that all subsequent comparisons use cosine similarity. The encoder $\phi^\text{FM}$ is used only for curation: \ours{} outputs observation identifiers, and the selected subset can subsequently be used to pretrain any architecture or objective, including with a different sensing modality.

Local and global diversity capture complementary properties of an observation. A locally uniform observation may contain globally distinctive content, while a locally diverse observation may be redundant with many similar observations. We therefore measure the two separately.

\begin{table*}[t]
\centering
\caption{\textbf{Downstream performance on GEO-Bench.}
The first row reports performance after pretraining on the full archive; all other values are differences relative to this baseline. Classification uses accuracy except for BigEarthNet ($\mu$F1), while segmentation uses mIoU. SDD denotes budget-controlled SemDeDup. Means are computed over unrounded scores.}
\label{tab:geobench_curation}
    \scriptsize
\setlength{\tabcolsep}{2pt}
\renewcommand{\arraystretch}{1.03}
\renewcommand{\applycolorwidth}{1.9em}

\resizebox{\textwidth}{!}{%
\begin{tabular}{
    l l@{\;}c@{\;}
    @{\quad}
    c c c c c c
    @{\;\;\;\;}
    c c c c c c
}
\toprule
\multicolumn{3}{c}{Curation}
&
\multicolumn{6}{c}{Classification}
&
\multicolumn{6}{c}{Segmentation} \\
\cmidrule(lr){1-3}
\cmidrule(lr){4-9}
\cmidrule(lr){10-15}

\makecell{Criterion/\\features} & \makecell{Sampling \\ strategy} & Kept
& \makecell{So2\\Sat} & \makecell{Big\\Earth\\Net} & \makecell{Forest\\Net} & \makecell{Euro\\SAT} & \makecell{PV4\\GER} & \bf Mean
& \makecell{Cash\\ew} & \makecell{Chesa\\peake} & \makecell{SA\\Crop} & \makecell{Neon\\Tree} & \makecell{PV4\\GER} & \bf Mean \\
\midrule

\multicolumn{2}{l}{\emph{Full dataset}} & 100\%
& 42.0 & 53.4 & %
47.9 & 78.1 & 92.9 & 62.9
& 36.1 & 57.4 & 27.1 & 53.5 & 76.9 & 50.2 \\

\midrule

\multicolumn{2}{l}{\emph{Official subset}} & 11\%
& \applycolordeltaA{+0.3}
& \applycolordeltaB{-1.5}
& \applycolordeltaD{-3.3}
& \applycolordeltaE{-1.6}
& \applycolordeltaF{+1.2}
& \applycolordeltaG{-1.0}
& \applycolordeltaH{-1.6}
& \applycolordeltaI{+0.7}
& \applycolordeltaJ{-0.6}
& \applycolordeltaK{-0.2}
& \applycolordeltaL{-0.8}
& \applycolordeltaM{-0.5} \\

\midrule

Uniform & random & 1\%
& \applycolordeltaA{+0.5}
& \applycolordeltaB{-3.5}
& \applycolordeltaD{-1.9}
& \applycolordeltaE{+0.5}
& \applycolordeltaF{0.0}
& \applycolordeltaG{-0.9}
& \applycolordeltaH{-2.4}
& \applycolordeltaI{-2.6}
& \applycolordeltaJ{-1.1}
& \applycolordeltaK{-0.9}
& \applycolordeltaL{-4.0}
& \applycolordeltaM{-2.2} \\

Europe & subset & 1\%
& \applycolordeltaA{-1.2}
& \applycolordeltaB{-0.1}
& \applycolordeltaD{-1.1}
& \applycolordeltaE{+1.0}
& \applycolordeltaF{-0.6}
& \applycolordeltaG{-0.4}
& \applycolordeltaH{+1.0}
& \applycolordeltaI{+1.3}
& \applycolordeltaJ{-0.2}
& \applycolordeltaK{+0.1}
& \applycolordeltaL{+1.6}
& \applycolordeltaM{+0.8} \\

Urban & top-\% & 1\%
& \applycolordeltaA{-5.0}
& \applycolordeltaB{+0.2}
& \applycolordeltaD{-0.7}
& \applycolordeltaE{+3.1}
& \applycolordeltaF{-0.1}
& \applycolordeltaG{-0.5}
& \applycolordeltaHb{+5.6}
& \applycolordeltaI{+1.9}
& \applycolordeltaJ{-1.1}
& \applycolordeltaK{-0.1}
& \applycolordeltaLb{+3.9}
& \applycolordeltaM{+2.0} \\

Land cover & stratified & 1\%
& \applycolordeltaA{+1.0}
& \applycolordeltaB{+0.8}
& \applycolordeltaD{+1.1}
& \applycolordeltaE{+2.2}
& \applycolordeltaF{+0.4}
& \applycolordeltaG{+1.1}
& \applycolordeltaH{+1.4}
& \applycolordeltaI{+1.5}
& \applycolordeltaJ{-0.1}
& \applycolordeltaK{-0.1}
& \applycolordeltaL{+3.0}
& \applycolordeltaM{+1.1} \\

Climate & stratified & 1\%
& \applycolordeltaA{-0.4}
& \applycolordeltaB{+1.2}
& \applycolordeltaD{+1.3}
& \applycolordeltaE{+2.7}
& \applycolordeltaF{+0.2}
& \applycolordeltaG{+1.0}
& \applycolordeltaH{+2.5}
& \applycolordeltaI{+2.6}
& \applycolordeltaJ{-0.2}
& \applycolordeltaK{0.0}
& \applycolordeltaL{+2.9}
& \applycolordeltaM{+1.6} \\

Soil & stratified & 1\%
& \applycolordeltaA{-0.7}
& \applycolordeltaB{-0.5}
& \applycolordeltaD{+0.2}
& \applycolordeltaE{+0.5}
& \applycolordeltaF{-0.1}
& \applycolordeltaG{-0.2}
& \applycolordeltaH{+1.5}
& \applycolordeltaI{+0.7}
& \applycolordeltaJ{0.0}
& \applycolordeltaK{-0.2}
& \applycolordeltaL{+0.3}
& \applycolordeltaM{+0.5} \\\greyrule

UniverSat & Hier. $k$-means & 1\%
& \applycolordeltaA{0.0}
& \applycolordeltaB{+0.2}
& \applycolordeltaD{+1.0}
& \applycolordeltaE{+1.4}
& \applycolordeltaF{-0.3}
& \applycolordeltaG{+0.4}
& \applycolordeltaH{-0.2}
& \applycolordeltaI{+0.7}
& \applycolordeltaJ{+0.4}
& \applycolordeltaK{-0.1}
& \applycolordeltaL{+1.0}
& \applycolordeltaM{+0.4} \\

UniverSat & pHash + Hier. $k$-m. & 1\%
& \applycolordeltaA{+0.3}
& \applycolordeltaB{-0.4}
& \applycolordeltaD{+1.1}
& \applycolordeltaE{+2.8}
& \applycolordeltaF{-0.3}
& \applycolordeltaG{+0.7}
& \applycolordeltaH{+0.7}
& \applycolordeltaI{+0.6}
& \applycolordeltaJ{+0.1}
& \applycolordeltaK{+0.1}
& \applycolordeltaL{+0.8}
& \applycolordeltaM{+0.4} \\

DINOv2 & SDD & 1\%
& \applycolordeltaAb{+1.7}
& \applycolordeltaB{-1.1}
& \applycolordeltaD{-1.2}
& \applycolordeltaE{+2.6}
& \applycolordeltaF{-0.5}
& \applycolordeltaG{+0.3}
& \applycolordeltaH{-1.9}
& \applycolordeltaI{-0.7}
& \applycolordeltaJ{+0.1}
& \applycolordeltaK{-0.4}
& \applycolordeltaL{-1.0}
& \applycolordeltaM{-0.8} \\

UniverSat & SDD & 1\%
& \applycolordeltaA{-0.8}
& \applycolordeltaB{+1.1}
& \applycolordeltaD{+0.1}
& \applycolordeltaE{+4.6}
& \applycolordeltaF{-0.1}
& \applycolordeltaG{+0.9}
& \applycolordeltaH{+2.5}
& \applycolordeltaI{+1.8}
& \applycolordeltaJb{+0.5}
& \applycolordeltaK{0.0}
& \applycolordeltaL{+2.4}
& \applycolordeltaM{+1.5} \\\greyrule

UniverSat & \bf \ours{} & 1\%
& \applycolordeltaA{+0.3}
& \applycolordeltaBb{+2.7}
& \applycolordeltaDb{+2.8}
& \applycolordeltaEb{+5.4}
& \applycolordeltaFb{+1.0}
& \applycolordeltaGb{+2.4}
& \applycolordeltaH{+4.2}
& \applycolordeltaIb{+2.9}
& \applycolordeltaJ{+0.2}
& \applycolordeltaKb{+0.2}
& \applycolordeltaL{+3.6}
& \applycolordeltaMb{+2.2} \\

\bottomrule
\end{tabular}
}

\end{table*}

\paragraph{Local diversity.}
We measure local diversity from the dispersion of the patch tokens, using a diagonal approximation of the coding rate from MCR$^2$~\citep{yu2020mcr2}. Let $t_{i,k,d}$ denote feature $d$ of token $k$ in observation $x_i$, and let $\bar{t}_{i,d}$ denote its mean across the $T$ tokens. We compute
\begin{align}
    \sigma_{i,d}^2
    &=
    \frac{1}{T}
    \sum_{k=1}^{T}
    \left(
        t_{i,k,d}-\bar{t}_{i,d}
    \right)^2,
    \qquad
    \epsilon_d
    =
    \operatorname*{median}_{1\le j\le N}\sigma_{j,d},
    \label{eq:sigma}
    \\
    L_i
    &=
    \frac{1}{2}
    \sum_{d=1}^{D}
    \log\left(
        1+\frac{\sigma_{i,d}^2}{\epsilon_d^2}
    \right).
    \label{eq:local}
\end{align}
Here, $\epsilon_d$ normalizes each feature by its typical variation across the archive. The score $L_i$ is large when the patch tokens vary strongly along several feature dimensions.

\paragraph{Global diversity.}
Global diversity measures how much new semantic content an observation contributes relative to other observations in the archive. For scalability, we partition the global embeddings $f_i$ into $K$ semantic clusters using $k$-means and restrict comparisons to observations within the same cluster.
We build on the principle of SemDeDup~\citep{abbas2023semdedup}, which measures the distinctiveness of each observation relative to those preceding it under a chosen ordering. This ordering determines which observation is favored when several are near-duplicates. While SemDeDup orders observations by their distance to the cluster centroid, we instead order them by decreasing local diversity $L_i$: if two observations contain essentially the same global content, we prefer the one with richer local content.

Let $\mathcal{P}_i$ denote the more locally diverse observations that precede $i$ within its cluster $c$,
\begin{align}
    \mathcal{P}_i
    &=
    \left\{
        j :
        c(j)=c(i),\;
        L_j>L_i
    \right\}.
    \label{eq:predecessors}
\end{align}
We define the global diversity of observation $i$ as its cosine distance to its closest predecessor,
\begin{align}
    G_i
    &=
    1-
    \max_{j\in\mathcal{P}_i}
    \cos\left(
        f_i,f_j
    \right),
    \label{eq:global}
\end{align}
with $G_i=2$ when $\mathcal{P}_i=\varnothing$. Thus, $G_i$ is large when the global content of observation $i$ is not already represented by a locally richer observation. When two observations are near-duplicates, the less locally diverse one consequently receives a small $G_i$.

\subsection{Budgeted diversity-aware selection}
\label{sec:method:prune}

Given a budget of $B$ observations, we must determine how much of the budget to allocate to each semantic cluster and which observations to retain within each cluster.

\paragraph{Combining local and global diversity.}
We combine the two diversity measures into a single score,
\begin{align}
    S_i
    &=
    \frac{\log G_i}{\operatorname{IQR}\left(\log G\right)}
    +
    \frac{L_i}{\operatorname{IQR}\left(L\right)}.
    \label{eq:diversity}
\end{align}
Here, $\operatorname{IQR}$ denotes the interquartile range over the archive, which puts the two terms on comparable scales. A high $S_i$ therefore identifies observations that are both globally distinctive and locally diverse.

\paragraph{Allocating the budget across clusters.}
Clusters dominated by near-duplicates should receive less of the budget than clusters containing many distinct observations. We therefore use the SemDeDup redundancy estimate to determine the budget of each cluster, independently of the observations ultimately selected.
Within each cluster, observations are ordered by decreasing distance to the cluster centroid. For each observation $i$, let $\mathcal{Q}_i$ denote the observations preceding it in this ordering, and define the redundancy $m_i$ as:
\begin{align}
    m_i
    &=
    \max_{j\in\mathcal{Q}_i}
    \cos\left(
        f_i,f_j
    \right).
    \label{eq:cluster_redundancy}
\end{align}
A large $m_i$ indicates that $i$ is already well represented by another observation in its cluster. This ordering is used only to estimate redundancy and allocate the budget.
We choose a single threshold $\tau_B$ such that exactly $B$ observations satisfy $m_i\leq\tau_B$ across the archive. The number of such observations in cluster $c$ defines its budget,
\begin{align}
    B_c
    &=
    \left|
        \left\{
            i :
            c(i)=c,\;
            m_i\leq\tau_B
        \right\}
    \right|,
    \qquad
    \sum_c B_c=B.
    \label{eq:cluster_budget}
\end{align}
Thus, $B_c$ estimates how many non-redundant observations cluster $c$ can contribute: redundant clusters receive a smaller budget, while more heterogeneous clusters receive a larger one.

\begin{figure*}[t]
\centering
\input{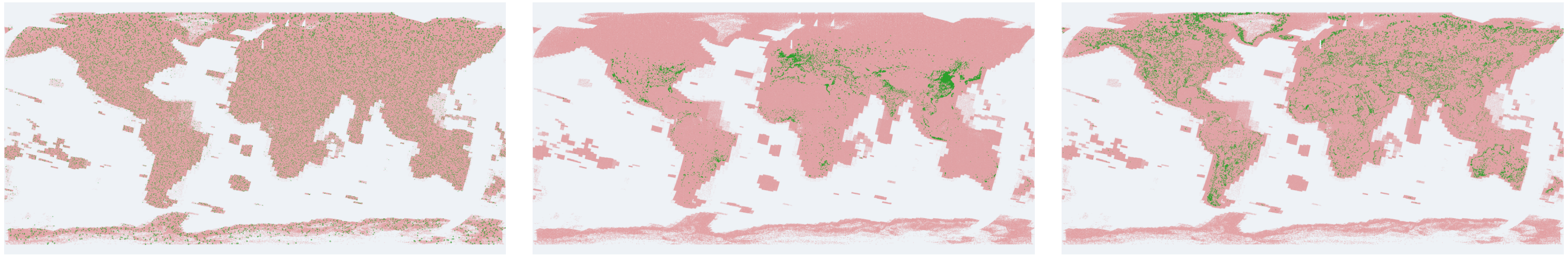}
\caption{\textbf{Spatial distribution of selected observations.}
Major-TOM observations retained at a $1\%$ budget by (a) random sampling,
(b) prioritization by built-up fraction, and (c) \ours{}.
\textcolor{green!55!black}{Green} observations are retained and
\textcolor{red!75!black}{red} observations are discarded.}
\label{fig:location_analysis}
\end{figure*}

\paragraph{Selection within each cluster.}
The redundancy scores $m_i$ determine only the cluster budgets. Once $B_c$ is fixed, \ours{} retains the $B_c$ observations in cluster $c$ with the largest joint diversity score $S_i$. The redundancy estimate therefore determines \emph{how many} observations are retained from each semantic cluster, while $S_i$ determines \emph{which} observations represent it.

\section{Experiments}
\label{sec:experiments}
\subsection{Evaluation protocol}
\label{sec:exp}

We evaluate curation methods through the quality of the representations learned from their selected subsets. We use Major-TOM Core-S2L2A, a near-contiguous and minimally curated global archive of $2{,}238{,}687$ single-date Sentinel-2 $10\times10$\,km observations covering the Earth's landmasses and coastlines. Its scale and limited prior filtering make it particularly well-suited for studying data curation; additional details are provided in \cref{sec:supp:majortom}. For each curation method, we select a fixed-size subset, pretrain a new model from scratch on the selected observations, and evaluate its frozen representations on downstream tasks (\cref{fig:pipeline}).

\paragraph{Curation parameters.}
Unless stated otherwise, each method retains $1\%$ of the archive. We curate using either the EO-specific UniverSat representation~\citep{perron2025universat}, computed from ten Sentinel-2 bands, or DINOv2 features extracted from the RGB bands only~\citep{czerkawski2024global}. We use these two because embeddings of the full archive are publicly available on Hugging Face for both. These representations are used only for curation: after selection, only the observation identifiers are retained, and no curation features are passed to pretraining.

\paragraph{Self-supervised pretraining.}
For the main experiments, we pretrain a ViT-B from scratch with MAE~\citep{he2022mae} on random $256\times256$ pixel crops from the selected observations. We use AdamW with identical hyperparameters across subsets (\cref{sec:supp:pretraining}) and match the optimization budget by defining each epoch as $2{,}000$ optimizer steps with a global batch of $256$, independently of subset size. Smaller subsets are therefore revisited more often. All MAE models are evaluated after $90$ epochs, corresponding to $180$k optimizer steps and approximately $46$M crops, with no early stopping or checkpoint selection, so that every model has exactly the same training budget. The plateau learning-rate schedule monitors MAE reconstruction loss on a fixed held-out $1\%$ validation split of Major-TOM, shared across all runs and never used for training.

\paragraph{Downstream evaluation.}
We evaluate the frozen representations on $5$ classification and $5$ semantic segmentation tasks from GEO-Bench \citep{lacoste2023geo}; five of them come from Landsat-8 or aerial imagery, whose bands we map onto the matching Sentinel-2 bands, setting the others to zero. We exclude m-nz-cattle, whose label maps would have to be resized, and m-brick-kiln, whose training was unstable for most methods. For classification, we train a linear classifier on the {mean- or max-pooled frozen patch tokens, the pooling being selected on the validation set,} to predict the $K$ classes. For segmentation, we apply a linear classifier to each frozen patch representation, predicting $P\times P\times K$ logits corresponding to $K$ class scores for each pixel in the patch. For each task, we sweep the probe learning rate and weight decay, select the best probe on the validation set, and report test performance. For classification, we also report a $k$-nearest-neighbor classifier on the frozen image embeddings, which doesn't requires training probes (details in \cref{sec:supp:knn}). Probing and pretraining hyperparameters are given in \cref{sec:supp:probing,sec:supp:pretraining}.

\subsection{Competing curation strategies}
\label{sec:exp:baselines}

We compare \ours{} against alternative strategies for selecting a budget of $B$ observations from the archive $\cX$. Unless stated otherwise, all methods are evaluated under the same pretraining and probing protocol of \cref{sec:exp}.

\paragraph{Reference splits.}
We report two configurations that do not match the target budget and therefore serve as reference points rather than direct competitors.
\emph{Full dataset} uses the complete Major-TOM archive; proportionally longer training did not help, as validation performance peaked within the optimization budget used for the curated subsets.
\emph{Official subset} uses Core-S2L2A-249k, the $249$k-tile subset released with Major-TOM \citep{francis2024majortom}, corresponding to approximately $11\%$ of the full archive. This subset is constructed through systematic geographic subsampling of the Major-TOM grid and thus provides a natural reference for geographically curated pretraining data.

\paragraph{Geographic priors.}
We consider three hand-designed geographic strategies. \emph{Uniform} partitions the globe into latitude-longitude bins and allocates the budget equally across occupied bins. \emph{Europe}, motivated by the strong geographic transfer reported by \citet{kaur2026pretrainwhere}, samples uniformly within $25^\circ$W--$45^\circ$E and $34^\circ$--$72^\circ$N. Finally, motivated by the effectiveness of urban imagery for SSL pretraining \citep{manas2021seasonal,wang2023ssl4eo,stewart2023ssl4eol}, \emph{Urban} retains the observations with the largest built-up fraction according to ESA WorldCover \citep{zanaga2022worldcover}.

\paragraph{Metadata stratification.}
We also consider curation based on geographic metadata, following the widespread use of such attributes to construct diverse EO datasets \citep{schmitt2019sen12ms,blumenstiel2025terramesh,jakubik2025terramind}. We use three categorical attributes: \emph{Climate}, defined by the K\"oppen--Geiger climate zones at $1$\,km resolution \citep{beck2023koppen}; \emph{Land cover}, from ESA WorldCover \citep{zanaga2022worldcover}; and \emph{Soil}, given by the most probable soil class from SoilGrids \citep{poggio2021soilgrids}. Each observation is assigned the majority class over its footprint.

For each attribute, classes receive equal shares of the budget whenever possible, with the allocation capped by the number of available observations and the remainder redistributed among the other classes. This deliberately flattens the metadata distribution, in the flavor of balanced-cluster curation \citep{vo2024automatic}.

\paragraph{Semantic curation.}
We evaluate feature-based curation methods using UniverSat or DINOv2 representations. We adapt SemDeDup~\citep{abbas2023semdedup} to our fixed-budget setting and compare it with hierarchical $k$-means~\citep{vo2024automatic}, which recursively partitions the embedding space and samples across clusters to promote semantic coverage. Where indicated, we additionally apply perceptual-hash (\emph{pHash}) deduplication~\citep{zauner2010phash} beforehand to remove near-identical images.

\begin{table}[t]
\centering
\newcommand{\applycolorsetCorrelation}[1]{%
  \pgfmathsetmacro{\t}{max(0,min(1,#1))}%
  \pgfmathsetmacro{\red}{1-\t}%
  \edef\mycellcolor{\noexpand\cellcolor[rgb]{\red,1,1}}%
  \mycellcolor%
}

\newcommand{\applycolorCorrelation}[1]{%
  \applycolorsetCorrelation{#1}%
  \makebox[\applycolorwidth][c]{#1}%
}
\newcommand{\applycolorCorrelationbf}[1]{%
  \applycolorsetCorrelation{#1}%
  \makebox[\applycolorwidth][c]{\textbf{#1}}%
}

\footnotesize{
\setlength{\tabcolsep}{2.5pt}
\renewcommand{\applycolorwidth}{2.0em}
\begin{tabular}{l l c c c c c c}
\toprule
Criterion & Sampling
& \makecell{Geo.\\cells}
& Climate
& Soil
& \makecell{Land\\cover}
& \makecell{Built-\\up}
& Europe \\
\midrule

Uniform & random
& \applycolorCorrelationbf{0.26}
& \applycolorCorrelation{0.08}
& \applycolorCorrelation{0.02}
& \applycolorCorrelation{0.03}
& \applycolorCorrelation{0.00}
& \applycolorCorrelation{0.00} \\\greyrule

Climate & stratified
& \applycolorCorrelation{0.17}
& \applycolorCorrelationbf{0.29}
& \applycolorCorrelation{0.05}
& \applycolorCorrelation{0.04}
& \applycolorCorrelation{0.01}
& \applycolorCorrelation{0.01} \\

Soil & stratified
& \applycolorCorrelation{0.11}
& \applycolorCorrelation{0.05}
& \applycolorCorrelationbf{0.25}
& \applycolorCorrelation{0.04}
& \applycolorCorrelation{0.01}
& \applycolorCorrelation{0.00} \\

Land cover & stratified
& \applycolorCorrelation{0.10}
& \applycolorCorrelation{0.05}
& \applycolorCorrelation{0.06}
& \applycolorCorrelationbf{0.37}
& \applycolorCorrelation{0.17}
& \applycolorCorrelation{0.01} \\

Urban & top-\%
& \applycolorCorrelation{0.32}
& \applycolorCorrelation{0.18}
& \applycolorCorrelation{0.19}
& \applycolorCorrelation{0.38}
& \applycolorCorrelationbf{0.79}
& \applycolorCorrelation{0.04} \\

Europe & subset
& \applycolorCorrelation{0.35}
& \applycolorCorrelation{0.18}
& \applycolorCorrelation{0.10}
& \applycolorCorrelation{0.07}
& \applycolorCorrelation{0.03}
& \applycolorCorrelationbf{0.36} \\\greyrule

DINOv2 & SemDeDup
& \applycolorCorrelation{0.08}
& \applycolorCorrelation{0.04}
& \applycolorCorrelation{0.04}
& \applycolorCorrelation{0.06}
& \applycolorCorrelation{0.00}
& \applycolorCorrelation{0.00} \\

UniverSat & SemDeDup
& \applycolorCorrelation{0.07}
& \applycolorCorrelation{0.04}
& \applycolorCorrelation{0.04}
& \applycolorCorrelation{0.05}
& \applycolorCorrelation{0.01}
& \applycolorCorrelation{0.00} \\\greyrule

UniverSat & \ours{}
& \applycolorCorrelation{0.10}
& \applycolorCorrelation{0.05}
& \applycolorCorrelation{0.06}
& \applycolorCorrelation{0.05}
& \applycolorCorrelation{0.02}
& \applycolorCorrelation{0.00} \\

\bottomrule
\end{tabular}
}

\caption{\textbf{Association between curation decisions and metadata attributes.} All subsets retain $1\%$ of the archive.
For categorical attributes, we report Cram\'er's $V$ between the keep/discard decision and each attribute; for built-up fraction, we report point-biserial correlation. Higher values indicate stronger dependence of the selection on that attribute. \textbf{Bold} marks the attribute explicitly targeted by each handcrafted strategy.}
\label{tab:keep_correlation}
\end{table}

\begin{figure}[t]
    \centering
    \newcommand{\samplab}[2]{\begin{tabular}[t]{@{}c@{}}\scriptsize #1\\[-4.5pt]\tiny #2\end{tabular}}
\setlength{\tabcolsep}{0pt}
\begin{tabular}{@{}l@{\;}c@{\,}c@{\,}c@{\,}c@{\,}c@{}}
    \rotatebox{90}{\makebox[0.185\linewidth][c]{\small Forest}} &
    \includegraphics[width=0.185\linewidth, height=0.185\linewidth]{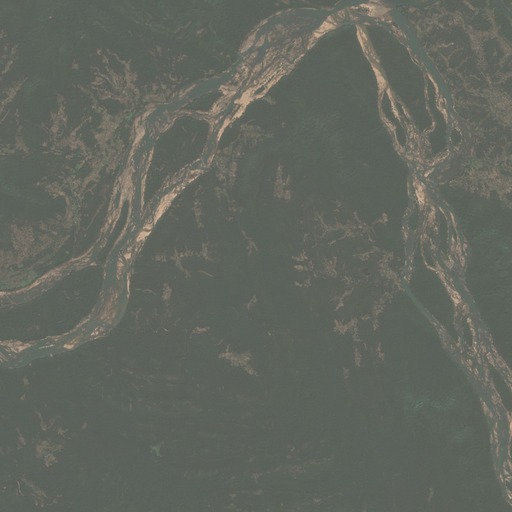}
    &
    \includegraphics[width=0.185\linewidth, height=0.185\linewidth]{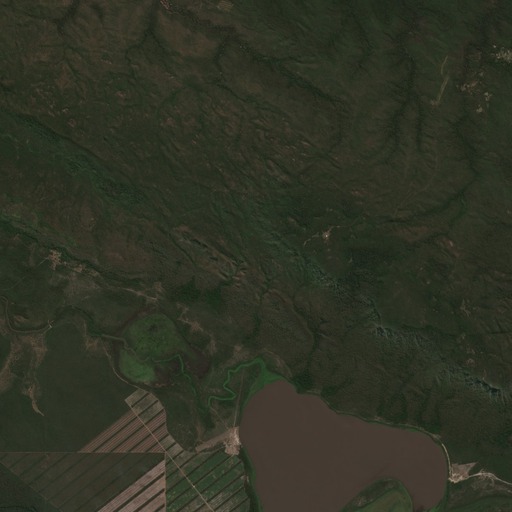}
    &
    \includegraphics[width=0.185\linewidth, height=0.185\linewidth]{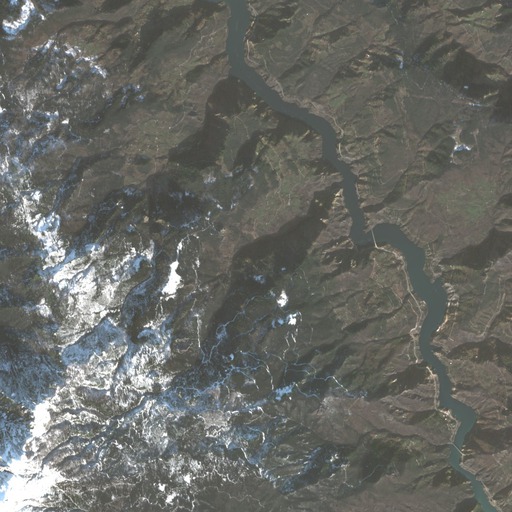}
    &
    \includegraphics[width=0.185\linewidth, height=0.185\linewidth]{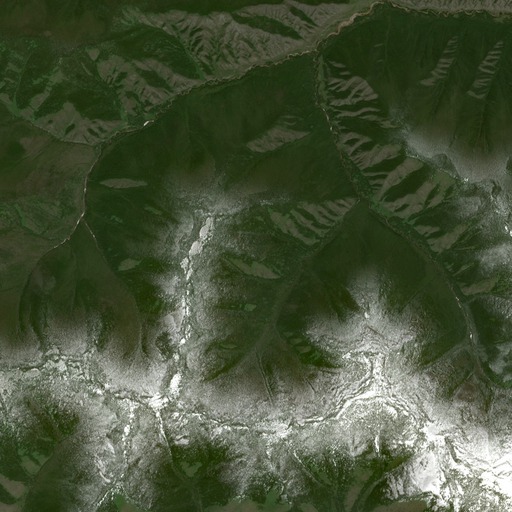}
    &
    \includegraphics[width=0.185\linewidth, height=0.185\linewidth]{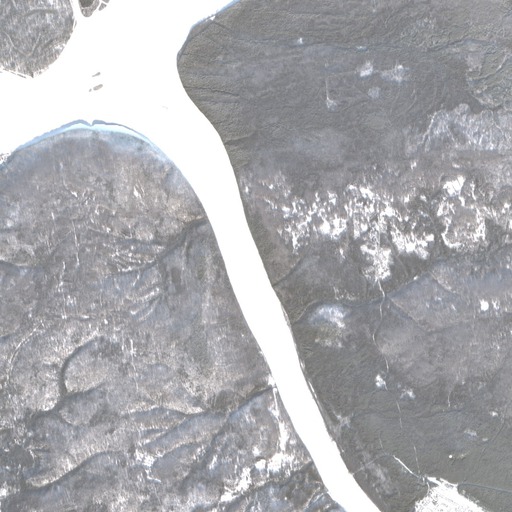}
    \\[-3.5pt]
    & \samplab{Chhattisgarh}{India}
    & \samplab{Santa Cruz}{Bolivia}
    & \samplab{Artvin}{Turkey}
    & \samplab{Tuva}{Russia}
    & \samplab{Krasnoyarsk}{Russia}
    \\[0pt]
    \rotatebox{90}{\makebox[0.185\linewidth][c]{\small Urban}} &
    \includegraphics[width=0.185\linewidth, height=0.185\linewidth]{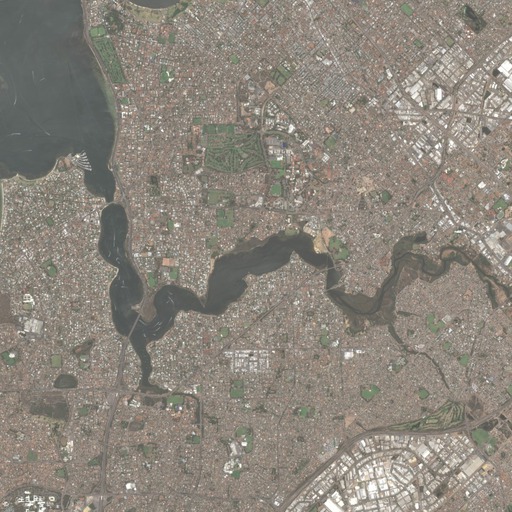}
    &
    \includegraphics[width=0.185\linewidth, height=0.185\linewidth]{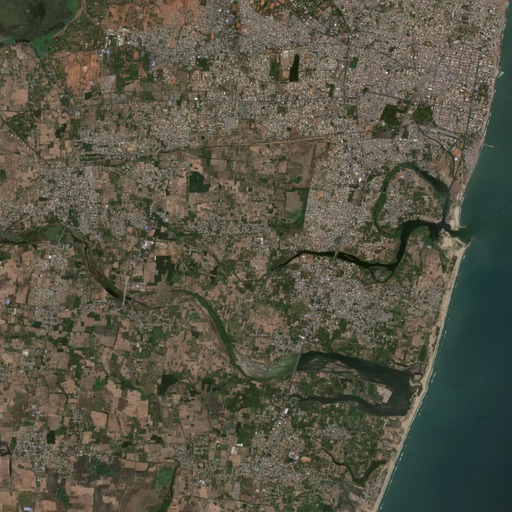}
    &
    \includegraphics[width=0.185\linewidth, height=0.185\linewidth]{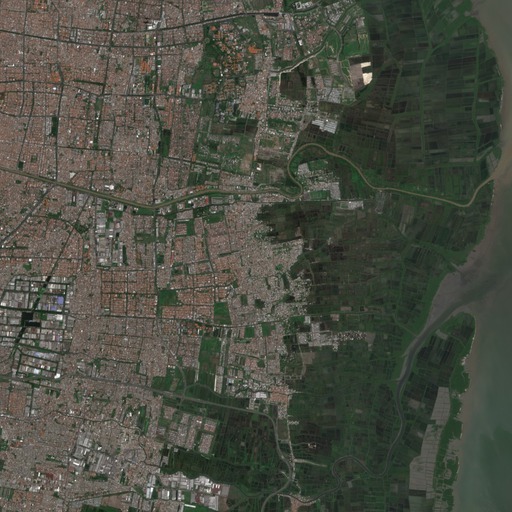}
    &
    \includegraphics[width=0.185\linewidth, height=0.185\linewidth]{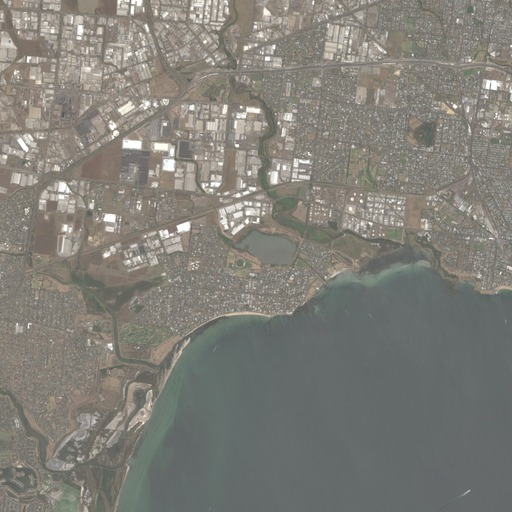}
    &
    \includegraphics[width=0.185\linewidth, height=0.185\linewidth]{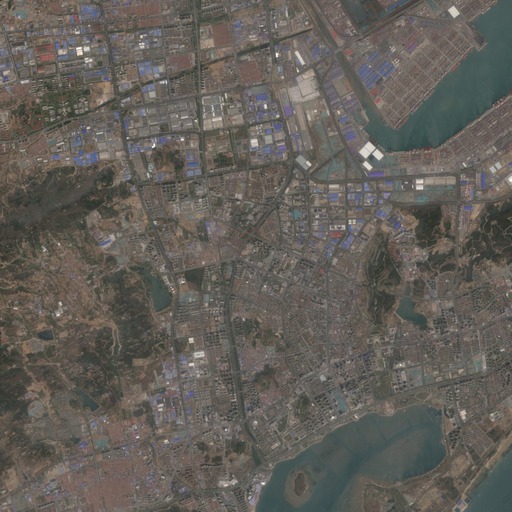}
    \\[-3.5pt]
    & \samplab{Perth}{Australia}
    & \samplab{Puducherry}{India}
    & \samplab{Surabaya}{Indonesia}
    & \samplab{Melbourne}{Australia}
    & \samplab{Qingdao}{China}
\end{tabular}

\caption{\textbf{Selected observations.}
We group observations retained by \ours{} at a $1\%$ budget into forest ($>80\%$ tree cover) and urban ($>30\%$ built-up cover) subsets, then sample five images from each. These observations span diverse biomes and urban morphologies while exhibiting rich local structure.}
    \label{fig:samples}
\end{figure}

\subsection{Results and analysis}
\label{sec:exp:downstream}

We report downstream classification and segmentation performance in
\cref{tab:geobench_curation}. Despite retaining only {$1\%$}
of the original archive, most curation strategies match or outperform
pretraining on the full dataset, showing that the composition of the
pretraining set matters at least as much as its size. \ours{} improves over the full archive by $+2.4$ points on classification and $+2.2$ mIoU on segmentation, the best mean on both tasks.

\paragraph{Curation can offset a {$100\times$} data reduction.}
Uniformly sampling only $1\%$ of the archive degrades performance by $0.9$ points in classification and $2.2$ mIoU in segmentation relative to full-data pretraining. Targeted curation largely closes or reverses this gap. Hand-designed EO priors are already competitive: climate and land-cover stratification improve both classification and segmentation, while urban prioritization favors segmentation at the expense of classification.

These strategies retain markedly different parts of the archive (\cref{fig:location_analysis}). Random sampling thins all regions uniformly, while urban prioritization concentrates on densely built areas. In contrast, \ours{} remains broadly distributed over land and favors coastlines, while discarding open-ocean observations and strongly downsampling ice caps and tropical rainforests, which exhibit low local diversity.

\begin{figure}[t]
\centering
\input{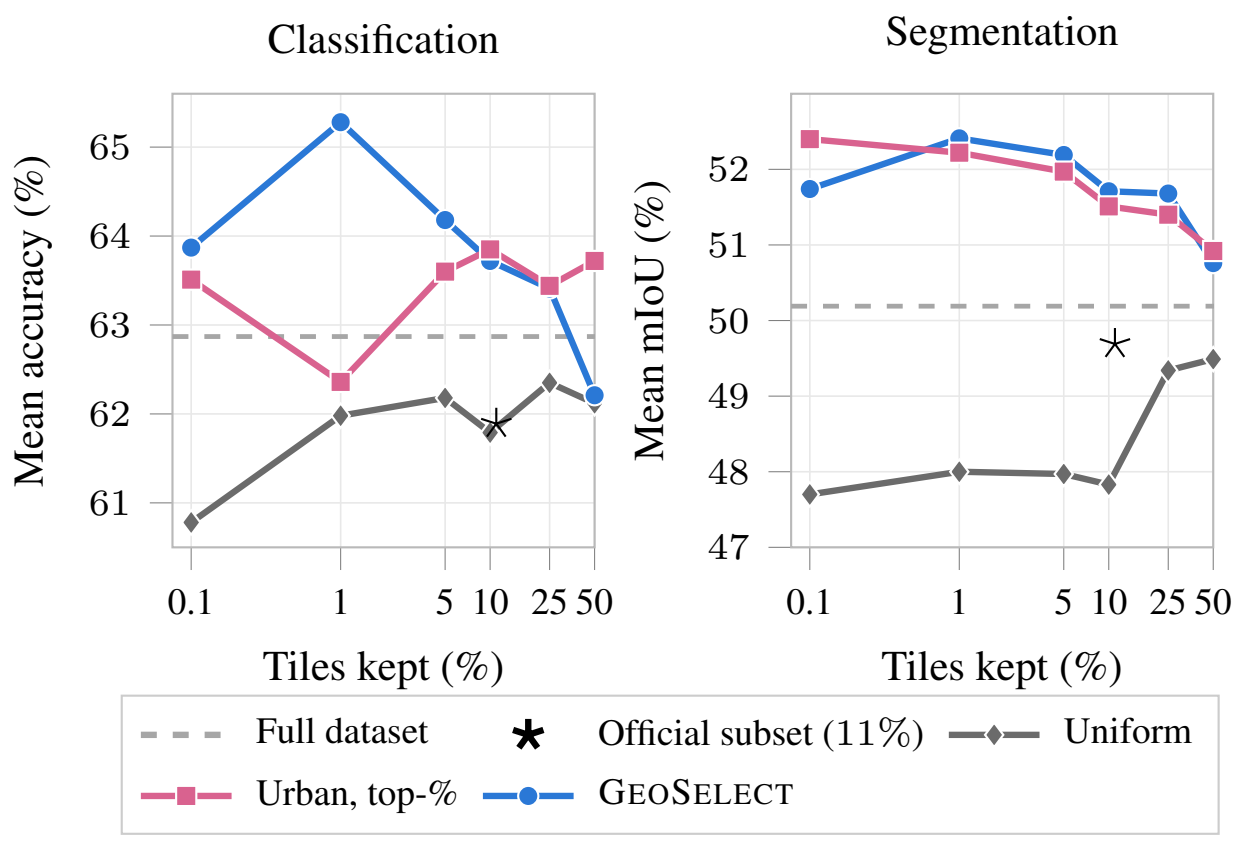}
\caption{\textbf{Downstream performance across curation budgets.}
Mean GEO-Bench classification accuracy (left) and segmentation mIoU (right) as a function of the fraction of Major-TOM retained for pretraining. The dashed line denotes full-archive pretraining and the star the {official subset (Core-S2L2A-249k, $11\%$ of the archive)}.}
\label{fig:keepratio}
\end{figure}

\begin{table}[t]
\centering
\caption{\textbf{Performance across SSL objectives.}
Mean GEO-Bench classification, with a $k$-NN classifier and linear probing (LP), and segmentation, with linear probing, after pretraining ViT-B with DINOv2 or LeJEPA on the same $10\%$ curated subsets.}
\centering
\footnotesize{
\setlength{\tabcolsep}{2pt}
\newcommand{\dgain}[1]{\,\textcolor{green!60!black}{\scriptsize#1}}
\resizebox{\linewidth}{!}{%
\begin{tabular}{l l c @{\hspace{4pt}} c c c @{\hspace{4pt}} c c c}
\toprule
\multicolumn{3}{c}{Curation} & \multicolumn{3}{c}{DINOv2} & \multicolumn{3}{c}{LeJEPA} \\
\cmidrule(lr){1-3}\cmidrule(lr){4-6}\cmidrule(lr){7-9}
 & & & \multicolumn{2}{c}{Classif.} & Seg. & \multicolumn{2}{c}{Classif.} & Seg. \\
\cmidrule(lr){4-5}\cmidrule(lr){6-6}\cmidrule(lr){7-8}\cmidrule(lr){9-9}
Criterion & Sampling & Kept & $k$-NN & LP & LP & $k$-NN & LP & LP \\
\midrule
\multicolumn{2}{l}{\emph{Full dataset}} & 100\% & 57.3 & 63.8 & 48.6 & 55.6 & 63.3 & 48.4 \\\greyrule
Urban & top-\% & 10\% & 61.0\dgain{+3.7} & 67.3\dgain{+3.5} & 51.8\dgain{+3.2} & 58.6\dgain{+3.0} & 65.7\dgain{+2.4} & 50.2\dgain{+1.8} \\\greyrule
UniverSat & SDD & 10\% & 60.2\dgain{+2.9} & 65.7\dgain{+1.9} & 50.8\dgain{+2.2} & 58.8\dgain{+3.2} & 64.5\dgain{+1.2} & 49.8\dgain{+1.4} \\
UniverSat & \bf \ours{} & 10\% & \textbf{61.1}\dgain{+3.8} & \textbf{67.6}\dgain{+3.8} & \textbf{52.0}\dgain{+3.4} & \textbf{60.1}\dgain{+4.5} & \textbf{67.3}\dgain{+4.0} & \textbf{51.0}\dgain{+2.6} \\
\bottomrule
\end{tabular}
}
}

\label{tab:ssl_transfer}
\end{table}

\begin{table}[t]
\centering
\caption{\textbf{Cross-modality curation from optical to radar.}
Sentinel-1 MAE pretraining on the full archive or on $1\%$ subsets curated from Sentinel-1 or co-registered Sentinel-2 embeddings. Evaluation uses the S1-native reBEN-S1 benchmark, with a $k$-NN classifier on pooled patch tokens (details in \cref{sec:supp:knn}) and linear probing (LP), and Sen1Floods11, with linear probing.}
\label{tab:modality_transfer}
\centering
\small{
\setlength{\tabcolsep}{2.5pt}
\newcommand{\dgain}[1]{\,\textcolor{green!60!black}{\scriptsize#1}}
\newcommand{\dloss}[1]{\,\textcolor{red!70!black}{\scriptsize#1}}
\newcommand{\dzero}[1]{\,{\scriptsize#1}}
\begin{tabular}{l l c @{\quad} c c @{\quad} c}
\toprule
\multicolumn{3}{c}{Pretraining} & \multicolumn{2}{c}{Classification} & Segmentation \\
\cmidrule(lr){1-3}\cmidrule(lr){4-5}\cmidrule(lr){6-6}
 & & & \multicolumn{2}{c}{reBEN-S1 $\mu$F1} & Sen1Floods11 \\
Archive & Curated from & Kept & $k$-NN & LP & LP mIoU \\
\midrule
Sentinel-1 & \emph{(full archive)} & 100\% & 54.0 & 59.6 & 79.0 \\\greyrule
Sentinel-1 & Sentinel-1 & 1\% & \textbf{56.3}\dgain{+2.3} & \textbf{61.2}\dgain{+1.6} & 79.0\dzero{0.0} \\\greyrule
Sentinel-1 & Sentinel-2 & 1\% & 53.5\dloss{-0.5} & 60.3\dgain{+0.7} & \textbf{79.1}\dgain{+0.1} \\
\bottomrule
\end{tabular}
}

\end{table}

\paragraph{Embedding-based curation.}
We now compare curation methods that directly operate in feature spaces, without explicit geographic or metadata criteria. As seen in \cref{tab:geobench_curation}, hierarchical $k$-means with UniverSat features improves classification by $+0.4$ points and segmentation by $+0.4$ mIoU; adding pHash increases the classification gain to $+0.7$ points. SemDeDup depends strongly on the representation used for curation: DINOv2 yields $+0.3$ points in classification but $-0.8$ mIoU in segmentation, whereas UniverSat reaches $+0.9$ and $+1.5$, respectively. With DINOv2, SemDeDup keeps open-water observations close to their share of the archive ($30\%$ against $33\%$), whereas UniverSat reduces it to $12\%$, suggesting that RGB features do not identify these observations as redundant.

\ours{} exceeds the strongest hand-designed baselines by $1.3$ points in classification (land-cover stratification) and $0.2$ mIoU in segmentation (urban prioritization).

These gains are not explained by geographic matching. On the European BigEarthNet benchmark, \ours{} gains $+2.7$ points, while the Europe-only split gives $-0.1$; \ours{}'s split is also nearly independent of the Europe indicator (Cram\'er's $V<0.01$, \cref{tab:keep_correlation}). The gains apply across four continents: Europe (EuroSAT and PV4GER), Asia (ForestNet), Africa (Cashew), and North America (Chesapeake).

\paragraph{Semantic curation differs from metadata heuristics.}
\label{sec:exp:correlation}
One might expect feature-based curation to be similar to standard EO heuristics, such as balancing climates or favoring urban areas. We test this by measuring the association between each method's keep/discard decisions and major metadata attributes (\cref{tab:keep_correlation}). As expected, handcrafted strategies are strongly associated with the attribute they explicitly target, whereas all feature-based methods show only weak associations with any metadata attribute (at most $0.10$).

More importantly, the feature-based methods have very similar association profiles despite substantially different downstream performance. Their gains therefore cannot be explained by better matching any of these metadata heuristics, but instead arise from distinctions within metadata classes that these coarse attributes do not capture{: the forest and urban observations retained by \ours{} span diverse biomes and urban morphologies (\cref{fig:samples})}. Consistently, \ours{} leaves most soil, land-cover, and climate distributions close to those of the full archive, except for particularly homogeneous classes such as ice caps and tropical rainforest, {which are downsampled to $13\%$ and $33\%$ of their share in the archive} because of their low local diversity (\cref{fig:keep_histograms,fig:qual_climates}).

\paragraph{Performance across curation budgets.}
\Cref{fig:keepratio} shows that \ours{} remains effective across a wide range of budgets. From $0.1\%$ to $25\%$ of the archive, it outperforms full-archive pretraining on both classification and segmentation, with the largest gains at $1\%$. Even at $0.1\%$, \ie, $2{,}246$ observations each revisited as about $20$k random crops, it improves by $+1.0$ points and $+1.5$ mIoU. Performance decreases as the budget grows, and the classification advantage disappears at $50\%$, consistent with curation becoming less selective. Uniform sampling remains substantially weaker throughout, confirming that the gains arise from which observations are retained rather than from reducing the dataset alone. Urban prioritization is particularly competitive for segmentation, where it performs comparably to \ours{} across budgets, but it {falls behind on classification at budgets of $5\%$ or less}.

\paragraph{Robustness across SSL objectives.}
\label{sec:exp:ssl_transfer}
We extend our evaluation beyond MAE by pretraining a ViT-B with two additional SSL objectives: DINOv2~\citep{oquab2024dinov2} with register tokens~\citep{darcet2024registers}, and LeJEPA~\citep{balestriero2025lejepa} (details in \cref{sec:supp:pretraining}).
\Cref{tab:ssl_transfer} shows that all curated subsets outperform full-archive pretraining for both objectives, with larger gains than for MAE at the same $10\%$ budget.
This is consistent with view-based SSL objectives benefiting particularly from observations whose different crops contain varied content.
\ours{} performs best for both objectives, improving linear-probing classification by $+3.8$ and $+4.0$ points and segmentation by $+3.4$ and $+2.6$ mIoU, respectively.
Urban prioritization is competitive with DINOv2 but falls further behind with LeJEPA, while SemDeDup consistently underperforms \ours{}.
Under $k$-NN evaluation, \ours{} performs best with LeJEPA and on par with urban prioritization with DINOv2, ahead of SemDeDup with both objectives (\cref{tab:ssl_transfer}; per-dataset results in \cref{tab:knn}).

\begin{table}[t]
\centering
\caption{\textbf{Combining global and local diversity.}
Mean GEO-Bench performance after MAE pretraining on $1\%$ of Major-TOM, with gains over the full archive. SemDeDup ranks observations by redundancy $m_i$; the local filter first discards the less diverse half by $L_i$. The last two variants retain SemDeDup's cluster budgets but rank observations by local diversity $L_i$ alone or the joint score $S_i$, which also penalizes near-duplicates.}
\label{tab:ablation}
\centering
\footnotesize{
\setlength{\tabcolsep}{3pt}
\newcommand{\dgain}[1]{\,\textcolor{green!60!black}{\scriptsize#1}}
\begin{tabular}{l c c c c}
\toprule
Selection & \makecell{Local \\ filter} & \makecell{Within-cluster \\ ranking} & Cls. & Seg. \\
\midrule
\emph{Full dataset} & & & 62.9 & 50.2 \\
\midrule
SemDeDup & none & 	Lowest \(m_i\) & 63.8\dgain{+0.9} & 51.7\dgain{+1.5} \\
SemDeDup & Top 50\% by \(L_i\) & Lowest \(m_i\) & 65.0\dgain{+2.1} & 52.2\dgain{+2.0} \\
Local ranking & none & Highest \(L_i\) & 64.9\dgain{+2.0} & 52.5\dgain{+2.3} \\
\ours{} & tie breaker in $G_i$ & Highest \(S_i\) & 65.3\dgain{+2.4} & 52.4\dgain{+2.2} \\
\bottomrule
\end{tabular}
}

\end{table}

\paragraph{Cross-modality curation.}
\label{sec:exp:modality_transfer}
We investigate whether \ours{} extends to radar imagery and whether curation based on optical observations can also benefit radar pretraining. We pretrain MAE models on the full Sentinel-1 archive of Major-TOM (Core-S1RTC) or on $1\%$ subsets curated using \ours{} applied to UniverSat embeddings of either Sentinel-1 observations or co-registered Sentinel-2 observations. We evaluate on the S1-native reBEN-S1~\citep{clasen2024reben} and Sen1Floods11~\citep{bonafilia2020sen1floods11} benchmarks (\cref{tab:modality_transfer}).

Curating from Sentinel-1 embeddings improves land-cover classification despite using $100\times$ less data, by $+2.3$ $\mu$F1 with $k$-NN and $+1.6$ with linear probing. Curating from co-registered Sentinel-2 embeddings also improves linear probing, by $+0.7$, but not $k$-NN classification. Both subsets preserve flood segmentation performance, within $0.1$ mIoU of full-archive pretraining. Semantic curation thus transfers to radar imagery, and optical embeddings can guide radar data selection, although embeddings of the target modality yield larger and more consistent gains.

\begin{table}[t]
\centering
\caption{\textbf{Curation of S2-NAIP.}
Mean GEO-Bench classification, with a $k$-NN classifier and linear probing (LP), and segmentation, with linear probing, after MAE pretraining on the full S2-NAIP archive or on $1\%$ subsets of it. \emph{Random} draws observations uniformly from the archive.}
\label{tab:s2naip_summary}
\centering
\footnotesize{
\setlength{\tabcolsep}{2pt}
\newcommand{\dgain}[1]{\,\textcolor{green!60!black}{\scriptsize#1}}
\newcommand{\dloss}[1]{\,\textcolor{red!70!black}{\scriptsize#1}}
\newcommand{\dzero}[1]{\,{\scriptsize#1}}
\begin{tabular}{l l c @{\hspace{6pt}} c c @{\hspace{6pt}} c}
\toprule
\multicolumn{3}{c}{Curation} & \multicolumn{2}{c}{Classification} & Segmentation \\
\cmidrule(lr){1-3}\cmidrule(lr){4-5}\cmidrule(lr){6-6}
Criterion & Sampling & Kept & $k$-NN & LP & LP \\
\midrule
\multicolumn{2}{l}{\emph{Full dataset}} & 100\% & 61.9 & 69.0 & 51.1 \\\greyrule
\multicolumn{2}{l}{Random} & 1\% & 62.1\dgain{+0.2} & 68.8\dloss{-0.2} & 50.4\dloss{-0.7} \\\greyrule
Urban & top-\% & 1\% & 63.1\dgain{+1.2} & 69.1\dgain{+0.1} & 48.5\dloss{-2.6} \\\greyrule
UniverSat & SDD & 1\% & 62.5\dgain{+0.6} & \textbf{69.4}\dgain{+0.4} & 50.2\dloss{-0.9} \\
UniverSat & \bf \ours{} & 1\% & \textbf{63.3}\dgain{+1.4} & 68.9\dloss{-0.1} & \textbf{51.3}\dgain{+0.2} \\
\bottomrule
\end{tabular}
}

\end{table}

\paragraph{Complementary roles of local and global diversity.}
\label{sec:exp:two_div}
Global and local diversity favor different observations. Among those retained by SemDeDup, $24\%$ fall in the less locally diverse half of the archive. Conversely, selecting solely by $L_i$ covers $45\%$ fewer geographic regions than SemDeDup (\cref{tab:two_div}). \ours{} balances both: only $5\%$ of its observations fall in the less locally diverse half, while retaining $93\%$ of SemDeDup's geographic coverage.

\Cref{tab:ablation} evaluates different ways to incorporate local diversity into SemDeDup. Standard SemDeDup improves classification and segmentation by $+0.9$ and $+1.5$ points over full-archive pretraining. Filtering out the less locally diverse half before applying SemDeDup increases these gains to $+2.1$ and $+2.0$. 

Alternatively, retaining SemDeDup's cluster budgets but ranking observations by $L_i$ yields $+2.0$ and $+2.3$, on par with our proposed approach. Local diversity thus accounts for most of the gain over SemDeDup.
However, ranking by $L_i$ alone can retain near-duplicates when both are locally diverse. This appears uncommon in our archive, as locally diverse observations tend to be spatially distinctive. Nevertheless, explicitly penalizing redundancy remains desirable, especially for multi-temporal archives where repeated observations may be both locally diverse and highly similar.

\paragraph{Curating multi-temporal data.}
\label{sec:exp:s2naip}
We also evaluate \ours{} on S2-NAIP~\citep{wolters2023zooming}, a multi-temporal archive of small tiles covering the contiguous United States. After removing duplicate acquisitions and filtering clouds\footnote{\raggedright\href{https://github.com/sentinel-hub/sentinel2-cloud-detector}{\nolinkurl{github.com/sentinel-hub/sentinel2-cloud-detector}}}, it contains $302$M observations across $19.3$M locations, with a median of $15$ acquisitions per location. We pretrain MAE on the full archive and on $1\%$ subsets (\cref{tab:s2naip_summary}; details in \cref{sec:supp:s2naip}). Even random sampling largely preserves classification performance, suggesting substantial spatial and temporal redundancy and limited room for improvement on single-date benchmarks. However, random sampling loses $0.7$ mIoU in segmentation, whereas \ours{} is the only method to preserve full-archive performance. \ours{} also performs best under $k$-NN evaluation, while matching full-archive classification performance under linear probing.

\paragraph{Efficiency.}
\label{sec:exp:efficiency}
Curation incurs a modest, one-time overhead relative to SSL pretraining. Its main cost is extracting patch-token statistics with UniverSat, requiring approximately $65$ A100 GPU-hours at $20$\,m resolution, far less than pretraining a foundation model. Clustering and selection take less than one minute on a GPU. Moreover, releasing patch-token statistics alongside global embeddings would allow the future curation of EO archives without rerunning the encoder. Once curated, the subset can be reused across pretraining runs, reducing storage and data transfer by $100\times$ while yielding better representations at matched training compute.

\paragraph{Limitations.}
Our study focuses on Major-TOM, chosen for its global coverage, limited prior curation, and publicly available embeddings. Evaluating already curated archives, such as SSL4EO-S12~\citep{wang2023ssl4eo}, would make it harder to isolate the effects of individual curation methods. We evaluate frozen representations through linear probing; assessing whether our conclusions extend to fine-tuning would require substantially more computation. Finally, pretraining costs limit independent runs, so small differences between methods should be interpreted cautiously.

\section{Conclusion}
\label{sec:conclusion}
We investigated the impact of data curation on self-supervised EO pretraining, comparing geographic, metadata-based, and semantic selection strategies under a common evaluation protocol. We introduced \ours{}, a simple semantic curation method that jointly promotes global diversity across observations and local diversity within each observation. On Major-TOM, pretraining on only $1\%$ of the archive yields better representations than full-data pretraining and outperforms alternative curation strategies at the same budget. On the multi-temporal S2-NAIP, it retains full-archive performance at the same budget and is the only method to preserve segmentation performance. As public access to EO data grows, our results show that curation can yield smaller, more easily shared pretraining datasets that also improve representation quality.

\section*{Acknowledgments}
This work was supported by the French Space Agency PhD program (CNES) and the French National Institute of Geographic and Forest Information (IGN). 
This work was supported by ANR project SHARP (ANR-23-PEIA-0008) in the context of the PEPRIA.
It was granted access to the HPC resources of IDRIS under the allocations AD011014719R2, AD011015196R2, and A0200116178 made by GENCI. We thank Jordi Inglada for inspiring discussions and valuable feedback.

\FloatBarrier
{
    \small
    \bibliographystyle{ieeenat_fullname}
    \bibliography{biblio}

\begin{thebibliography}{58}
\providecommand{\natexlab}[1]{#1}
\providecommand{\url}[1]{\texttt{#1}}
\expandafter\ifx\csname urlstyle\endcsname\relax
  \providecommand{\doi}[1]{doi: #1}\else
  \providecommand{\doi}{doi: \begingroup \urlstyle{rm}\Url}\fi

\bibitem[Abbas et~al.(2023)Abbas, Tirumala, Simig, Ganguli, and
  Morcos]{abbas2023semdedup}
Amro Abbas, Kushal Tirumala, D{\'a}niel Simig, Surya Ganguli, and Ari~S.
  Morcos.
\newblock {SemDeDup: D}ata-efficient learning at web-scale through semantic
  deduplication.
\newblock In \emph{ICLR Workshop MRL}, 2023.

\bibitem[Abbas et~al.(2024)Abbas, Rusak, Tirumala, Brendel, Chaudhuri, and
  Morcos]{abbas2024dbp}
Amro Abbas, Evgenia Rusak, Kushal Tirumala, Wieland Brendel, Kamalika
  Chaudhuri, and Ari~S. Morcos.
\newblock Effective pruning of web-scale datasets based on complexity of
  concept clusters.
\newblock In \emph{ICLR}, 2024.

\bibitem[Adler et~al.(2025)Adler, Hackel, Sumbul, and
  Demir]{redundancyaware2025}
Mathis~J{\"u}rgen Adler, Leonard Hackel, Gencer Sumbul, and Beg{\"u}m Demir.
\newblock Redundancy-aware pretraining of vision-language foundation models in
  remote sensing.
\newblock In \emph{IGARSS}, 2025.

\bibitem[Astruc et~al.(2025)Astruc, Gonthier, Mallet, and
  Landrieu]{astruc2024anysat}
Guillaume Astruc, Nicolas Gonthier, Cl{\'e}ment Mallet, and Loic Landrieu.
\newblock {AnySat: O}ne {E}arth observation model for many resolutions, scales,
  and modalities.
\newblock In \emph{CVPR}, 2025.

\bibitem[Astruc et~al.(2026)Astruc, Trulls, Hosang, Landrieu, and
  Sarlin]{astruc2026unigeoclip}
Guillaume Astruc, Eduard Trulls, Jan Hosang, Loic Landrieu, and Paul-Edouard
  Sarlin.
\newblock {UniGeoCLIP: U}nified geospatial contrastive learning.
\newblock In \emph{CVPR Workshop EarthVision}, 2026.

\bibitem[Balestriero and LeCun(2025)]{balestriero2025lejepa}
Randall Balestriero and Yann LeCun.
\newblock {LeJEPA}: Provable and scalable self-supervised learning without the
  heuristics.
\newblock \emph{arXiv:2511.08544}, 2025.

\bibitem[Barseghyan et~al.(2025)Barseghyan, Vanyan, Tamazyan, Shelhamer, and
  Khachatrian]{barseghyan2025less}
Alvard Barseghyan, Ani Vanyan, Hakob Tamazyan, Evan Shelhamer, and Hrant
  Khachatrian.
\newblock Less is more? {D}ata specialization for self-supervised remote
  sensing models.
\newblock In \emph{ICML Workshop TerraBytes}, 2025.

\bibitem[Beck et~al.(2023)Beck, McVicar, Vergopolan, Berg, Lutsko, Dufour,
  Zeng, Jiang, van Dijk, and Miralles]{beck2023koppen}
Hylke~E. Beck, Tim~R. McVicar, Noemi Vergopolan, Alexis Berg, Nicholas~J.
  Lutsko, Ambroise Dufour, Zhenzhong Zeng, Xin Jiang, Albert I. J.~M. van Dijk,
  and Diego~G. Miralles.
\newblock High-resolution (1 km) {K}\"oppen--{G}eiger maps for 1901--2099 based
  on constrained {CMIP6} projections.
\newblock \emph{Scientific Data}, 2023.

\bibitem[Blumenstiel et~al.(2025)Blumenstiel, Fraccaro, Marsocci, Jakubik,
  Maurogiovanni, Czerkawski, Sedona, Cavallaro, Brunschwiler,
  Bernab{\'e}-Moreno, and Long{\'e}p{\'e}]{blumenstiel2025terramesh}
Benedikt Blumenstiel, Paolo Fraccaro, Valerio Marsocci, Johannes Jakubik,
  Stefano Maurogiovanni, Mikolaj Czerkawski, Rocco Sedona, Gabriele Cavallaro,
  Thomas Brunschwiler, Juan Bernab{\'e}-Moreno, and Nicolas Long{\'e}p{\'e}.
\newblock {TerraMesh: A} planetary mosaic of multimodal {E}arth observation
  data.
\newblock In \emph{CVPR Workshop EarthVision}, 2025.

\bibitem[Bonafilia et~al.(2020)Bonafilia, Tellman, Anderson, and
  Issenberg]{bonafilia2020sen1floods11}
Derrick Bonafilia, Beth Tellman, Tyler Anderson, and Erica Issenberg.
\newblock {Sen1Floods11}: {A} georeferenced dataset to train and test deep
  learning flood algorithms for {Sentinel-1}.
\newblock In \emph{CVPR Workshops}, 2020.

\bibitem[Brown et~al.(2020)Brown, Mann, Ryder, Subbiah, Kaplan, Dhariwal,
  Neelakantan, Shyam, Sastry, Askell, et~al.]{brown2020language}
Tom Brown, Benjamin Mann, Nick Ryder, Melanie Subbiah, Jared~D. Kaplan,
  Prafulla Dhariwal, Arvind Neelakantan, Pranav Shyam, Girish Sastry, Amanda
  Askell, et~al.
\newblock Language models are few-shot learners.
\newblock In \emph{NeurIPS}, 2020.

\bibitem[Burgert et~al.(2026)Burgert, Hackel, Rota, and
  Demir]{burgert2026georank}
Tom Burgert, Leonard Hackel, Paolo Rota, and Beg{\"u}m Demir.
\newblock Rank-based geographical regularization: {R}evisiting contrastive
  self-supervised learning for multispectral remote sensing imagery.
\newblock In \emph{WACV}, 2026.

\bibitem[Caron et~al.(2019)Caron, Bojanowski, Mairal, and Joulin]{caron}
Mathilde Caron, Piotr Bojanowski, Julien Mairal, and Armand Joulin.
\newblock Unsupervised pre-training of image features on non-curated data.
\newblock In \emph{ICCV}, 2019.

\bibitem[Clasen et~al.(2025)Clasen, Hackel, Burgert, Sumbul, Demir, and
  Markl]{clasen2024reben}
Kai~Norman Clasen, Leonard Hackel, Tom Burgert, Gencer Sumbul, Beg{\"u}m Demir,
  and Volker Markl.
\newblock {reBEN: Refined BigEarthNet} dataset for remote sensing image
  analysis.
\newblock In \emph{IGARSS}, 2025.

\bibitem[Czerkawski et~al.(2024)Czerkawski, Kluczek, and
  Bojanowski]{czerkawski2024global}
Mikolaj Czerkawski, Marcin Kluczek, and J{\k{e}}drzej~S. Bojanowski.
\newblock Global and dense embeddings of {E}arth: {M}ajor {TOM} floating in the
  latent space.
\newblock \emph{arXiv:2412.05600}, 2024.

\bibitem[Darcet et~al.(2024)Darcet, Oquab, Mairal, and
  Bojanowski]{darcet2024registers}
Timoth{\'e}e Darcet, Maxime Oquab, Julien Mairal, and Piotr Bojanowski.
\newblock Vision transformers need registers.
\newblock In \emph{ICLR}, 2024.

\bibitem[Francis and Czerkawski(2024)]{francis2024majortom}
Alistair Francis and Mikolaj Czerkawski.
\newblock Major {TOM}: {E}xpandable datasets for {E}arth observation.
\newblock In \emph{IGARSS}, 2024.

\bibitem[Gadre et~al.(2023)Gadre, Ilharco, Fang, Hayase, Smyrnis, Nguyen,
  et~al.]{gadre2023datacomp}
Samir~Yitzhak Gadre, Gabriel Ilharco, Alex Fang, Jonathan Hayase, Georgios
  Smyrnis, Thao Nguyen, et~al.
\newblock {DataComp: I}n search of the next generation of multimodal datasets.
\newblock In \emph{NeurIPS Datasets and Benchmarks}, 2023.

\bibitem[Gao et~al.(2020)Gao, Biderman, Black, Golding, Hoppe, Foster, Phang,
  He, Thite, Nabeshima, et~al.]{gao2020pile}
Leo Gao, Stella Biderman, Sid Black, Laurence Golding, Travis Hoppe, Charles
  Foster, Jason Phang, Horace He, Anish Thite, Noa Nabeshima, et~al.
\newblock The {P}ile: {A}n 800{GB} dataset of diverse text for language
  modeling.
\newblock \emph{arXiv:2101.00027}, 2020.

\bibitem[Gaydon et~al.(2024)Gaydon, Daab, and Roche]{gaydon2024fractal}
Charles Gaydon, Michel Daab, and Floryne Roche.
\newblock {FRACTAL}: {A}n ultra-large-scale aerial lidar dataset for 3{D}
  semantic segmentation of diverse landscapes.
\newblock \emph{arXiv:2405.04634}, 2024.

\bibitem[Hackel et~al.(2026)Hackel, Burgert, and Demir]{hackel2026slimmability}
Leonard Hackel, Tom Burgert, and Beg{\"u}m Demir.
\newblock How much of a model do we need? {R}edundancy and slimmability in
  remote sensing foundation models.
\newblock In \emph{NeurIPS}, 2026.

\bibitem[He et~al.(2022)He, Chen, Xie, Li, Doll{\'a}r, and Girshick]{he2022mae}
Kaiming He, Xinlei Chen, Saining Xie, Yanghao Li, Piotr Doll{\'a}r, and Ross
  Girshick.
\newblock Masked autoencoders are scalable vision learners.
\newblock In \emph{CVPR}, 2022.

\bibitem[Hoffmann et~al.(2022)Hoffmann, Borgeaud, Mensch, Buchatskaya, Cai,
  Rutherford, de~Las~Casas, Hendricks, Welbl, Clark, et~al.]{chinchilla}
Jordan Hoffmann, Sebastian Borgeaud, Arthur Mensch, Elena Buchatskaya, Trevor
  Cai, Eliza Rutherford, Diego de Las~Casas, Lisa~Anne Hendricks, Johannes
  Welbl, Aidan Clark, et~al.
\newblock An empirical analysis of compute-optimal large language model
  training.
\newblock In \emph{NeurIPS}, 2022.

\bibitem[Jakubik et~al.(2025)Jakubik, Yang, Blumenstiel, Scheurer, Sedona,
  Maurogiovanni, Bosmans, Dionelis, Marsocci, Kopp,
  et~al.]{jakubik2025terramind}
Johannes Jakubik, Felix Yang, Benedikt Blumenstiel, Erik Scheurer, Rocco
  Sedona, Stefano Maurogiovanni, Jente Bosmans, Nikolaos Dionelis, Valerio
  Marsocci, Niklas Kopp, et~al.
\newblock {TerraMind: L}arge-scale generative multimodality for {E}arth
  observation.
\newblock In \emph{ICCV}, 2025.

\bibitem[Jiang et~al.(2024)Jiang, Scheibenreif, and Borth]{jiang2024less}
Xuemei Jiang, Linus Scheibenreif, and Damian Borth.
\newblock Less is more: {A}ctive self-supervised learning in remote sensing.
\newblock In \emph{IGARSS}. IEEE, 2024.

\bibitem[Kaur et~al.(2026)Kaur, Purohit, Muhawenayo, Rolf, and
  Kerner]{kaur2026pretrainwhere}
Amandeep Kaur, Mirali Purohit, Gedeon Muhawenayo, Esther Rolf, and Hannah
  Kerner.
\newblock Pretrain where? {I}nvestigating how pretraining data diversity
  impacts geospatial foundation model performance.
\newblock In \emph{CVPR Workshop EarthVision}, 2026.

\bibitem[Kerdreux et~al.(2025)Kerdreux, Tuel, Febvre, Mouche, and
  Chapron]{kerdreux2025efficient}
Thomas Kerdreux, Alexandre Tuel, Quentin Febvre, Alexis Mouche, and Bertrand
  Chapron.
\newblock Efficient self-supervised learning for {E}arth observation via
  dynamic dataset curation.
\newblock In \emph{CVPR Workshop MORSE}, 2025.

\bibitem[Lacoste et~al.(2023)Lacoste, Lehmann, Rodriguez, Sherwin, Kerner,
  L{\"u}tjens, Irvin, Dao, Alemohammad, Drouin, et~al.]{lacoste2023geo}
Alexandre Lacoste, Nils Lehmann, Pau Rodriguez, Evan Sherwin, Hannah Kerner,
  Bj{\"o}rn L{\"u}tjens, Jeremy Irvin, David Dao, Hamed Alemohammad, Alexandre
  Drouin, et~al.
\newblock {GEO-Bench: T}oward foundation models for {E}arth monitoring.
\newblock In \emph{NeurIPS Datasets and Benchmarks}, 2023.

\bibitem[Lauren{\c{c}}on et~al.(2022)Lauren{\c{c}}on, Saulnier, Wang, Akiki,
  Villanova~del Moral, Le~Scao, Von~Werra, Mou, Gonz{\'a}lez~Ponferrada,
  Nguyen, et~al.]{laurenccon2022bigscience}
Hugo Lauren{\c{c}}on, Lucile Saulnier, Thomas Wang, Christopher Akiki, Albert
  Villanova~del Moral, Teven Le~Scao, Leandro Von~Werra, Chenghao Mou, Eduardo
  Gonz{\'a}lez~Ponferrada, Huu Nguyen, et~al.
\newblock The {BigScience ROOTS} corpus: {A} 1.6 {TB} composite multilingual
  dataset.
\newblock In \emph{NeurIPS Datasets and Benchmarks}, 2022.

\bibitem[Maharana et~al.(2024)Maharana, Yadav, and Bansal]{maharana2024d2}
Adyasha Maharana, Prateek Yadav, and Mohit Bansal.
\newblock {$\mathbb{D}^2$} pruning: Message passing for balancing diversity \&
  difficulty in data pruning.
\newblock In \emph{ICLR}, 2024.

\bibitem[Ma{\~n}as et~al.(2021)Ma{\~n}as, Lacoste, {Gir{\'o}-i-Nieto}, Vazquez,
  and Rodr{\'\i}guez]{manas2021seasonal}
Oscar Ma{\~n}as, Alexandre Lacoste, Xavier {Gir{\'o}-i-Nieto}, David Vazquez,
  and Pau Rodr{\'\i}guez.
\newblock Seasonal contrast: {U}nsupervised pre-training from uncurated remote
  sensing data.
\newblock In \emph{ICCV}, 2021.

\bibitem[Mendieta et~al.(2023)Mendieta, Han, Shi, Zhu, and
  Chen]{mendieta2023gfm}
Mat{\'i}as Mendieta, Boran Han, Xingjian Shi, Yi Zhu, and Chen Chen.
\newblock Towards geospatial foundation models via continual pretraining.
\newblock In \emph{ICCV}, 2023.

\bibitem[Muhtar et~al.(2025)Muhtar, Zhang, Li, Gu, He, Xiao, and
  Zhang]{muhtar2026quality}
Dilxat Muhtar, Enzhuo Zhang, Zhenshi Li, Feng Gu, Yanglangxing He, Pengfeng
  Xiao, and Xueliang Zhang.
\newblock Quality-driven curation of remote sensing vision-language data via
  learned scoring models.
\newblock In \emph{NeurIPS}, 2025.

\bibitem[Oquab et~al.(2024)Oquab, Darcet, Moutakanni, Vo, Szafraniec, Khalidov,
  et~al.]{oquab2024dinov2}
Maxime Oquab, Timoth{\'e}e Darcet, Th{\'e}o Moutakanni, Huy~V. Vo, Marc
  Szafraniec, Vasil Khalidov, et~al.
\newblock {DINOv2: L}earning robust visual features without supervision.
\newblock \emph{TMLR}, 2024.

\bibitem[Penedo et~al.(2023)Penedo, Malartic, Hesslow, Cojocaru, Alobeidli,
  Cappelli, Pannier, Almazrouei, and Launay]{penedo2023refinedweb}
Guilherme Penedo, Quentin Malartic, Daniel Hesslow, Ruxandra Cojocaru, Hamza
  Alobeidli, Alessandro Cappelli, Baptiste Pannier, Ebtesam Almazrouei, and
  Julien Launay.
\newblock The {R}efined{W}eb dataset for {F}alcon {LLM}: {O}utperforming
  curated corpora with web data only.
\newblock In \emph{NeurIPS Datasets and Benchmarks}, 2023.

\bibitem[Perera et~al.(2026)Perera, Witharana, Manos, Pimenta, and
  Liljedahl]{perera2026clustering}
Amal~Shehan Perera, Chandi Witharana, Elias Manos, Michael Pimenta, and Anna~K.
  Liljedahl.
\newblock Clustering guided domain-specific pretrained foundation model for
  very high-resolution {A}rctic remote sensing.
\newblock \emph{arXiv:2605.30467}, 2026.

\bibitem[Perron et~al.(2026)Perron, Astruc, Gonthier, Mallet, and
  Landrieu]{perron2025universat}
Yohann Perron, Guillaume Astruc, Nicolas Gonthier, Cl{\'e}ment Mallet, and Loic
  Landrieu.
\newblock {UniverSat: R}esolution- and modality-agnostic transformers for
  {E}arth observation.
\newblock In \emph{NeurIPS}, 2026.

\bibitem[Plekhanova et~al.(2025)Plekhanova, Robert, Dollinger, Arens, Brun,
  Wegner, and Zimmermann]{plekhanova2025ssl4eco}
Elena Plekhanova, Damien Robert, Johannes Dollinger, Emilia Arens, Philipp
  Brun, Jan~Dirk Wegner, and Niklaus~E. Zimmermann.
\newblock {SSL4Eco: A} global seasonal dataset for geospatial foundation models
  in ecology.
\newblock In \emph{CVPR Workshop EarthVision}, 2025.

\bibitem[Poggio et~al.(2021)Poggio, de~Sousa, Batjes, Heuvelink, Kempen,
  Ribeiro, and Rossiter]{poggio2021soilgrids}
Laura Poggio, Luis~M. de Sousa, Niels~H. Batjes, Gerard B.~M. Heuvelink, Bas
  Kempen, Eloi Ribeiro, and David Rossiter.
\newblock {SoilGrids} 2.0: {P}roducing soil information for the globe with
  quantified spatial uncertainty.
\newblock \emph{SOIL}, 2021.

\bibitem[Purohit et~al.(2025)Purohit, Muhawenayo, Rolf, and
  Kerner]{purohit2025spatial}
Mirali Purohit, Gedeon Muhawenayo, Esther Rolf, and Hannah Kerner.
\newblock How does the spatial distribution of pre-training data affect
  geospatial foundation models?
\newblock In \emph{AAAI Workshop Good Data for Generative AI}, 2025.

\bibitem[Roscher et~al.(2024)Roscher, Ru{\ss}wurm, Gevaert, Kampffmeyer, dos
  Santos, Vakalopoulou, H{\"a}nsch, Hansen, Nogueira, Prexl, and
  Tuia]{roscher2024better}
Ribana Roscher, Marc Ru{\ss}wurm, Caroline Gevaert, Michael Kampffmeyer,
  Jefersson~A. dos Santos, Maria Vakalopoulou, Ronny H{\"a}nsch, Stine Hansen,
  Keiller Nogueira, Jonathan Prexl, and Devis Tuia.
\newblock Better, not just more: {D}ata-centric machine learning for {E}arth
  observation.
\newblock \emph{IEEE Geoscience and Remote Sensing Magazine}, 12\penalty0
  (4):\penalty0 335--355, 2024.

\bibitem[Schmitt et~al.(2019)Schmitt, Hughes, Qiu, and Zhu]{schmitt2019sen12ms}
Michael Schmitt, Lloyd~Haydn Hughes, Chunping Qiu, and Xiao~Xiang Zhu.
\newblock {SEN12MS} -- {A} curated dataset of georeferenced multi-spectral
  {S}entinel-1/2 imagery for deep learning and data fusion.
\newblock \emph{ISPRS Annals}, 2019.

\bibitem[Sim{\'e}oni et~al.(2026)Sim{\'e}oni, Vo, Seitzer, Baldassarre, Oquab,
  Jose, Khalidov, Szafraniec, Yi, Ramamonjisoa, et~al.]{simeoni2025dinov3}
Oriane Sim{\'e}oni, Huy~V. Vo, Maximilian Seitzer, Federico Baldassarre, Maxime
  Oquab, Cijo Jose, Vasil Khalidov, Marc Szafraniec, Seungeun Yi, Micha{\"e}l
  Ramamonjisoa, et~al.
\newblock {DINOv3}.
\newblock \emph{TMLR}, 2026.

\bibitem[Sorscher et~al.(2022)Sorscher, Geirhos, Shekhar, Ganguli, and
  Morcos]{sorscher2022beyond}
Ben Sorscher, Robert Geirhos, Shashank Shekhar, Surya Ganguli, and Ari~S.
  Morcos.
\newblock Beyond neural scaling laws: {B}eating power law scaling via data
  pruning.
\newblock In \emph{NeurIPS}, 2022.

\bibitem[Stewart et~al.(2023)Stewart, Lehmann, Corley, Wang, Chang, Ait
  Ali~Braham, Sehgal, Robinson, and Banerjee]{stewart2023ssl4eol}
Adam~J. Stewart, Nils Lehmann, Isaac~A. Corley, Yi Wang, Yi-Chia Chang, Nassim
  Ait Ali~Braham, Shradha Sehgal, Caleb Robinson, and Arindam Banerjee.
\newblock {SSL4EO-L: D}atasets and foundation models for {L}andsat imagery.
\newblock In \emph{NeurIPS Datasets and Benchmarks}, 2023.

\bibitem[Szwarcman et~al.(2026)Szwarcman, Roy, Fraccaro, G{\'\i}slason,
  Blumenstiel, Ghosal, de~Oliveira, de~Sousa~Almeida, Sedona, Kang,
  et~al.]{szwarcman2024prithvi}
Daniela Szwarcman, Sujit Roy, Paolo Fraccaro, {\TH}orsteinn~El{\'\i}
  G{\'\i}slason, Benedikt Blumenstiel, Rinki Ghosal, Pedro~Henrique de
  Oliveira, Joao~Lucas de Sousa~Almeida, Rocco Sedona, Yanghui Kang, et~al.
\newblock {Prithvi-EO-2.0: A} versatile multitemporal foundation model for
  {E}arth observation applications.
\newblock \emph{IEEE Transactions on Geoscience and Remote Sensing},
  64:\penalty0 1--20, 2026.

\bibitem[Tirumala et~al.(2023)Tirumala, Simig, Aghajanyan, and
  Morcos]{tirumala2023d4}
Kushal Tirumala, D{\'a}niel Simig, Armen Aghajanyan, and Ari~S. Morcos.
\newblock {D4: I}mproving {LLM} pretraining via document de-duplication and
  diversification.
\newblock In \emph{NeurIPS Datasets and Benchmarks}, 2023.

\bibitem[Velazquez et~al.(2025)Velazquez, Rodriguez, Alonso, Gonfaus, Gonzalez,
  Richarte, Marin, Bengio, and Lacoste]{velazquez2025earthview}
Diego Velazquez, Pau Rodriguez, Sergio Alonso, Josep~M. Gonfaus, Jordi
  Gonzalez, Gerardo Richarte, Javier Marin, Yoshua Bengio, and Alexandre
  Lacoste.
\newblock {EarthView: A} large scale remote sensing dataset for
  self-supervision.
\newblock In \emph{WACV Workshops}, 2025.

\bibitem[Vo et~al.(2024)Vo, Khalidov, Darcet, Moutakanni, Smetanin, Szafraniec,
  et~al.]{vo2024automatic}
Huy~V. Vo, Vasil Khalidov, Timoth{\'e}e Darcet, Th{\'e}o Moutakanni, Nikita
  Smetanin, Marc Szafraniec, et~al.
\newblock Automatic data curation for self-supervised learning: {A}
  clustering-based approach.
\newblock \emph{TMLR}, 2024.

\bibitem[Wang et~al.(2023)Wang, Braham, Xiong, Liu, Albrecht, and
  Zhu]{wang2023ssl4eo}
Yi Wang, Nassim Ait~Ali Braham, Zhitong Xiong, Chenying Liu, Conrad~M.
  Albrecht, and Xiao~Xiang Zhu.
\newblock {SSL4EO-S12: A} large-scale multimodal, multitemporal dataset for
  self-supervised learning in {E}arth observation.
\newblock \emph{IEEE Geoscience and Remote Sensing Magazine}, 2023.

\bibitem[Wei et~al.(2025)Wei, Dong, Lai, Yang, Luo, Zhang, Yang, Yuan, Zhao,
  Luo, and Fu]{wei2025rsprune}
Fan Wei, Runmin Dong, Yushan Lai, Yixiang Yang, Zhaoyang Luo, Jinxiao Zhang,
  Miao Yang, Shuai Yuan, Jiyao Zhao, Bin Luo, and Haohuan Fu.
\newblock {RS-Prune: T}raining-free data pruning at high ratios for efficient
  remote sensing diffusion foundation models.
\newblock \emph{arXiv:2512.23239}, 2025.

\bibitem[Wolters et~al.(2023)Wolters, Bastani, and
  Kembhavi]{wolters2023zooming}
Piper Wolters, Favyen Bastani, and Aniruddha Kembhavi.
\newblock Zooming out on zooming in: {A}dvancing super-resolution for remote
  sensing.
\newblock \emph{arXiv:2311.18082}, 2023.

\bibitem[Xie et~al.(2023)Xie, Santurkar, Ma, and Liang]{xie2023data}
Sang~Michael Xie, Shibani Santurkar, Tengyu Ma, and Percy~S. Liang.
\newblock Data selection for language models via importance resampling.
\newblock In \emph{NeurIPS}, 2023.

\bibitem[Xiong et~al.(2024)Xiong, Wang, Zhang, Stewart, Hanna, Borth,
  Papoutsis, Le~Saux, Camps-Valls, and Zhu]{xiong2024dofa}
Zhitong Xiong, Yi Wang, Fahong Zhang, Adam~J. Stewart, Jo{\"e}lle Hanna, Damian
  Borth, Ioannis Papoutsis, Bertrand Le~Saux, Gustau Camps-Valls, and
  Xiao~Xiang Zhu.
\newblock Neural plasticity-inspired multimodal foundation model for {E}arth
  observation.
\newblock \emph{arXiv:2403.15356}, 2024.

\bibitem[Yu et~al.(2020)Yu, Chan, You, Song, and Ma]{yu2020mcr2}
Yaodong Yu, Kwan Ho~Ryan Chan, Chong You, Chaobing Song, and Yi Ma.
\newblock Learning diverse and discriminative representations via the principle
  of maximal coding rate reduction.
\newblock In \emph{NeurIPS}, 2020.

\bibitem[Zanaga et~al.(2022)Zanaga, Van De~Kerchove, Daems, De~Keersmaecker,
  Brockmann, Kirches, Wevers, Cartus, Santoro, Fritz, Lesiv, Herold,
  Tsendbazar, Xu, Ramoino, and Arino]{zanaga2022worldcover}
Daniele Zanaga, Ruben Van De~Kerchove, Dirk Daems, Wanda De~Keersmaecker,
  Carsten Brockmann, Grit Kirches, Jan Wevers, Oliver Cartus, Maurizio Santoro,
  Steffen Fritz, Myroslava Lesiv, Martin Herold, Nandin-Erdene Tsendbazar,
  Panpan Xu, Fabrizio Ramoino, and Olivier Arino.
\newblock {ESA WorldCover} 10 m 2021 v200.
\newblock Zenodo, 2022.

\bibitem[Zauner(2010)]{zauner2010phash}
Christoph Zauner.
\newblock Implementation and benchmarking of perceptual image hash functions.
\newblock Master's thesis, Upper Austria University of Applied Sciences,
  Hagenberg, 2010.

\bibitem[Zhu et~al.(2026)Zhu, Xiong, Wang, Stewart, Heidler, Wang, Yuan,
  Dujardin, Xu, and Shi]{zhu2026foundations}
Xiao~Xiang Zhu, Zhitong Xiong, Yi Wang, Adam~J. Stewart, Konrad Heidler,
  Yuanyuan Wang, Zhenghang Yuan, Thomas Dujardin, Qingsong Xu, and Yilei Shi.
\newblock On the foundations of {E}arth foundation models.
\newblock \emph{Communications Earth \& Environment}, 2026.

\end{thebibliography}
}
\label{page:main_end}

\clearpage
\setcounter{topnumber}{99}\setcounter{dbltopnumber}{2}
\twocolumn[{\centering\Large\textbf{Appendix}\par\vspace{1.0em}}]
\appendix
\setcounter{figure}{0}
\setcounter{table}{0}
\renewcommand{\thefigure}{S\arabic{figure}}
\renewcommand{\thetable}{S\arabic{table}}
\renewcommand{\theHfigure}{S\arabic{figure}}
\renewcommand{\theHtable}{S\arabic{table}}
\newcommand{\suppcite}[1]{~\cite{#1}}
\newcommand{\mainref}[2]{\cref{#1}}
\newcommand{\mainrefnum}[2]{\ref{#1}}
\newcommand{\arxivversion}{1}
\providecommand{\mainref}[2]{#2}
\providecommand{\mainrefnum}[2]{#2}
\providecommand{\suppcite}[1]{}
\providecommand{\arxivversion}{0}
\setcounter{dbltopnumber}{3}
\ifnum\arxivversion=1 This appendix is organized as follows.\else This supplementary material is organized as follows.\fi{} \Cref{sec:supp:qualitative} illustrates each component of \ours{} with qualitative examples. \Cref{sec:supp:analysis} analyzes the selected subsets through their local and global diversity and their metadata distributions. \Cref{sec:supp:results} reports additional results: per-dataset results across SSL objectives at $10\%$ and $1\%$ budgets\ifnum\arxivversion=0, a $k$-NN evaluation, and cross-modality curation at a $1\%$ budget\else{} and a $k$-NN evaluation\fi. \Cref{sec:supp:s2naip} applies \ours{} to a second, much larger archive, S2-NAIP. \Cref{sec:supp:setup} details the implementation: the Major-TOM archive, the configuration of \ours{}, and the linear-probing and pretraining hyperparameters. \ifnum\arxivversion=1 Figures and tables numbered with an S belong to this appendix.\else Figures, tables, sections, and equations with plain numbers (\eg, \mainref{tab:geobench_curation}{Table~1}) refer to the main paper.\fi{}

\section{Qualitative examples}
\label{sec:supp:qualitative}
We illustrate the selection mechanism of \ours{} on Major-TOM observations. Unless otherwise stated, retained and discarded observations refer to the $1\%$ subset of \mainref{tab:geobench_curation}{Table~1}.

\begin{figure*}[t]
\centering
\newcommand{\qsp}[1]{\includegraphics[width=0.1175\linewidth]{images/qual/#1}}
\setlength{\tabcolsep}{0.5pt}\renewcommand{\arraystretch}{0.4}
\begin{tabular}{@{}r@{\hspace{2pt}}cccccccc@{}}
& \multicolumn{8}{c}{\small Percentile of local diversity $L_i$ in the archive} \\[3pt]
& \small $0.5$ & \small $10$ & \small $25$ & \small $50$ & \small $75$ & \small $90$ & \small $97$ & \small $99.5$ \\[2pt]
\rotatebox{90}{\makebox[0.1175\linewidth][c]{\scriptsize Image}} &
\qsp{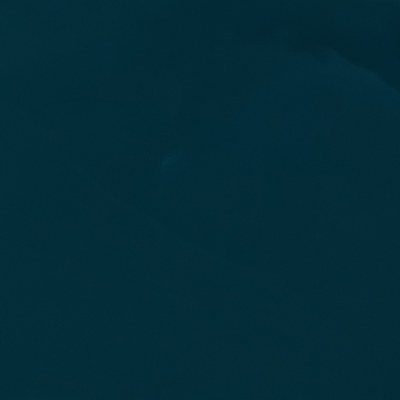} & \qsp{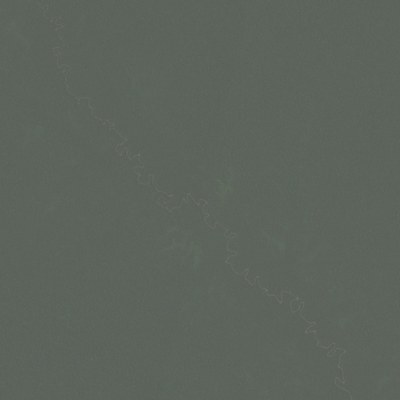} & \qsp{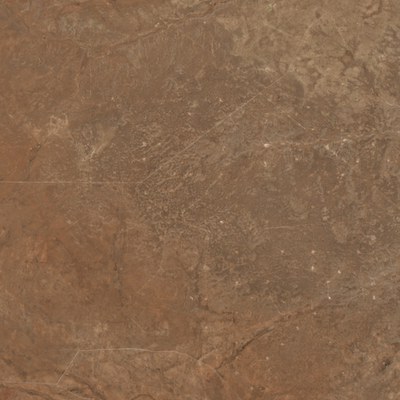} & \qsp{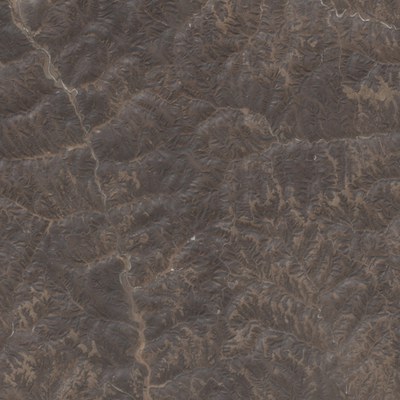} &
\qsp{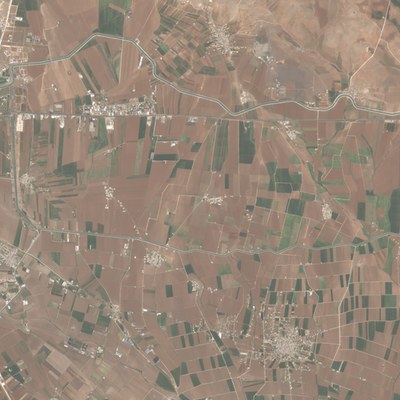} & \qsp{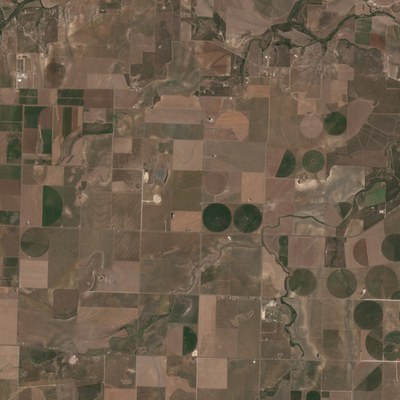} & \qsp{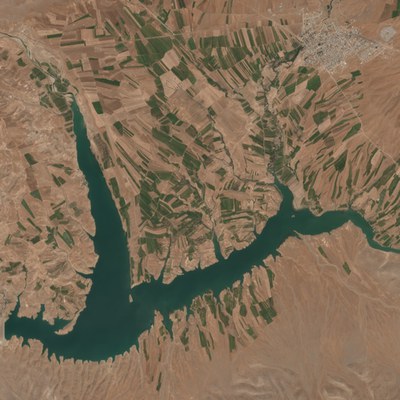} & \qsp{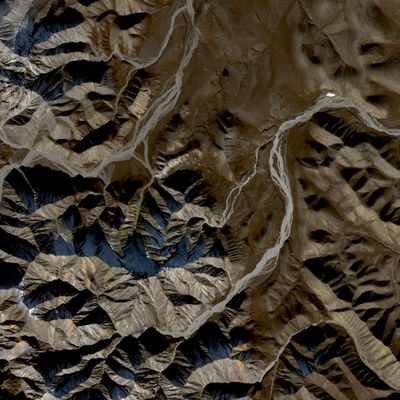} \\
\rotatebox{90}{\makebox[0.1175\linewidth][c]{\scriptsize Features}} &
\qsp{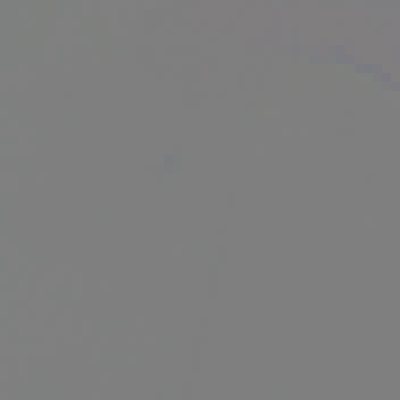} & \qsp{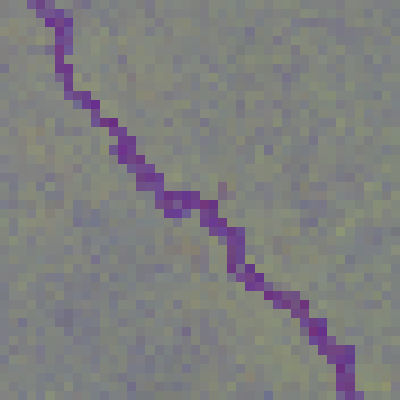} & \qsp{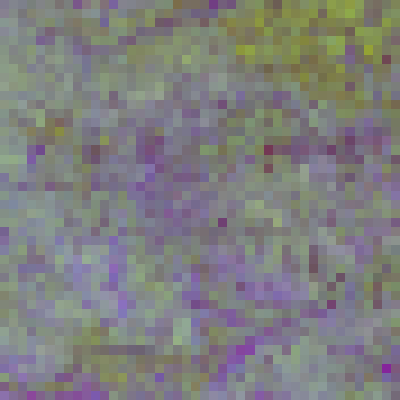} & \qsp{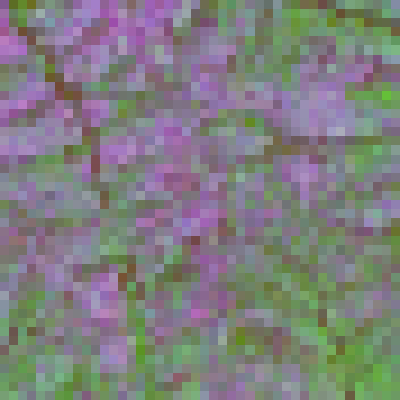} &
\qsp{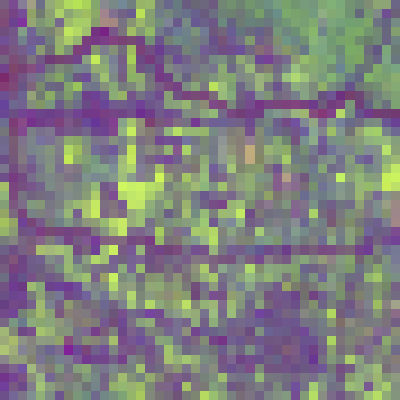} & \qsp{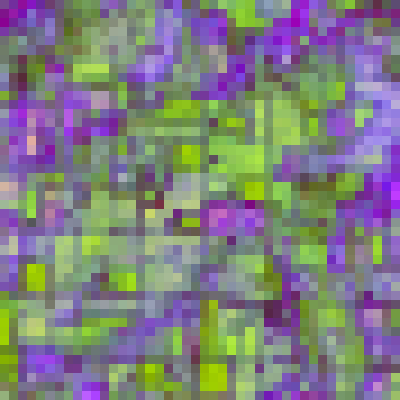} & \qsp{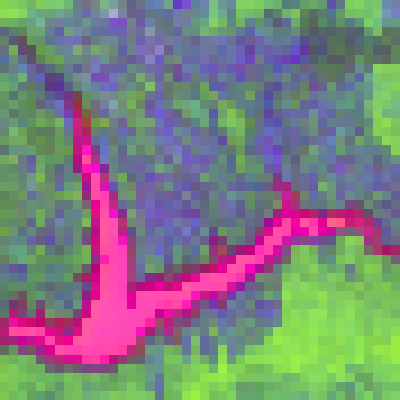} & \qsp{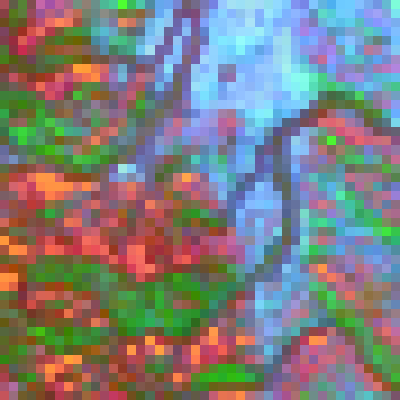} \\[4pt]
\rotatebox{90}{\makebox[0.1175\linewidth][c]{\scriptsize Image}} &
\qsp{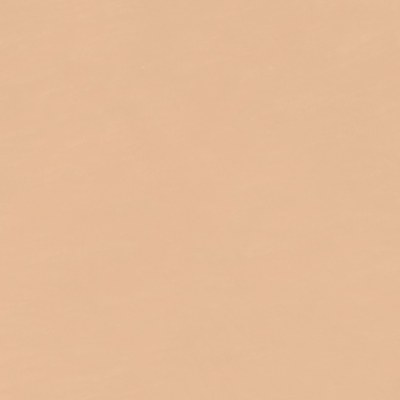} & \qsp{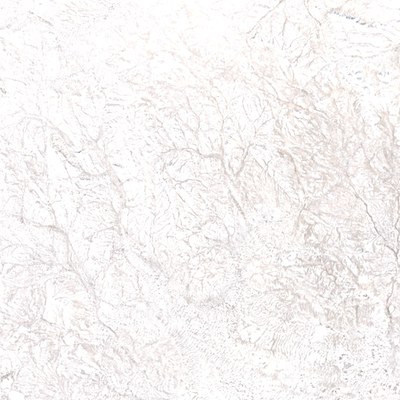} & \qsp{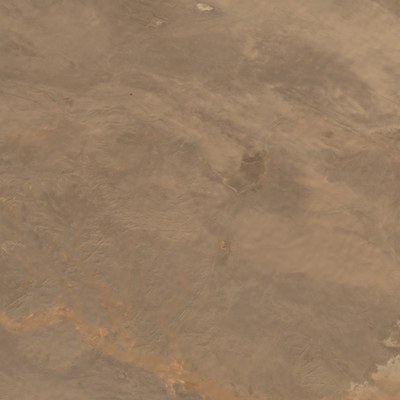} & \qsp{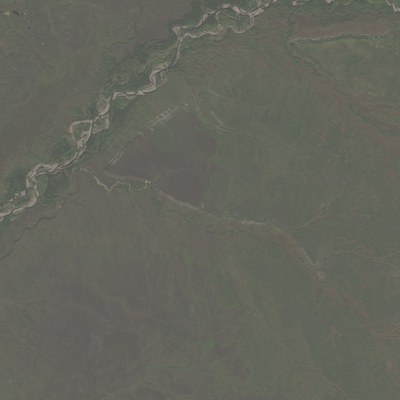} &
\qsp{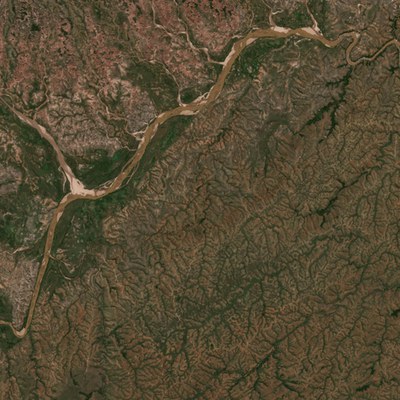} & \qsp{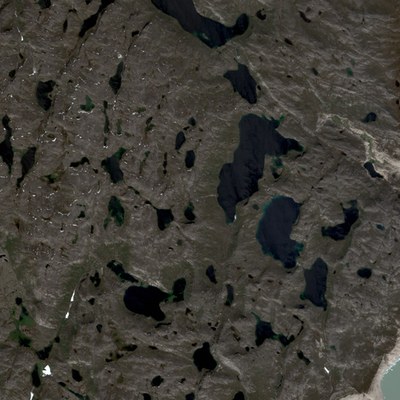} & \qsp{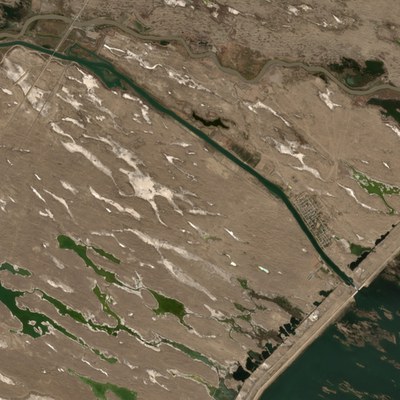} & \qsp{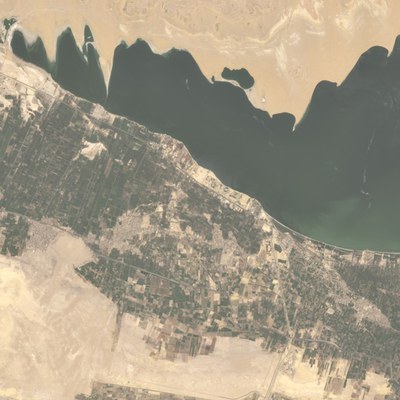} \\
\rotatebox{90}{\makebox[0.1175\linewidth][c]{\scriptsize Features}} &
\qsp{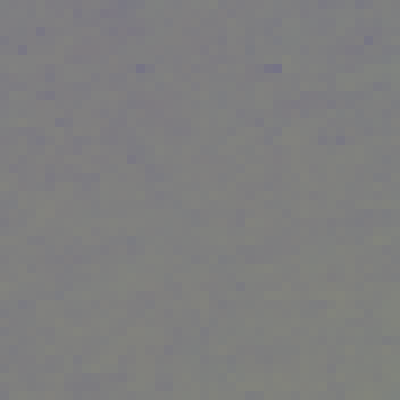} & \qsp{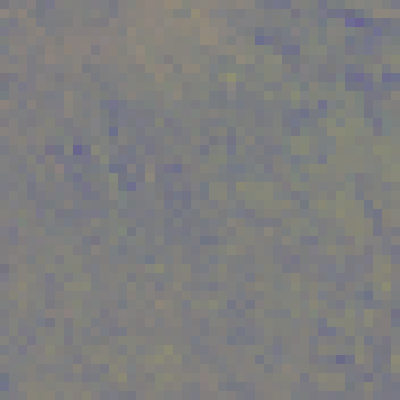} & \qsp{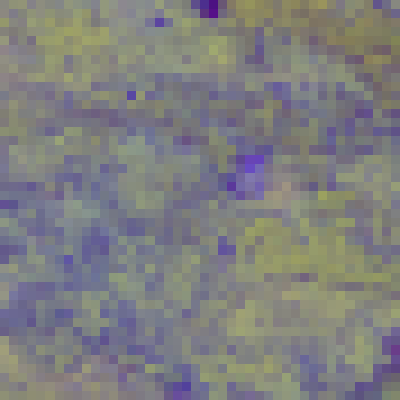} & \qsp{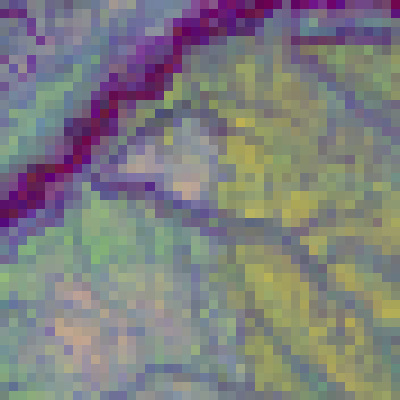} &
\qsp{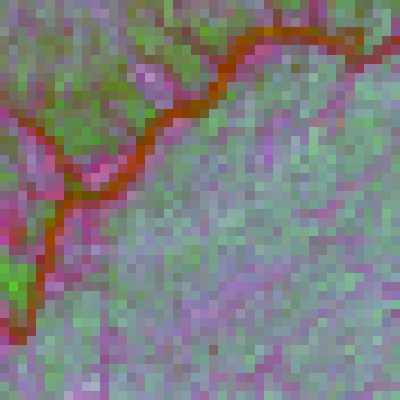} & \qsp{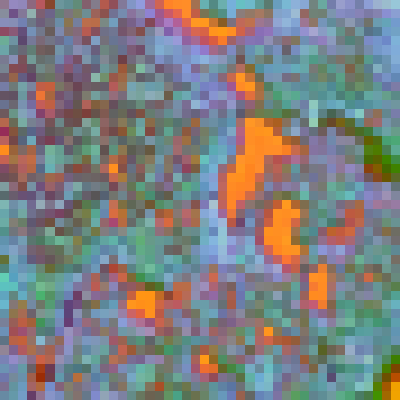} & \qsp{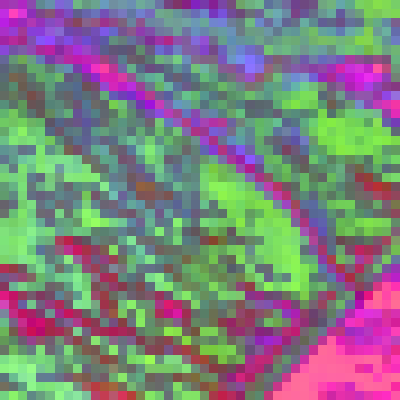} & \qsp{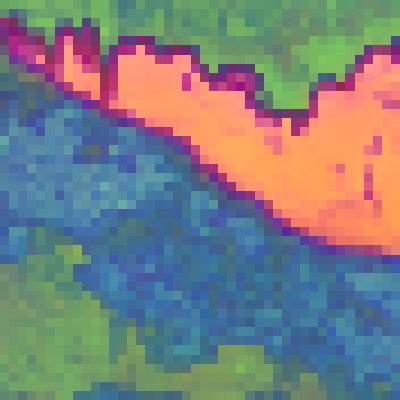} \\
\end{tabular}

\caption{\textbf{Local diversity across the archive.}
Two observations per percentile of $L_i$ (columns). Below each image, we visualize its UniverSat patch tokens, scaled by $1/\epsilon_d$ as in \mainref{eq:local}{Eq.~(3)}, centered on their within-image mean, and projected onto three principal components fitted jointly across all examples. Colors follow a shared scale, with gray corresponding to zero deviation from the mean.}
\label{fig:qual_spectrum}
\end{figure*}

\begin{figure*}[t]
\centering
\newcommand{\qal}[1]{\includegraphics[width=0.086\linewidth]{images/qual/#1}}
\newcommand{\qan}[2]{\rotatebox{90}{\makebox[0.086\linewidth][c]{\scriptsize\shortstack{$N_c=#1$\\[1pt]$B_c=#2$}}}}
\setlength{\tabcolsep}{0.5pt}\renewcommand{\arraystretch}{0.4}
\begin{tabular}{@{}r@{\hspace{2pt}}ccccc@{\hspace{8pt}}r@{\hspace{2pt}}ccccc@{}}
& \multicolumn{5}{c}{\small Low retention rate} & & \multicolumn{5}{c}{\small High retention rate} \\[3pt]
\qan{1{,}323}{1} & \qal{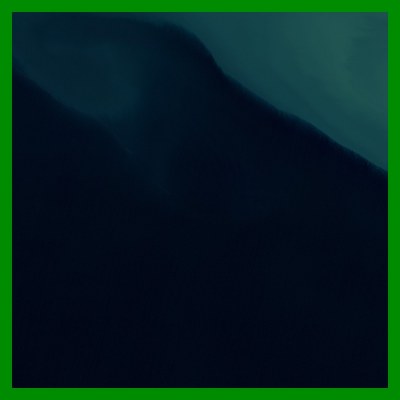} & \qal{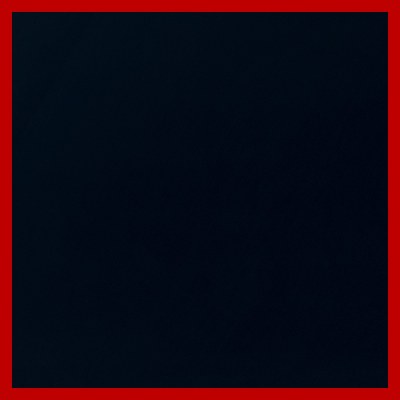} & \qal{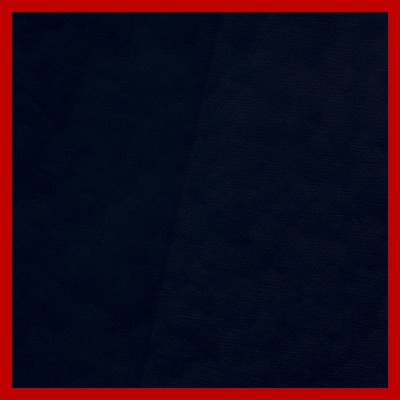} & \qal{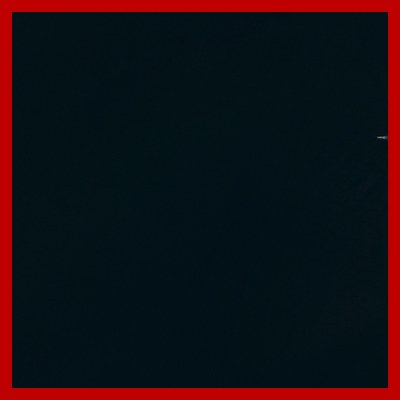} & \qal{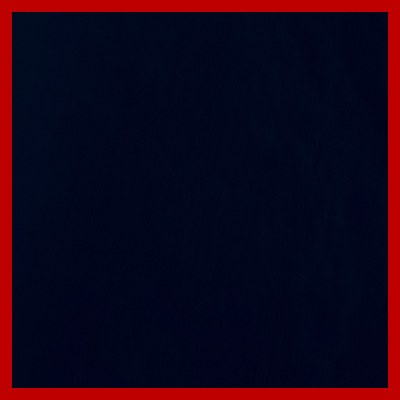} &
\qan{124}{69} & \qal{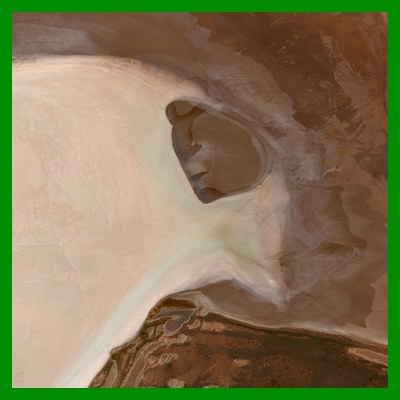} & \qal{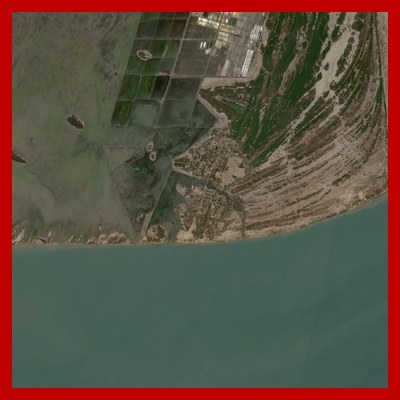} & \qal{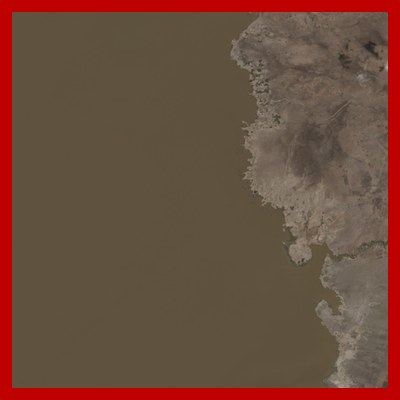} & \qal{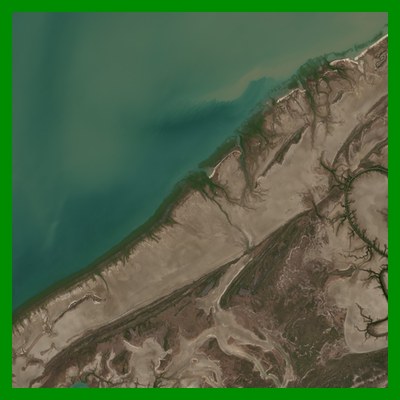} & \qal{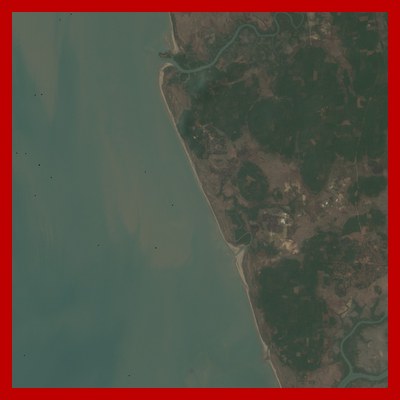} \\
\qan{679}{1} & \qal{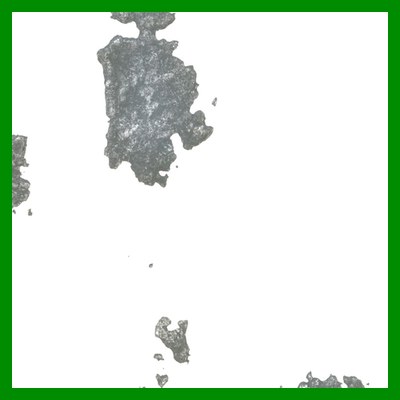} & \qal{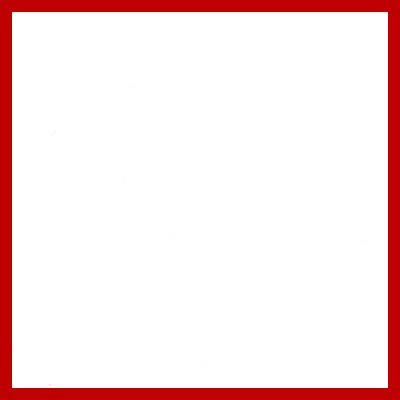} & \qal{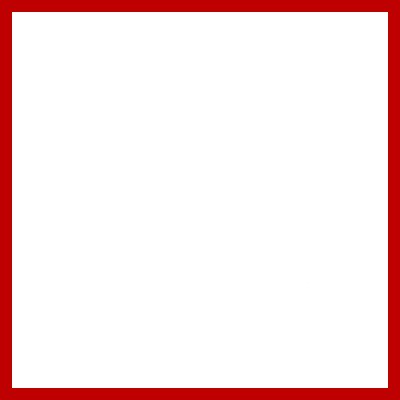} & \qal{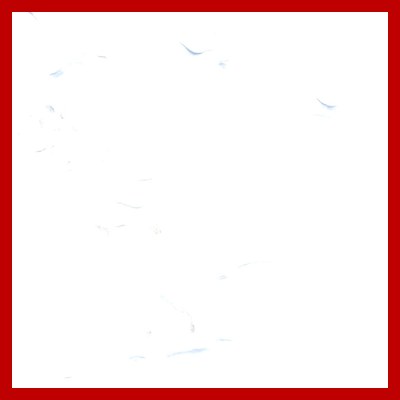} & \qal{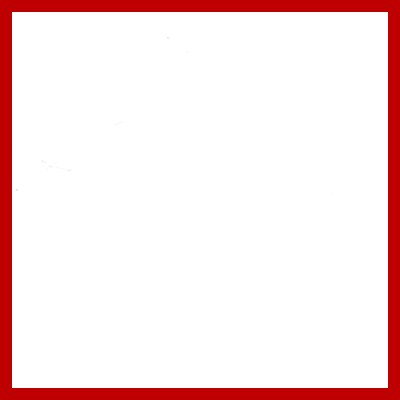} &
\qan{139}{25} & \qal{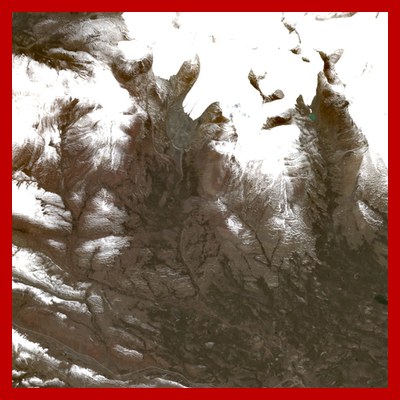} & \qal{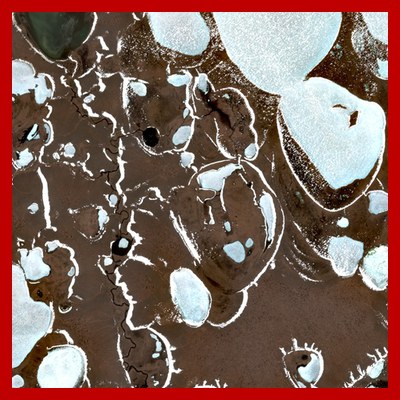} & \qal{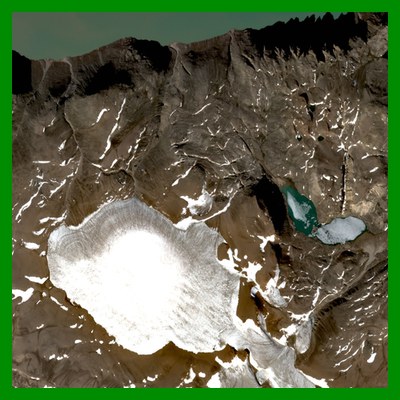} & \qal{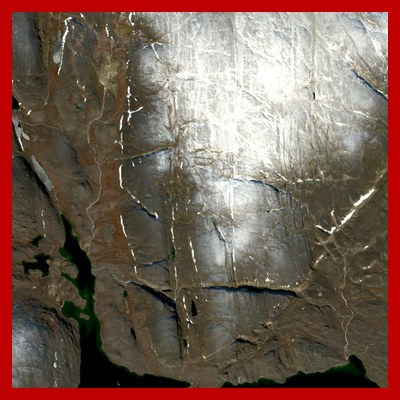} & \qal{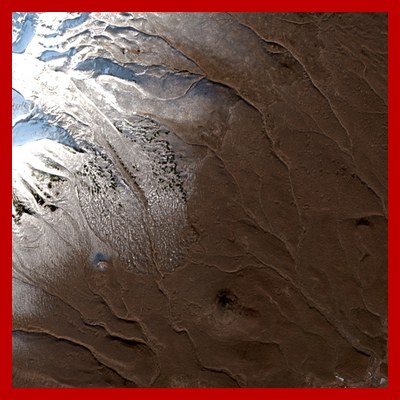} \\
\qan{569}{1} & \qal{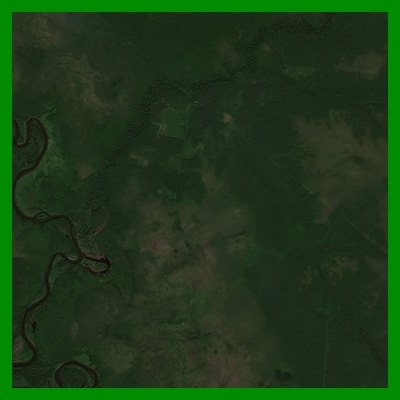} & \qal{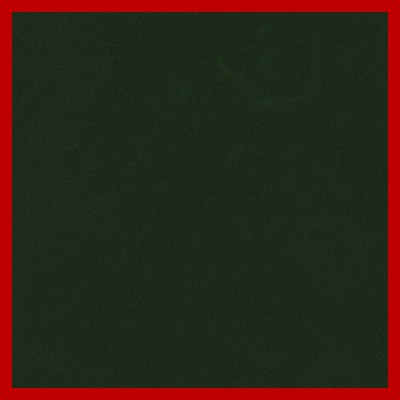} & \qal{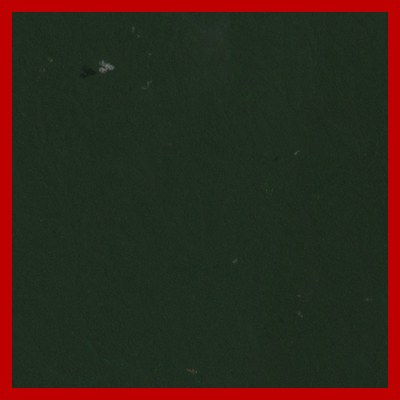} & \qal{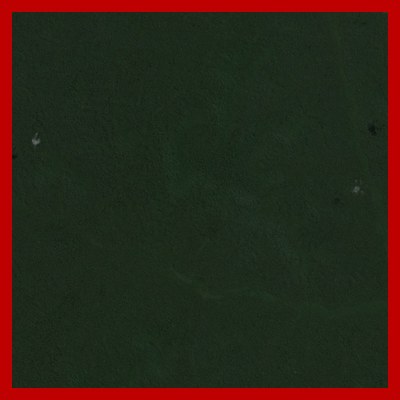} & \qal{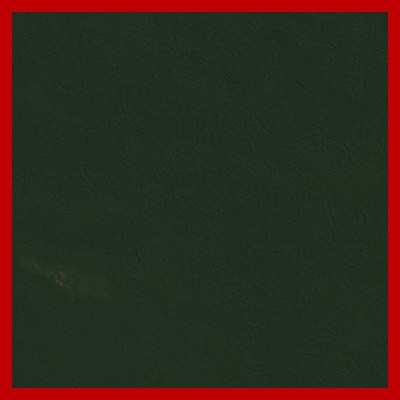} &
\qan{121}{23} & \qal{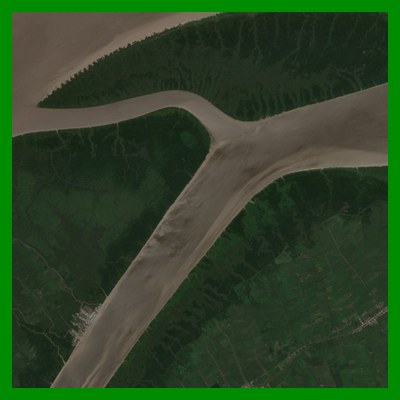} & \qal{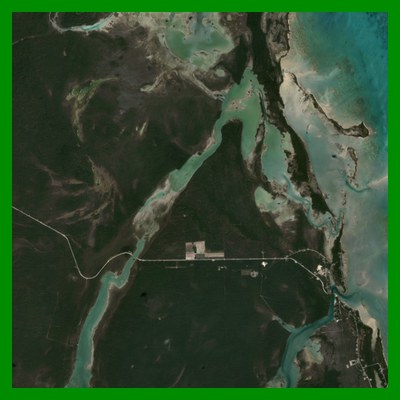} & \qal{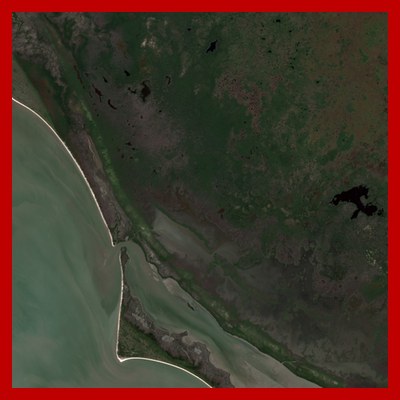} & \qal{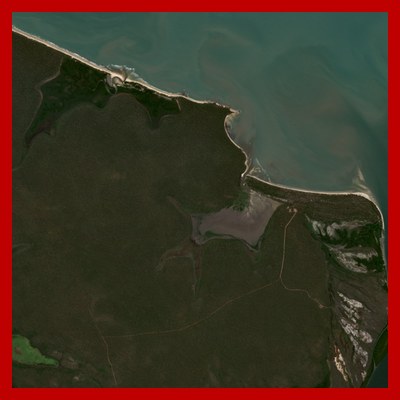} & \qal{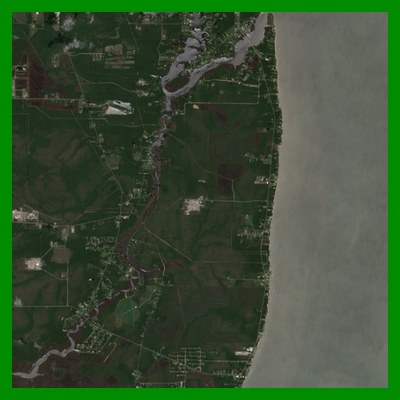} \\
\qan{466}{1} & \qal{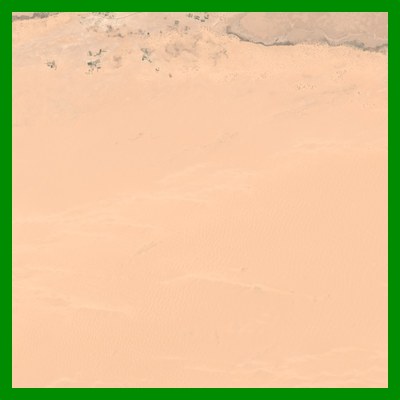} & \qal{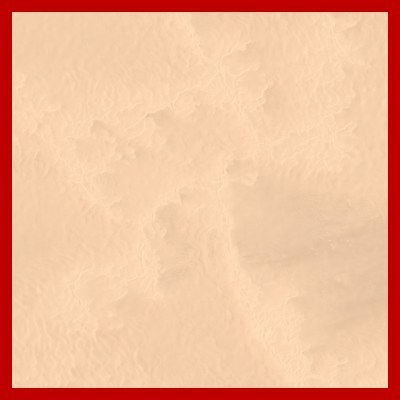} & \qal{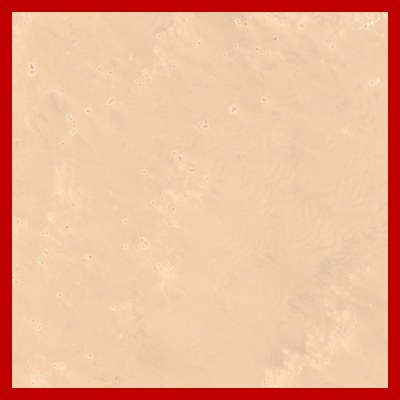} & \qal{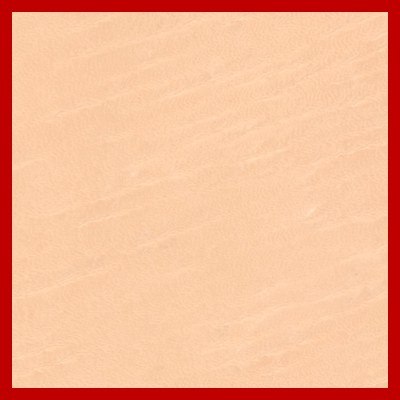} & \qal{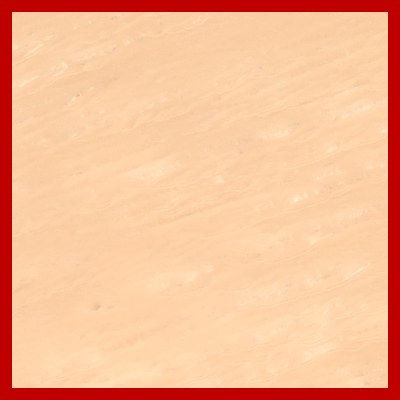} &
\qan{104}{28} & \qal{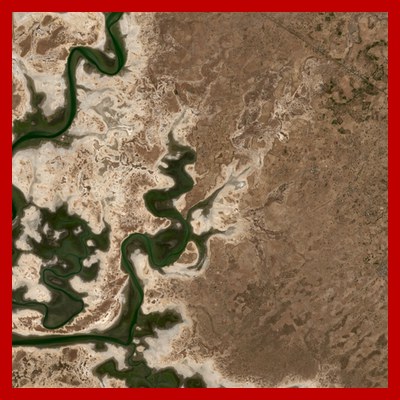} & \qal{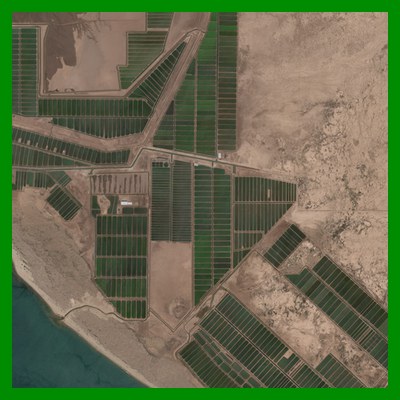} & \qal{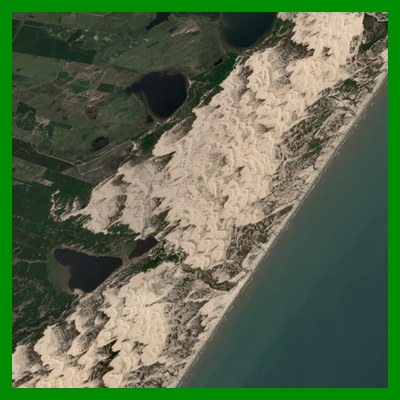} & \qal{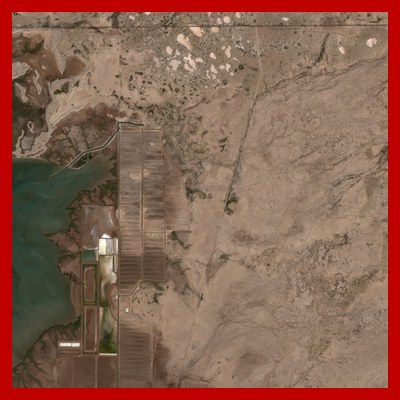} & \qal{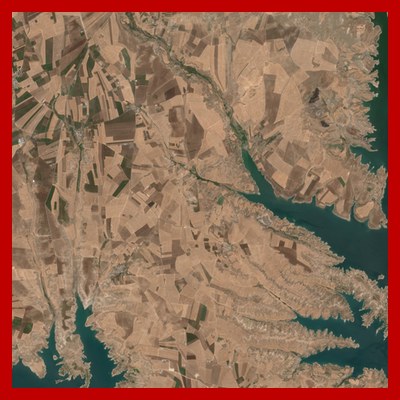} \\
\end{tabular}

\caption{\textbf{Budget allocation across semantic clusters.}
Examples of clusters with low (left) and high (right) retention rates at a $1\%$ curation budget, paired by semantic content. $N_c$ denotes the cluster size and $B_c$ its allocated budget. For each cluster, we show either the single retained observation and four randomly sampled discarded ones (left), or five randomly sampled observations (right). \textcolor{green!55!black}{Green} borders indicate retained observations and \textcolor{red!75!black}{red} borders discarded ones.}
\label{fig:qual_alloc}
\end{figure*}

\paragraph{Local diversity.}
\Cref{fig:qual_spectrum} illustrates how local diversity $L_i$ reflects the variability of patch tokens within an observation. Low-diversity observations depict homogeneous surfaces, such as open water, sand, or snow, and exhibit little feature variation. High-diversity observations contain richer spatial structures, such as agricultural parcels, river networks, coastlines, and mountainous terrain.

\begin{figure*}[t]
\centering
\newcommand{\qdp}[1]{\includegraphics[width=0.116\linewidth]{images/qual/#1}}
\newcommand{\qdl}[2]{\multicolumn{2}{c}{\scriptsize $#1$, #2}}
\setlength{\tabcolsep}{0.5pt}\renewcommand{\arraystretch}{0.4}
\begin{tabular}{@{}r@{\hspace{2pt}}cc@{\hspace{5pt}}cc@{\hspace{5pt}}cc@{\hspace{5pt}}cc@{}}
\rotatebox{90}{\makebox[0.116\linewidth][c]{\scriptsize Neighboring}} &
\qdp{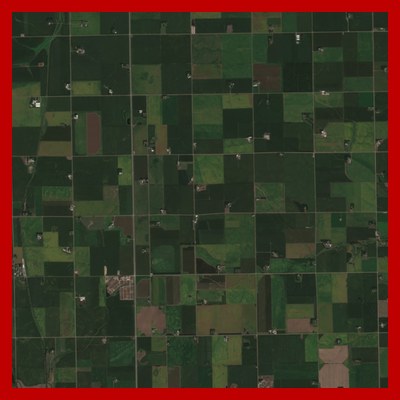} & \qdp{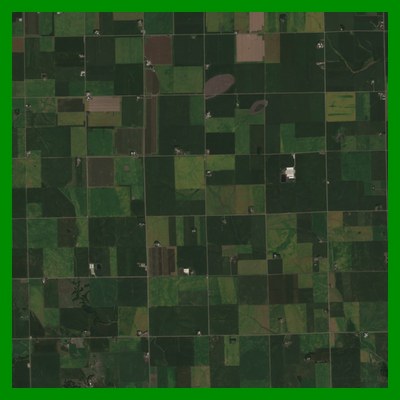} & \qdp{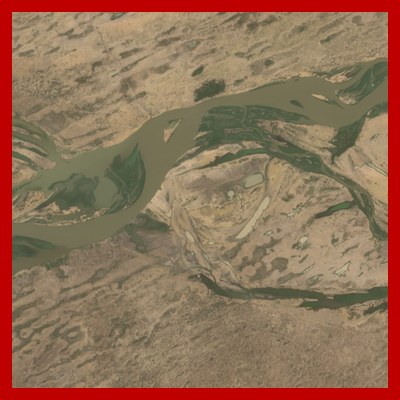} & \qdp{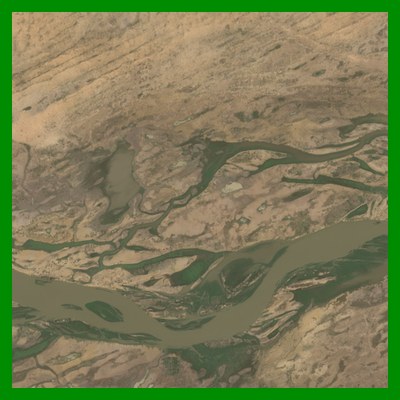} &
\qdp{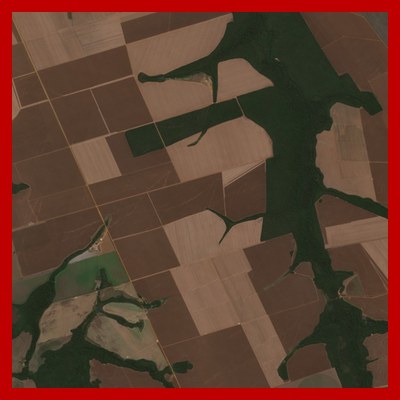} & \qdp{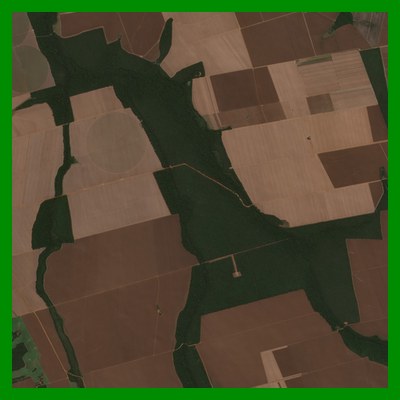} & \qdp{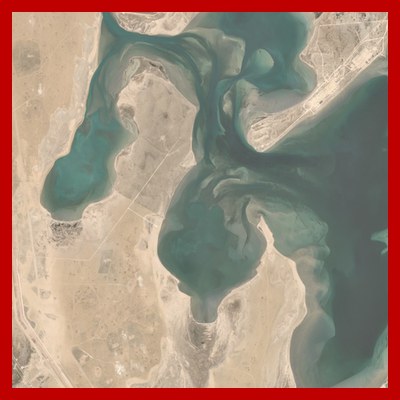} & \qdp{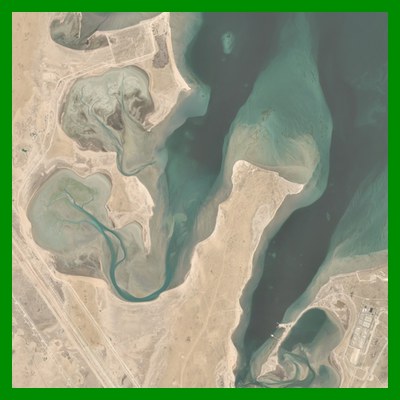} \\[2pt]
& \qdl{1.1}{10\,km} & \qdl{1.4}{10\,km} & \qdl{1.7}{13\,km} & \qdl{1.8}{12\,km} \\[6pt]
\rotatebox{90}{\makebox[0.116\linewidth][c]{\scriptsize Distant}} &
\qdp{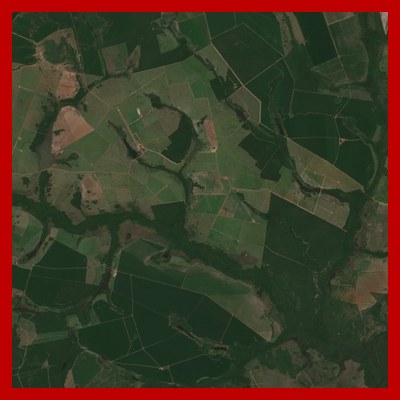} & \qdp{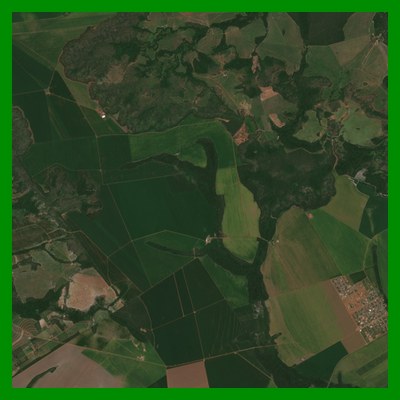} & \qdp{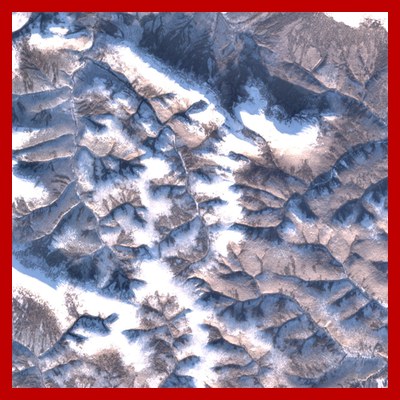} & \qdp{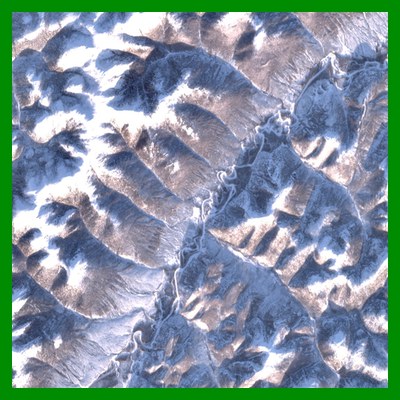} &
\qdp{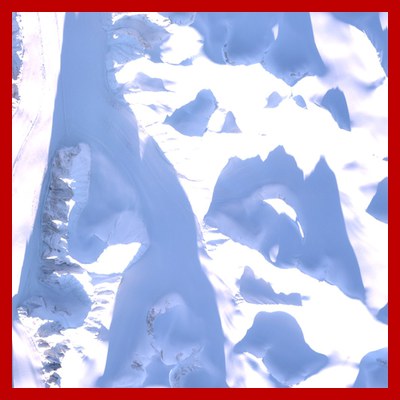} & \qdp{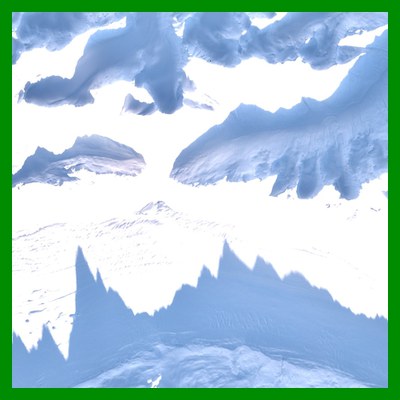} & \qdp{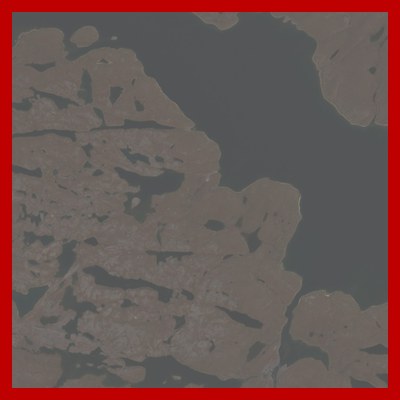} & \qdp{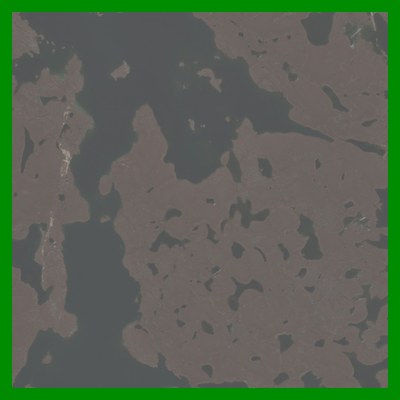} \\[2pt]
& \qdl{1.9}{580\,km} & \qdl{1.9}{335\,km} & \qdl{1.9}{2{,}916\,km} & \qdl{2.0}{1{,}918\,km} \\
\end{tabular}

\caption{\textbf{Near-duplicates removed by global diversity.} Observations retained by ranking on $L_i$ alone (\emph{Local ranking} in \mainref{tab:ablation}{Table~5}) but discarded by \ours{} (\textcolor{red!75!black}{red}), next to their most similar observation retained by \ours{} (\textcolor{green!55!black}{green}). Both observations of each pair belong to the same cluster and are retained by local ranking; only the global diversity $G_i$ leads \ours{} to keep a single one. Under each pair, the cosine distance between their global embeddings ($\times10^{-3}$, as in \cref{tab:two_div}) and the geographic distance between them. Top: pairs less than $30$\,km apart (\emph{Neighboring}); bottom: more than $300$\,km apart (\emph{Distant}).}
\label{fig:qual_dups}
\end{figure*}

\paragraph{Budget allocation.}
\Cref{fig:qual_alloc} illustrates how the budget allocation of \mainref{eq:cluster_budget}{Eq.~(8)} adapts to semantic redundancy. Clusters of near-identical observations, such as open water, ice sheets, closed rainforest, or sand seas, receive a single slot for several hundred observations. Clusters with similar content but more varied spatial structures retain a much larger fraction of their observations. Overall, $10\%$ of the clusters receive over a third of the total budget. Although the allocation relies solely on global embeddings, the retention rate $B_c/N_c$ of a cluster correlates with the median local diversity of its observations (Spearman $\rho=0.57$ across clusters), as locally uniform observations also tend to be similar to one another. Over a third of the clusters receive a single slot, allocated to their most locally diverse observation.

\begin{figure*}[t]
\centering
\newcommand{\qcl}[1]{\includegraphics[width=0.13\linewidth]{images/qual/#1}}
\setlength{\tabcolsep}{0.5pt}\renewcommand{\arraystretch}{0.4}
\begin{tabular}{@{}r@{\hspace{2pt}}cccc@{\hspace{6pt}}ccc@{}}
& \multicolumn{4}{c}{\small Retained by \ours{}} & \multicolumn{3}{c}{\small Discarded} \\[3pt]
\rotatebox{90}{\makebox[0.13\linewidth][c]{\scriptsize Ice cap (EF)}} &
\qcl{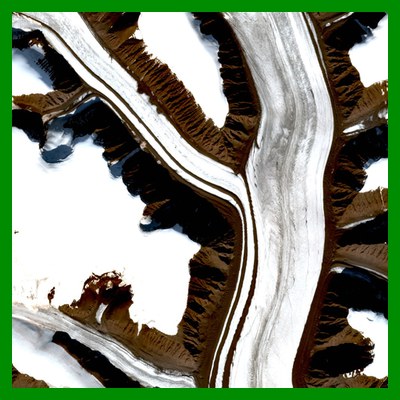} & \qcl{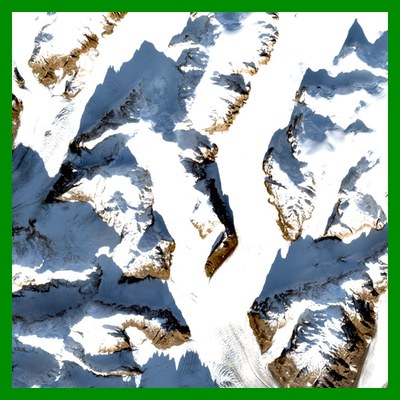} & \qcl{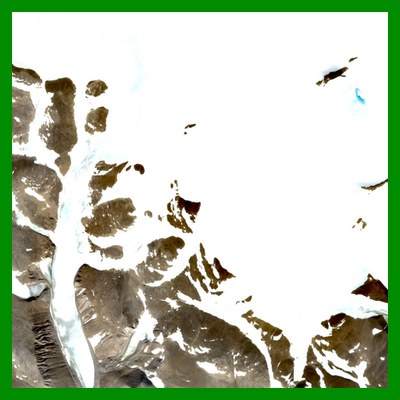} & \qcl{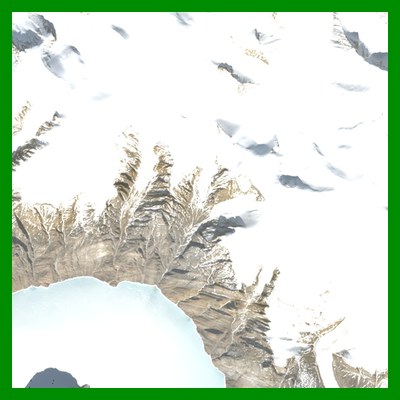} & \qcl{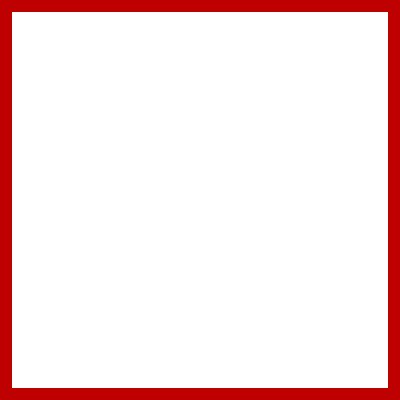} & \qcl{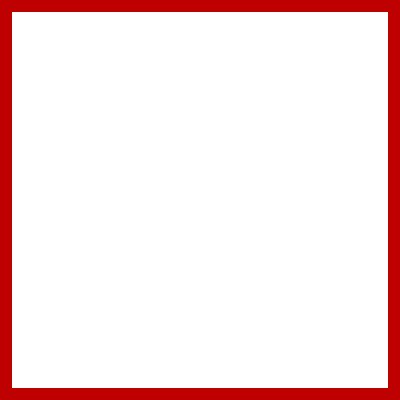} & \qcl{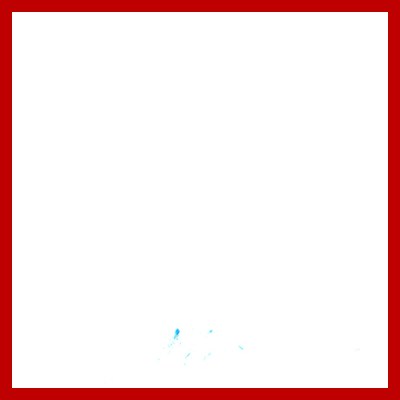} \\
\rotatebox{90}{\makebox[0.13\linewidth][c]{\scriptsize Rainforest (Af)}} &
\qcl{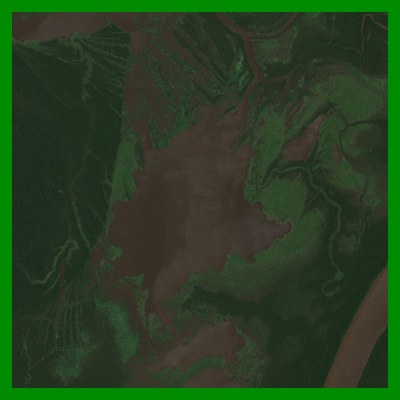} & \qcl{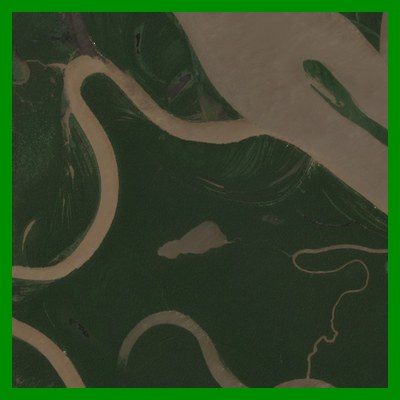} & \qcl{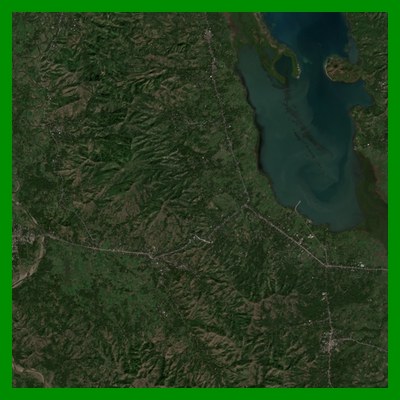} & \qcl{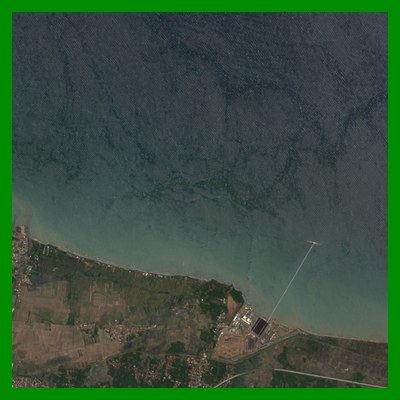} & \qcl{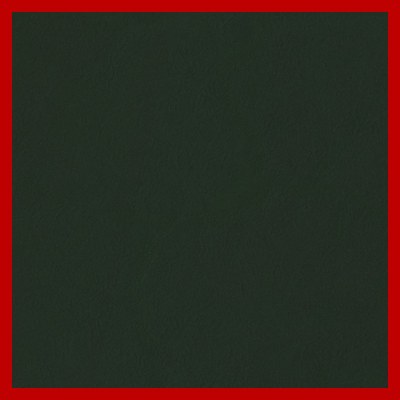} & \qcl{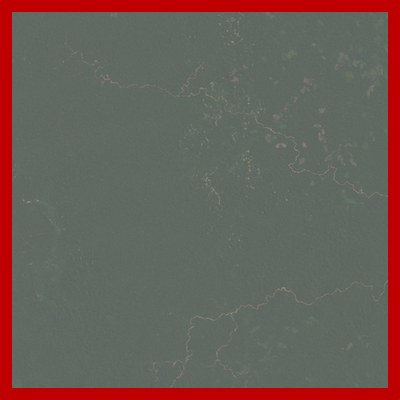} & \qcl{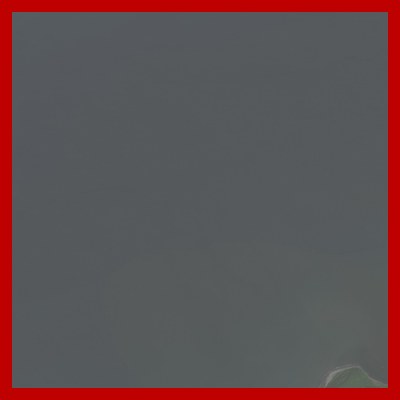} \\
\rotatebox{90}{\makebox[0.13\linewidth][c]{\scriptsize Hot desert (BWh)}} &
\qcl{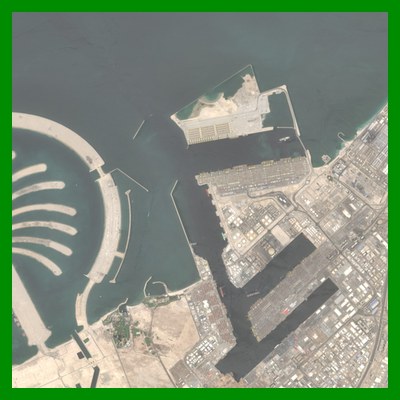} & \qcl{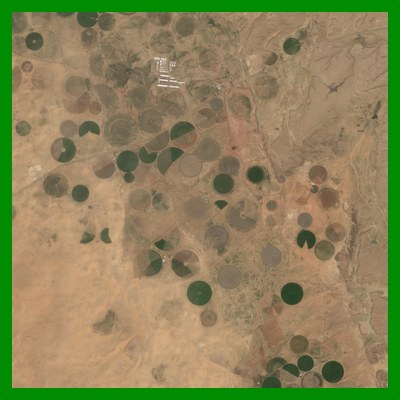} & \qcl{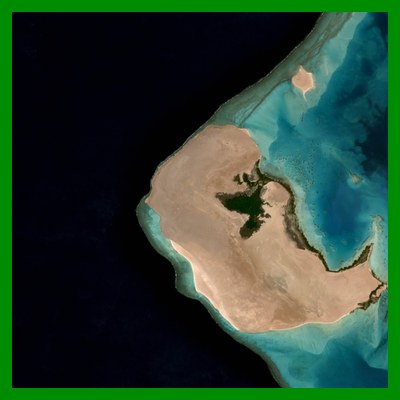} & \qcl{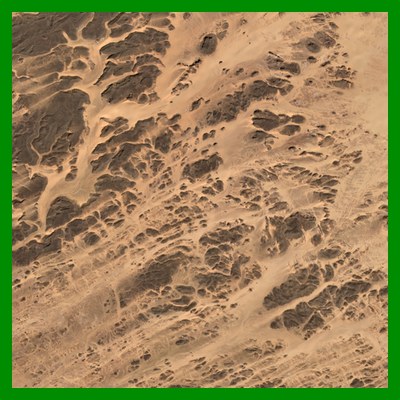} & \qcl{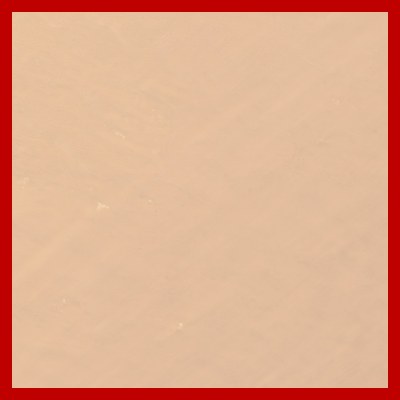} & \qcl{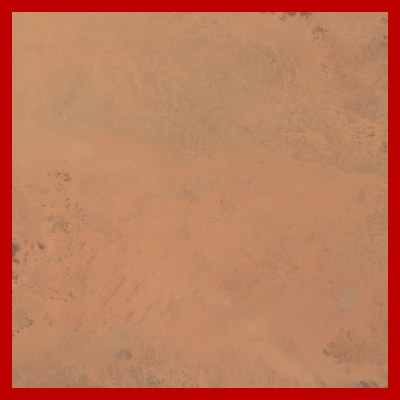} & \qcl{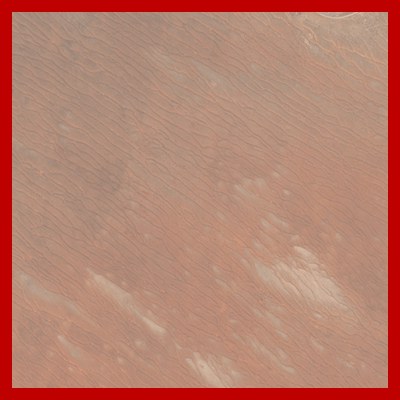} \\
\end{tabular}

\caption{\textbf{Selection within homogeneous climates.}
Examples from three K\"oppen--Geiger climates: ice cap (EF), tropical rainforest (Af), and hot desert (BWh), retained by \ours{} (left, \textcolor{green!55!black}{green}) or discarded (right, \textcolor{red!75!black}{red}).}
\label{fig:qual_climates}
\end{figure*}

\begin{figure*}[t]
\centering
\newcommand{\qrd}[1]{\includegraphics[width=0.1225\linewidth]{images/qual/#1}}
\setlength{\tabcolsep}{0.5pt}\renewcommand{\arraystretch}{0.4}
\begin{tabular}{@{}cccccccc@{}}
\qrd{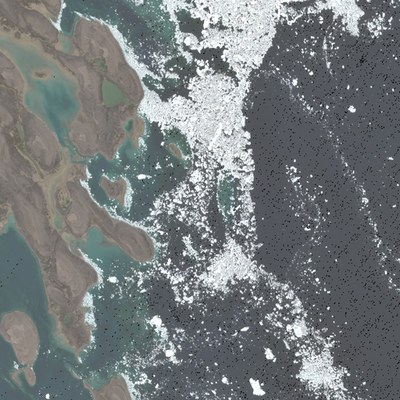} & \qrd{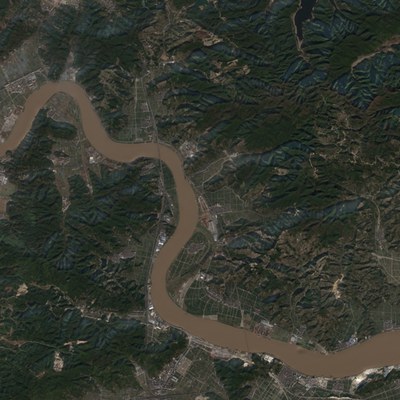} & \qrd{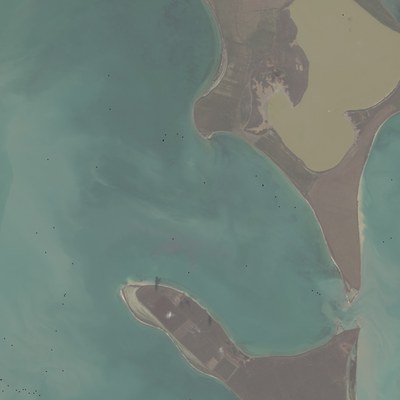} & \qrd{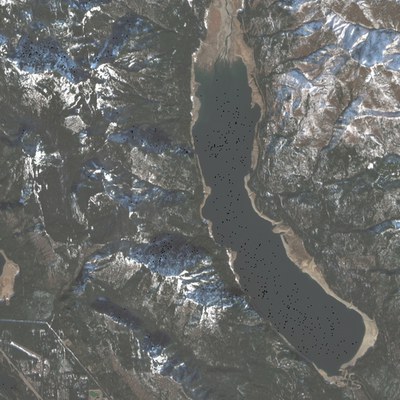} & \qrd{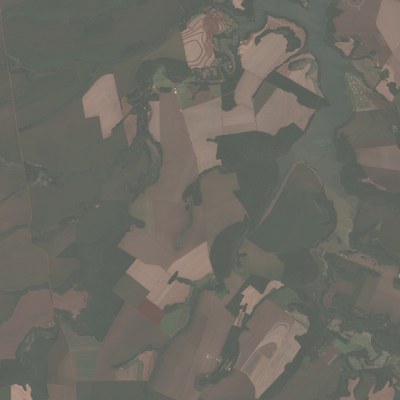} & \qrd{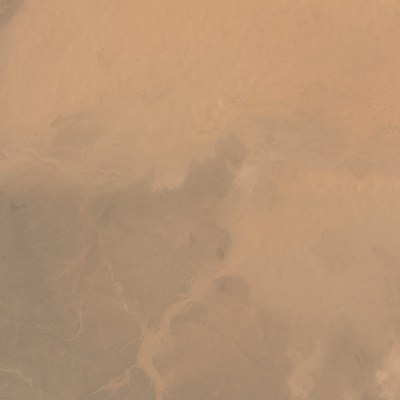} & \qrd{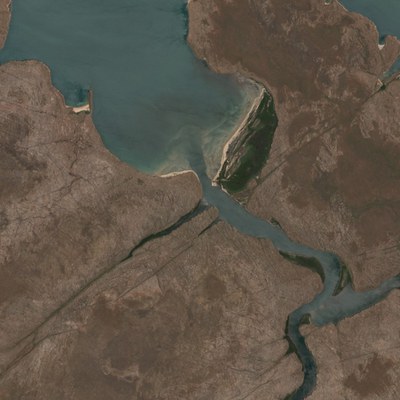} & \qrd{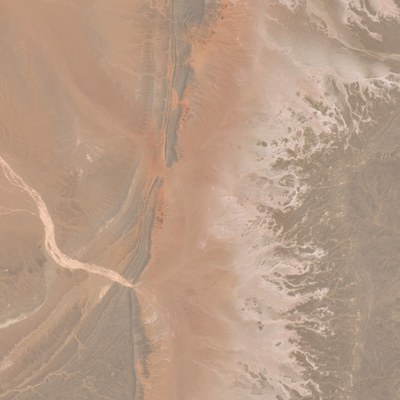} \\
\qrd{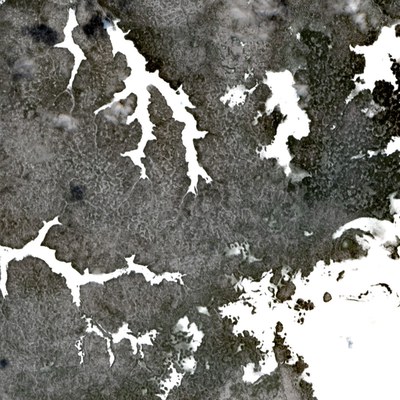} & \qrd{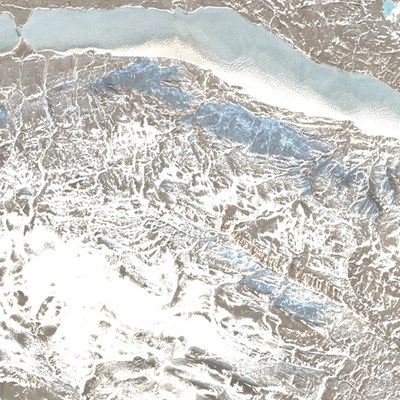} & \qrd{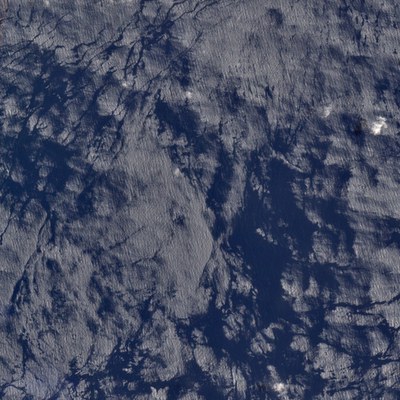} & \qrd{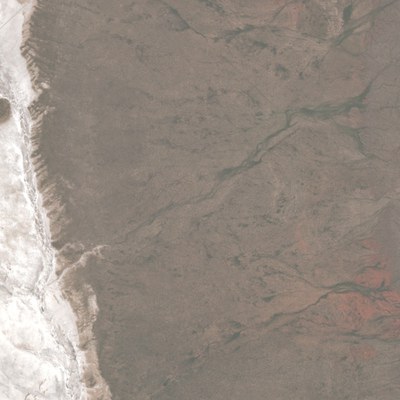} & \qrd{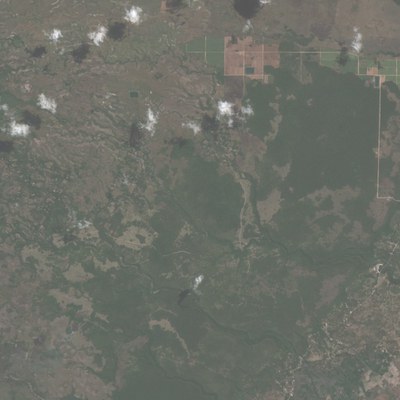} & \qrd{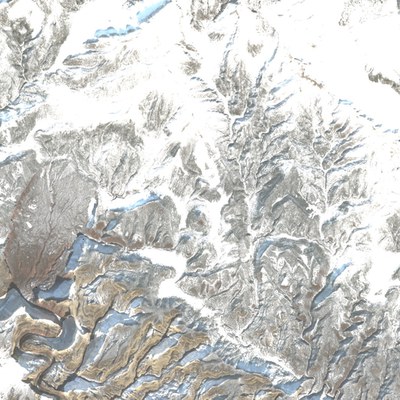} & \qrd{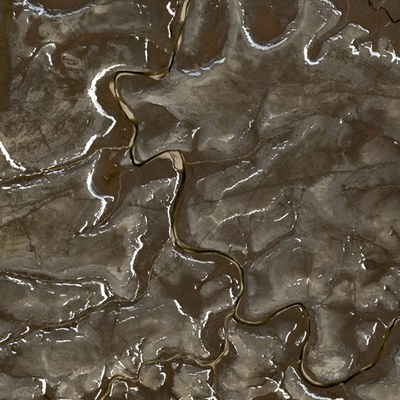} & \qrd{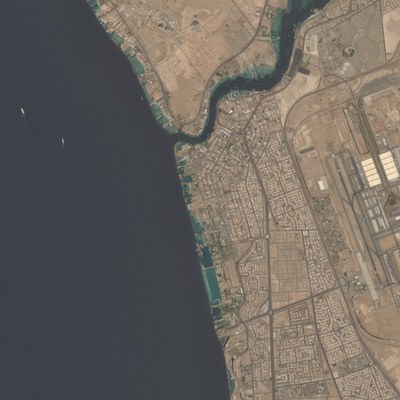} \\
\qrd{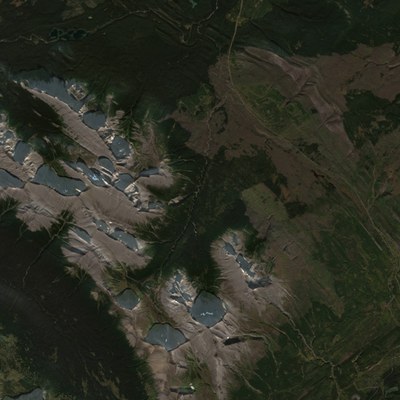} & \qrd{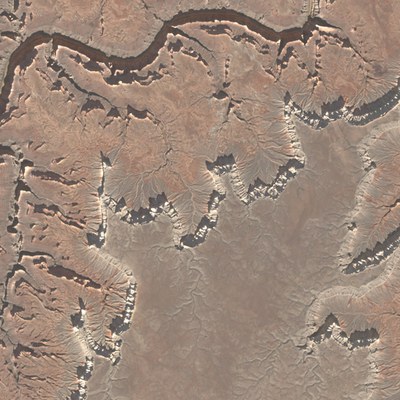} & \qrd{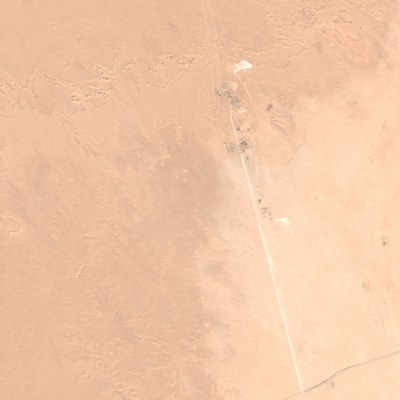} & \qrd{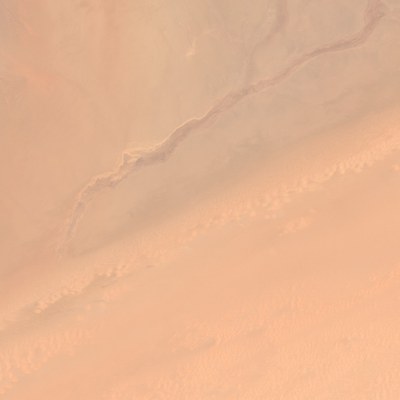} & \qrd{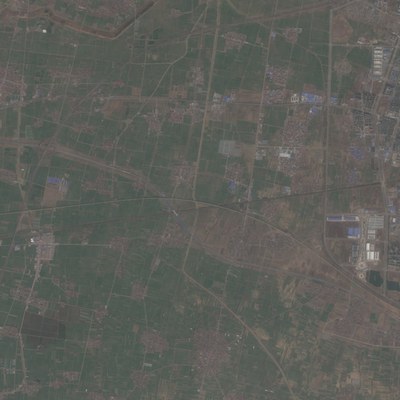} & \qrd{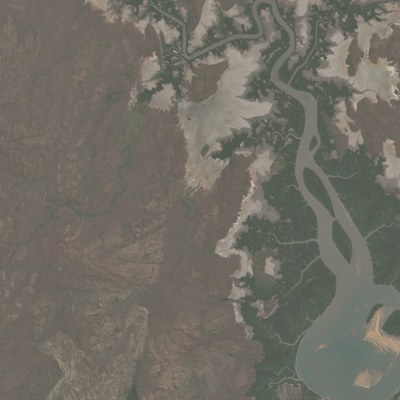} & \qrd{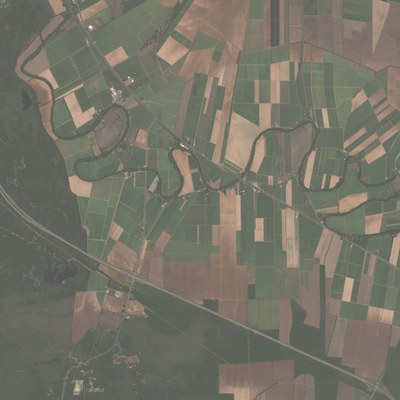} & \qrd{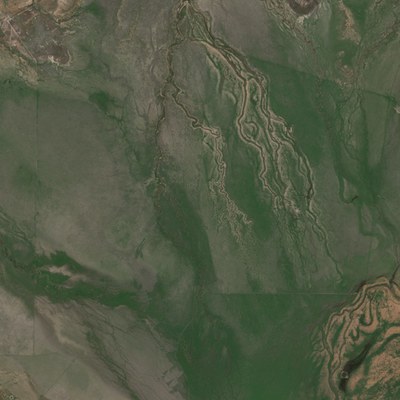} \\
\end{tabular}

\caption{\textbf{Random sample of the curated subset.}
Twenty-four observations drawn uniformly at random from the $1\%$ \ours{} subset, without further selection.}
\label{fig:qual_random}
\end{figure*}

\begin{figure*}[t]
\centering
\newcommand{\qfl}[1]{\includegraphics[width=0.1175\linewidth]{images/qual/#1}}
\setlength{\tabcolsep}{0.5pt}\renewcommand{\arraystretch}{0.4}
\begin{tabular}{@{}cc@{\hspace{5pt}}cc@{\hspace{5pt}}cc@{\hspace{5pt}}cc@{}}
\qfl{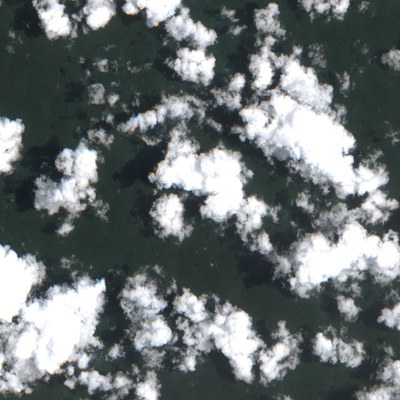} & \qfl{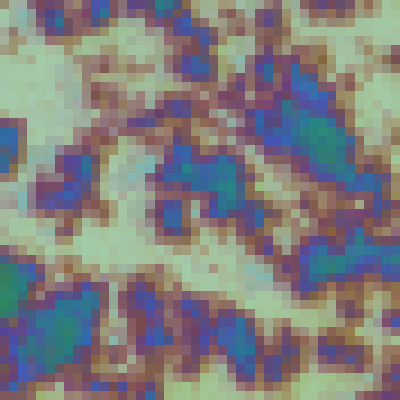} & \qfl{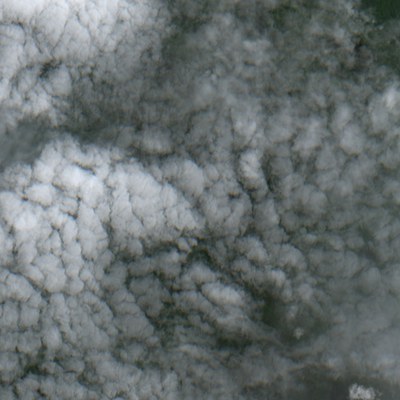} & \qfl{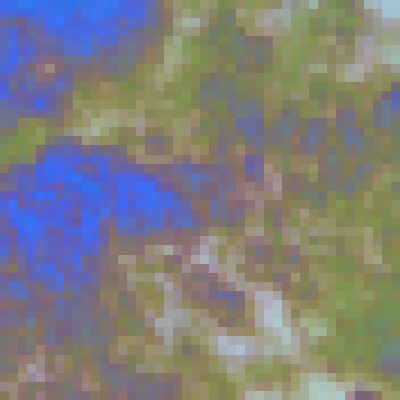} &
\qfl{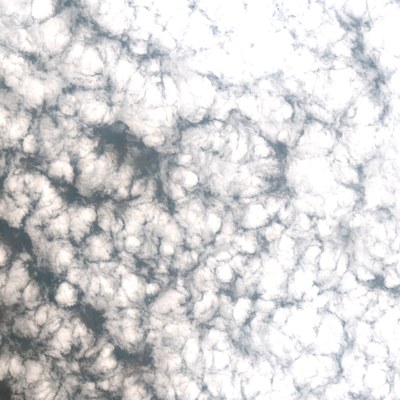} & \qfl{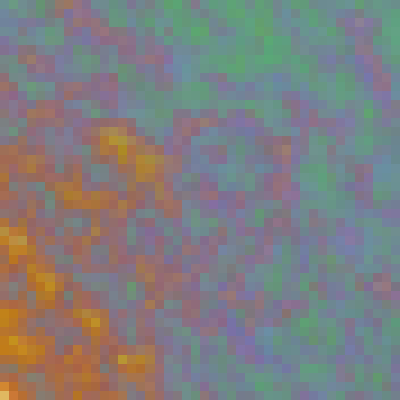} & \qfl{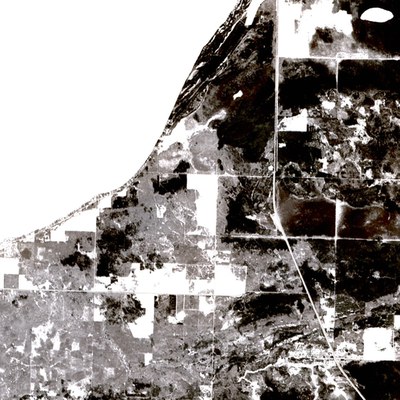} & \qfl{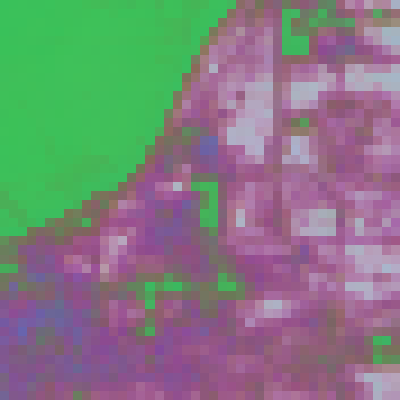} \\[2pt]
\multicolumn{2}{c}{\small (a)} & \multicolumn{2}{c}{\small (b)} & \multicolumn{2}{c}{\small (c)} & \multicolumn{2}{c}{\small (d)} \\
\end{tabular}

\caption{\textbf{Failure cases.}
Observations retained by \ours{} whose high local diversity is largely attributable to clouds (a--c) or, more ambiguously, to snow (d), with their patch-token visualizations. Principal components are fitted jointly across the four examples.}
\label{fig:qual_failures}
\end{figure*}

\paragraph{Near-duplicates.} Ranking by $L_i$ alone within the same cluster budgets (\emph{Local ranking} in \mainref{tab:ablation}{Table~5}) differs from \ours{} on over a quarter of its observations, as $S_i$ trades local diversity against distinctiveness within each cluster. Some of these differences are near-duplicates: \cref{fig:qual_dups} pairs observations retained by local ranking but discarded by \ours{} with their most similar observation that \ours{} retains. Both observations of each pair belong to the same cluster, are retained by local ranking, and are nearly identical in content: only the global diversity $G_i$ leads \ours{} to keep a single one. Such pairs are rarely neighboring cells: their median distance is $1{,}500$\,km, and fewer than $10\%$ lie within $50$\,km. This redundancy is therefore semantic rather than spatial, and spatial thinning would not remove it. On Major-TOM, which observes each location once, removing such pairs has little measurable effect downstream: \emph{Local ranking} and \ours{} perform on par (\mainref{tab:ablation}{Table~5}), although the subset of \ours{} is less redundant (\cref{tab:two_div}).

\paragraph{Homogeneous climates.}
Even in climates dominated by homogeneous observations, \ours{} identifies locally diverse scenes (\cref{fig:qual_climates}): glaciated relief and ice-cap margins, rivers and coastlines in tropical rainforests, and irrigated fields and rocky relief in hot deserts. Rather than excluding these climates altogether, \ours{} preferentially retains their more structured observations.

\paragraph{Random draw.}
\Cref{fig:qual_random} shows a random sample of the curated subset. Most observations exhibit varied spatial structures, although some homogeneous scenes remain, reflecting the preservation of semantic coverage: at this budget, every cluster retains at least one observation.

\paragraph{Failure cases.}
Local diversity can also arise from atmospheric phenomena rather than from the underlying scene: residual clouds artificially increase $L_i$, favoring observations in which the ground is partially obscured (\cref{fig:qual_failures}a--c). More stringent cloud filtering or cloud-aware representations could mitigate this limitation. Snow cover also increases $L_i$ (\cref{fig:qual_failures}d), but whether it is a desirable source of diversity depends on the downstream application.

\section{Analysis of the selected subsets}
\label{sec:supp:analysis}
We profile the diversity of the $1\%$ subsets (\cref{sec:supp:two_div}) and their metadata distributions (\cref{sec:supp:correlation}).
\begin{table*}[t]
\centering
\begin{minipage}[c]{0.51\textwidth}
\caption{\textbf{Local and global diversity of $1\%$ subsets.}
We compare a random draw (not the geographic \emph{Uniform} baseline of \mainref{tab:geobench_curation}{Table~1}), SemDeDup, the observations of highest local diversity $L_i$ regardless of clusters (\emph{Local div.\ only}), the \emph{Local ranking} of \mainref{tab:ablation}{Table~5}, which ranks by $L_i$ within clusters, and \ours{}, using UniverSat embeddings. Local diversity is characterized by the median percentile of $L_i$ relative to the full archive and the proportion of observations in its less diverse half. Global diversity is characterized by the median cosine distance ($\times 10^{-3}$) to the nearest retained neighbor within the same cluster and the effective number of $100$\,km regions (exponential of the entropy of their distribution). Arrows indicate the direction of greater diversity.}
\label{tab:two_div}
\end{minipage}
\hfill
\begin{minipage}[c]{0.46\textwidth}
\centering
\small
\setlength{\tabcolsep}{3.5pt}
\begin{tabular}{lcccc}
\toprule
& \multicolumn{2}{c}{Local} & \multicolumn{2}{c}{Global} \\
\cmidrule(lr){2-3}\cmidrule(lr){4-5}
Selection & $L_i$ pctl.\,$\uparrow$ & low-$L_i$\,$\downarrow$ & NN dist.\,$\uparrow$ & regions\,$\uparrow$ \\
\midrule
Random            & 50 & 50\% & \phantom{0}5.0 & 10.2k \\
SemDeDup          & 81 & 24\% & 24.9 & \phantom{0}7.3k \\
Local div.\ only  & 99 & \phantom{0}0\% & \phantom{0}6.2 & \phantom{0}4.0k \\
Local ranking     & 94 & \phantom{0}5\% & 10.1 & \phantom{0}6.5k \\
\ours{}          & 93 & \phantom{0}5\% & 15.1 & \phantom{0}6.7k \\
\bottomrule
\end{tabular}

\end{minipage}
\vspace{1.5\baselineskip}
\end{table*}
\begin{figure*}[t]
\centering
\input{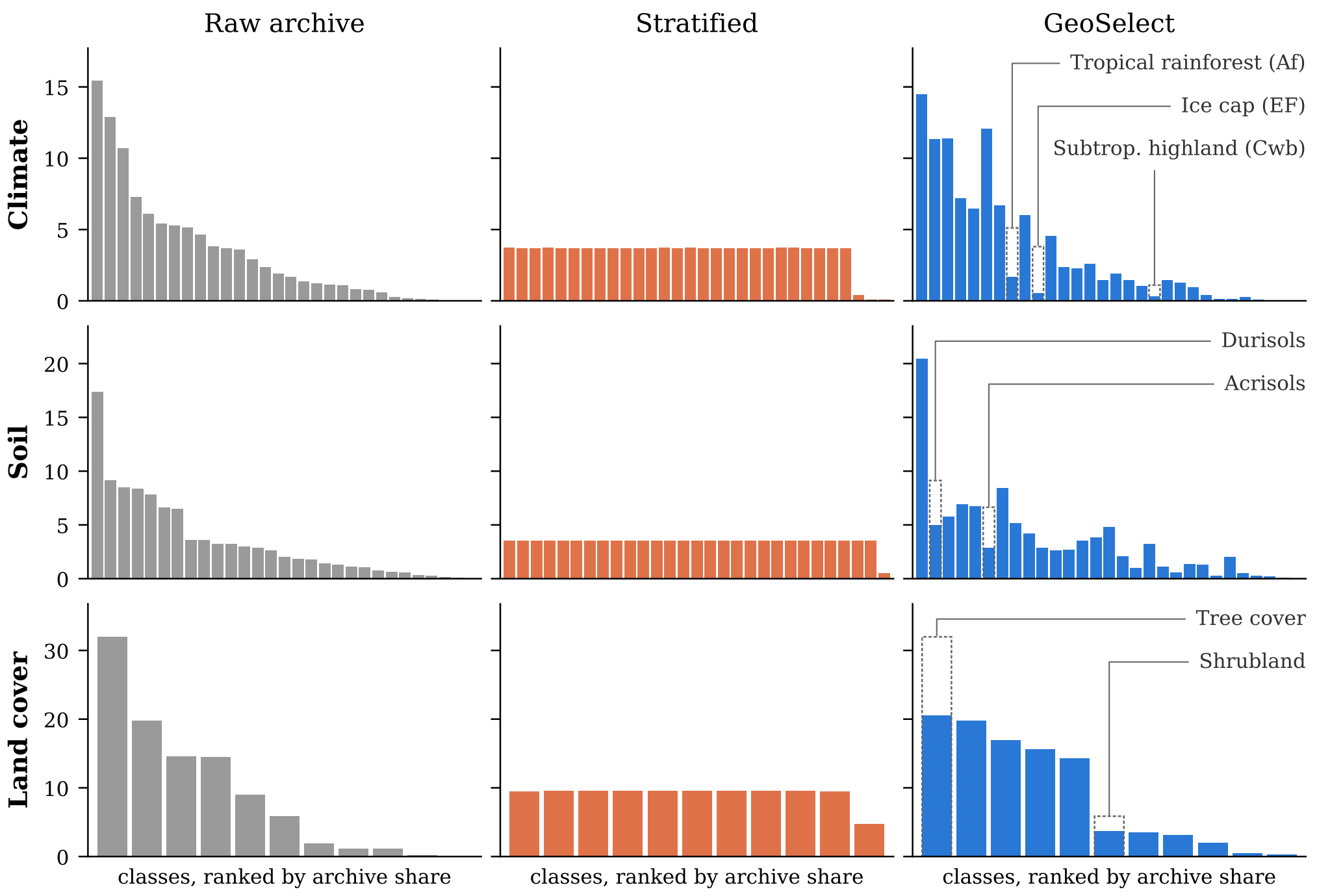}
\caption{\textbf{Metadata distributions after curation.}
Distribution of climate (K\"oppen--Geiger), soil (SoilGrids), and land-cover (ESA WorldCover) classes in the full archive, after stratification by the corresponding attribute, and after curation with \ours{} (both at a $1\%$ budget). Shares are computed over the observations with a class (the open ocean has none), and classes are ordered by decreasing frequency in the full archive. In the right column, dashed outlines show the archive share of the classes with the largest drops, which are named.}
\label{fig:keep_histograms}
\end{figure*}
\begin{table*}[t]
\centering
\caption{\textbf{Per-dataset results across SSL objectives.} GEO-Bench linear probing of DINOv2 and LeJEPA models pretrained on the full archive, on the $10\%$ subsets of \mainref{tab:ssl_transfer}{Table~3}, or on the $1\%$ subsets of \mainref{tab:geobench_curation}{Table~1}; \emph{Kept} is the fraction of the archive retained. \emph{Urban} retains the observations with the largest built-up fraction, and SDD is budget-controlled SemDeDup. The first row of each block reports the score of full-archive pretraining with that objective; all other entries are differences relative to it, shaded green for gains and red for losses. Classification uses accuracy except for BigEarthNet ($\mu$F1), segmentation uses mIoU, and \emph{Mean} averages the five tasks of each group. Bold marks the best curated subset in each column for each objective and budget.\ifnum\arxivversion=1\relax\ Means are computed over unrounded scores.\fi}
\label{tab:ssl_per_dataset}
\ifnum\arxivversion=1 \scriptsize
\setlength{\tabcolsep}{2pt}
\renewcommand{\arraystretch}{1.03}
\renewcommand{\applycolorwidth}{1.9em}
\resizebox{\textwidth}{!}{%
\begin{tabular}{l l l@{\;}c@{\;} @{\quad} c c c c c c @{\;\;\;\;} c c c c c c}
\toprule
& \multicolumn{3}{c}{Curation} & \multicolumn{6}{c}{Classification} & \multicolumn{6}{c}{Segmentation} \\
\cmidrule(lr){2-4}\cmidrule(lr){5-10}\cmidrule(lr){11-16}
SSL & \makecell{Criterion/\\features} & \makecell{Sampling \\ strategy} & Kept
& \makecell{So2\\Sat} & \makecell{Big\\Earth\\Net} & \makecell{Forest\\Net} & \makecell{Euro\\SAT} & \makecell{PV4\\GER} & \bf Mean
& \makecell{Cash\\ew} & \makecell{Chesa\\peake} & \makecell{SA\\Crop} & \makecell{Neon\\Tree} & \makecell{PV4\\GER} & \bf Mean \\
\midrule
\multirow{7}{*}{DINOv2} & \multicolumn{2}{l}{\emph{Full dataset}} & 100\% & 40.4 & 54.2 & 46.5 & 84.2 & 93.9 & 63.8 & 28.8 & 57.9 & 25.7 & 53.1 & 77.4 & 48.6 \\
\cmidrule(l){2-16}
 & Urban & top-\% & 10\%
& \applycolordeltaA{+7.3}
& \applycolordeltaB{+3.6}
& \applycolordeltaD{+4.5}
& \applycolordeltaE{+1.8}
& \applycolordeltaF{+0.3}
& \applycolordeltaG{+3.5}
& \applycolordeltaHb{+6.8}
& \applycolordeltaI{+3.0}
& \applycolordeltaJb{+2.2}
& \applycolordeltaK{+0.2}
& \applycolordeltaL{+3.8}
& \applycolordeltaM{+3.2} \\\arrayrulecolor{black!30}\cmidrule(l){2-16}\arrayrulecolor{black}
 & UniverSat & SDD & 10\%
& \applycolordeltaA{+5.6}
& \applycolordeltaB{+1.9}
& \applycolordeltaD{+0.5}
& \applycolordeltaE{+1.0}
& \applycolordeltaF{+0.3}
& \applycolordeltaG{+1.9}
& \applycolordeltaH{+3.8}
& \applycolordeltaI{+2.9}
& \applycolordeltaJ{+1.7}
& \applycolordeltaK{+0.4}
& \applycolordeltaL{+2.4}
& \applycolordeltaM{+2.2} \\
 & UniverSat & \bf \ours{} & 10\%
& \applycolordeltaAb{+7.6}
& \applycolordeltaBb{+4.0}
& \applycolordeltaDb{+4.7}
& \applycolordeltaEb{+2.3}
& \applycolordeltaFb{+0.5}
& \applycolordeltaGb{+3.8}
& \applycolordeltaH{+5.9}
& \applycolordeltaIb{+3.5}
& \applycolordeltaJb{+2.2}
& \applycolordeltaKb{+0.7}
& \applycolordeltaLb{+4.8}
& \applycolordeltaMb{+3.4} \\
\cmidrule(l){2-16}
 & Urban & top-\% & 1\%
& \applycolordeltaA{+5.8}
& \applycolordeltaBb{+3.8}
& \applycolordeltaD{+3.8}
& \applycolordeltaEb{+2.9}
& \applycolordeltaF{+0.5}
& \applycolordeltaGb{+3.4}
& \applycolordeltaHb{+7.6}
& \applycolordeltaIb{+3.6}
& \applycolordeltaJ{+1.8}
& \applycolordeltaK{+0.2}
& \applycolordeltaL{+3.9}
& \applycolordeltaMb{+3.4} \\\arrayrulecolor{black!30}\cmidrule(l){2-16}\arrayrulecolor{black}
 & UniverSat & SDD & 1\%
& \applycolordeltaA{+4.8}
& \applycolordeltaB{+1.7}
& \applycolordeltaD{+2.7}
& \applycolordeltaE{+0.4}
& \applycolordeltaF{+0.6}
& \applycolordeltaG{+2.1}
& \applycolordeltaH{+3.1}
& \applycolordeltaI{+2.6}
& \applycolordeltaJ{+0.5}
& \applycolordeltaK{+0.2}
& \applycolordeltaL{+2.3}
& \applycolordeltaM{+1.7} \\
 & UniverSat & \bf \ours{} & 1\%
& \applycolordeltaAb{+7.6}
& \applycolordeltaB{+2.8}
& \applycolordeltaDb{+4.7}
& \applycolordeltaE{+0.8}
& \applycolordeltaFb{+1.1}
& \applycolordeltaGb{+3.4}
& \applycolordeltaH{+5.6}
& \applycolordeltaI{+3.3}
& \applycolordeltaJb{+2.0}
& \applycolordeltaKb{+0.4}
& \applycolordeltaLb{+4.4}
& \applycolordeltaM{+3.1} \\
\midrule
\multirow{7}{*}{LeJEPA} & \multicolumn{2}{l}{\emph{Full dataset}} & 100\% & 37.5 & 53.4 & 48.4 & 83.4 & 93.6 & 63.3 & 27.2 & 59.7 & 25.1 & 53.1 & 76.9 & 48.4 \\
\cmidrule(l){2-16}
 & Urban & top-\% & 10\%
& \applycolordeltaA{+8.2}
& \applycolordeltaB{+2.5}
& \applycolordeltaD{+0.2}
& \applycolordeltaE{+2.1}
& \applycolordeltaF{-0.8}
& \applycolordeltaG{+2.4}
& \applycolordeltaH{+3.7}
& \applycolordeltaI{+1.0}
& \applycolordeltaJ{+1.5}
& \applycolordeltaKb{+0.3}
& \applycolordeltaL{+2.6}
& \applycolordeltaM{+1.8} \\\arrayrulecolor{black!30}\cmidrule(l){2-16}\arrayrulecolor{black}
 & UniverSat & SDD & 10\%
& \applycolordeltaA{+2.5}
& \applycolordeltaB{+1.4}
& \applycolordeltaD{+0.5}
& \applycolordeltaE{+2.0}
& \applycolordeltaF{-0.3}
& \applycolordeltaG{+1.2}
& \applycolordeltaH{+2.7}
& \applycolordeltaI{+0.2}
& \applycolordeltaJ{+1.4}
& \applycolordeltaK{+0.2}
& \applycolordeltaL{+2.4}
& \applycolordeltaM{+1.4} \\
 & UniverSat & \bf \ours{} & 10\%
& \applycolordeltaAb{+8.8}
& \applycolordeltaBb{+4.2}
& \applycolordeltaDb{+1.4}
& \applycolordeltaEb{+3.6}
& \applycolordeltaFb{+1.9}
& \applycolordeltaGb{+4.0}
& \applycolordeltaHb{+4.0}
& \applycolordeltaIb{+1.7}
& \applycolordeltaJb{+2.8}
& \applycolordeltaK{-0.1}
& \applycolordeltaLb{+4.6}
& \applycolordeltaMb{+2.6} \\
\cmidrule(l){2-16}
 & Urban & top-\% & 1\%
& \applycolordeltaAb{+8.6}
& \applycolordeltaB{+3.0}
& \applycolordeltaD{+1.3}
& \applycolordeltaE{+2.1}
& \applycolordeltaF{-0.5}
& \applycolordeltaG{+2.9}
& \applycolordeltaHb{+4.4}
& \applycolordeltaI{+1.5}
& \applycolordeltaJ{+2.3}
& \applycolordeltaK{-0.1}
& \applycolordeltaL{+3.0}
& \applycolordeltaM{+2.2} \\\arrayrulecolor{black!30}\cmidrule(l){2-16}\arrayrulecolor{black}
 & UniverSat & SDD & 1\%
& \applycolordeltaA{+4.9}
& \applycolordeltaBb{+3.9}
& \applycolordeltaD{-0.3}
& \applycolordeltaE{+0.3}
& \applycolordeltaF{+0.3}
& \applycolordeltaG{+1.8}
& \applycolordeltaH{+1.6}
& \applycolordeltaI{+0.5}
& \applycolordeltaJ{+1.4}
& \applycolordeltaK{-0.1}
& \applycolordeltaL{+2.4}
& \applycolordeltaM{+1.2} \\
 & UniverSat & \bf \ours{} & 1\%
& \applycolordeltaA{+7.1}
& \applycolordeltaB{+3.2}
& \applycolordeltaDb{+1.5}
& \applycolordeltaEb{+4.0}
& \applycolordeltaFb{+1.2}
& \applycolordeltaGb{+3.4}
& \applycolordeltaH{+4.2}
& \applycolordeltaIb{+1.8}
& \applycolordeltaJb{+2.4}
& \applycolordeltaKb{0.0}
& \applycolordeltaLb{+4.5}
& \applycolordeltaMb{+2.6} \\
\bottomrule

\end{tabular}
}
\else\input{tables/ssl_per_dataset}\fi{}
\end{table*}
\begin{table*}[t]
\centering
\caption{\textbf{$k$-NN classification on GEO-Bench.} DINOv2 and LeJEPA models of \mainref{tab:ssl_transfer}{Table~3}, at a $10\%$ budget, presented as in \cref{tab:ssl_per_dataset}: full-archive scores, then differences shaded green for gains and red for losses, with bold for the best curated subset in each column. For each dataset, the image embedding (CLS token, mean patch token, or both), $k$, and distance are selected on the validation set; we report test accuracy, or $\mu$F1 for BigEarthNet.\ifnum\arxivversion=1\relax\ Means are computed over unrounded scores.\fi}
\label{tab:knn}
\ifnum\arxivversion=1 {
\small
\setlength{\tabcolsep}{3pt}
\renewcommand{\applycolorwidth}{2.1em}
\begin{tabular}{l l l c @{\quad} c c c c c c}
\toprule
& \multicolumn{3}{c}{Curation} & \multicolumn{6}{c}{Classification} \\
\cmidrule(lr){2-4}\cmidrule(lr){5-10}
SSL & \makecell{Criterion/\\features} & \makecell{Sampling\\strategy} & Kept & \makecell{So2\\Sat} & \makecell{Big\\Earth\\Net} & \makecell{Forest\\Net} & \makecell{Euro\\SAT} & \makecell{PV4\\GER} & \bf Mean \\
\midrule
\multirow{4}{*}{DINOv2} & \multicolumn{2}{l}{\emph{Full dataset}} & 100\% & 33.4 & 51.3 & 36.4 & 75.2 & 90.5 & 57.3 \\
\cmidrule(l){2-10}
 & Urban & top-\% & 10\% & \applycolor{+4.2}{0}{5} & \applycolor{+4.6}{0}{5} & \applycolorbf{+4.5}{0}{5} & \applycolorbf{+3.1}{0}{5} & \applycolorbf{+2.0}{0}{5} & \applycolor{+3.7}{0}{3} \\\arrayrulecolor{black!30}\cmidrule(l){2-10}\arrayrulecolor{black}
 & UniverSat & SDD & 10\% & \applycolor{+2.5}{0}{5} & \applycolor{+4.2}{0}{5} & \applycolor{+3.3}{0}{5} & \applycolor{+2.5}{0}{5} & \applycolor{+1.6}{0}{5} & \applycolor{+2.9}{0}{3} \\
 & UniverSat & \bf \ours{} & 10\% & \applycolorbf{+5.7}{0}{5} & \applycolorbf{+5.3}{0}{5} & \applycolor{+2.7}{0}{5} & \applycolorbf{+3.1}{0}{5} & \applycolor{+1.8}{0}{5} & \applycolorbf{+3.8}{0}{3} \\
\midrule
\multirow{4}{*}{LeJEPA} & \multicolumn{2}{l}{\emph{Full dataset}} & 100\% & 32.7 & 48.0 & 34.0 & 72.7 & 90.5 & 55.6 \\
\cmidrule(l){2-10}
 & Urban & top-\% & 10\% & \applycolor{+2.7}{0}{5} & \applycolor{+5.2}{0}{5} & \applycolor{+3.6}{0}{5} & \applycolor{+2.9}{0}{5} & \applycolor{+0.5}{0}{5} & \applycolor{+3.0}{0}{3} \\\arrayrulecolor{black!30}\cmidrule(l){2-10}\arrayrulecolor{black}
 & UniverSat & SDD & 10\% & \applycolor{+3.1}{0}{5} & \applycolor{+4.4}{0}{5} & \applycolor{+3.7}{0}{5} & \applycolorbf{+4.0}{0}{5} & \applycolorbf{+0.9}{0}{5} & \applycolor{+3.2}{0}{3} \\
 & UniverSat & \bf \ours{} & 10\% & \applycolorbf{+3.8}{0}{5} & \applycolorbf{+7.5}{0}{5} & \applycolorbf{+8.5}{0}{5} & \applycolor{+2.6}{0}{5} & \applycolor{+0.2}{0}{5} & \applycolorbf{+4.5}{0}{3} \\
\bottomrule
\end{tabular}
}
\else\input{tables/knn}\fi{}
\end{table*}

\subsection{Local and global diversity}
\label{sec:supp:two_div}
\Cref{tab:two_div} characterizes the subsets obtained by promoting local and global diversity separately or jointly (\mainref{sec:exp:two_div}{Sec.~4.3}).
SemDeDup achieves a fivefold increase in nearest-neighbor distance over random sampling, but a quarter of its observations fall in the less locally diverse half of the archive, which contains many observations dominated by bare ground, shrubland, or water.
SemDeDup and \ours{} share less than a third of their observations: SemDeDup retains distinctive observations that can be internally uniform, whereas \ours{} retains more locally diverse ones.
Selecting solely by $L_i$, regardless of clusters, has the opposite effect: the retained observations are highly diverse locally but nearly as redundant as those of a random subset. They also span fewer geographic regions, concentrating at higher latitudes and along coastlines. Ranking by $L_i$ within clusters (\emph{Local ranking}) recovers most of this coverage, but the retained observations remain more redundant than those of \ours{}: at nearly identical local diversity and geographic coverage, combining $L_i$ with the global diversity $G_i$ raises the nearest-neighbor distance by half, from $10.1$ to $15.1$, by discarding near-duplicates such as those of \cref{fig:qual_dups}.
By combining both criteria, \ours{} retains highly diverse observations while preserving over $90\%$ of SemDeDup's geographic coverage and $60\%$ of its nearest-neighbor distance.
\ifnum\arxivversion=0\relax
\begin{table*}[t]
\centering
\begin{minipage}[c]{0.51\textwidth}
\caption{\textbf{Cross-modality curation at $10\%$ and $1\%$.} Sentinel-1 MAE pretraining on the full archive or on subsets curated by \ours{} from Sentinel-1 or co-registered Sentinel-2 embeddings (\emph{Curated from}); \emph{Kept} is the fraction of the archive retained. reBEN-S1 is evaluated with a $k$-NN classifier on pooled patch tokens (\cref{sec:supp:knn}) and with a linear probe, Sen1Floods11 with a linear probe; linear-probing results at $10\%$ are those of \mainref{tab:modality_transfer}{Table~4}. Small numbers are differences with the full archive, computed before rounding.}
\label{tab:s1_1pct}
\end{minipage}
\hfill
\begin{minipage}[c]{0.46\textwidth}
\centering
\input{tables/modality_transfer_1pct}
\end{minipage}
\end{table*}
\fi

\subsection{Metadata distributions}
\label{sec:supp:correlation}
\Cref{fig:keep_histograms} complements the association analysis of \mainref{tab:keep_correlation}{Table~2} by showing how curation alters individual metadata distributions. While stratification explicitly balances classes, \ours{} roughly follows the distributions of the archive, apart from a few classes that it reduces markedly: the ice-cap (EF), tropical-rainforest (Af), and subtropical-highland (Cwb) climates, Durisols and Acrisols, and tree cover and shrubland. Tree cover, the most frequent land-cover class, drops from $32\%$ to $20\%$ of the labeled observations; \cref{fig:qual_climates} shows retained and discarded observations of the first two climates. This suggests that the benefits of semantic curation stem primarily from selection within metadata classes rather than from rebalancing them.

\section{Additional results}
\label{sec:supp:results}
\subsection{Per-dataset results across SSL objectives}
\label{sec:supp:ssl_per_dataset}
\Cref{tab:ssl_per_dataset} details, for each GEO-Bench dataset, the linear-probing results of the DINOv2 and LeJEPA models of \mainref{tab:ssl_transfer}{Table~3}. The gains of \ours{} are broad rather than driven by a few datasets: with both objectives, it improves over the full archive and over SemDeDup on every dataset except NeonTree with LeJEPA. They are largest on So2Sat, by about $8$ points, and smallest where there is little room for improvement: on NeonTree, which no curated subset changes by more than a point, and, with DINOv2, on PV4GER classification, where the full-archive model already exceeds $93\%$ accuracy.

\Cref{tab:ssl_per_dataset} also reports both objectives at a $1\%$ budget, on the subsets of \mainref{tab:geobench_curation}{Table~1}. \ours{} keeps most of its gains at this budget, slightly below those at $10\%$, and improves or matches the full archive on every dataset: as with MAE, a curated subset $100\times$ smaller than the archive yields better representations with both view-based objectives. \ours{} also remains ahead of SemDeDup with both objectives, by about $1.5$ points in both classification and segmentation. Urban prioritization is also competitive at this budget: with DINOv2, it matches \ours{} in classification and slightly exceeds it in segmentation ($+3.4$ versus $+3.1$ mIoU), whereas with LeJEPA it falls behind \ours{} in both, as at $10\%$.

\subsection{\texorpdfstring{$k$}{k}-NN evaluation}
\label{sec:supp:knn}
Linear probing involves training a parametric classifier and tuning its optimization hyperparameters. As a more direct evaluation of the learned representations, we classify frozen image embeddings using a $k$-nearest-neighbor classifier (\cref{tab:knn}). \ifnum\arxivversion=1 We consider the DINOv2 and LeJEPA models of \mainref{tab:ssl_transfer}{Table~3}, whose objectives explicitly supervise global image embeddings, as well as the MAE models of \cref{tab:modality_transfer,tab:s2naip_summary}, whose objective operates on patch tokens. The image embedding (CLS token, mean patch token, or their concatenation for DINOv2 and LeJEPA; mean- or max-pooled patch tokens for MAE), the number\else We consider the DINOv2 and LeJEPA models of \mainref{tab:ssl_transfer}{Table~3}, whose objectives explicitly supervise global image embeddings, unlike MAE, which operates on patch tokens. The image embedding (CLS token, mean patch token, or their concatenation), the number\fi{} of neighbors $k\in\{1,3,5,10,20\}$, and the distance are selected on each dataset's validation set. Distances are Euclidean, computed on the raw embeddings or after $\ell_2$ normalization (\ie, cosine), a concatenation being normalized as a whole. Neighbors vote with equal weights; for the multilabel BigEarthNet and reBEN-S1, each label is predicted independently, when a strict majority of the $k$ neighbors carries it, without a tuned threshold.

The $k$-NN results largely agree with linear probing: all curated subsets outperform the full archive with both SSL objectives. With DINOv2, \ours{} and urban prioritization perform equally well, ahead of SemDeDup. With LeJEPA, \ours{} performs best, notably on ForestNet and BigEarthNet, while SemDeDup overtakes urban prioritization. The benefits of curation are thus directly reflected in the geometry of the learned embeddings, without requiring a trained probe.

\ifnum\arxivversion=0\relax
\subsection{Cross-modality curation at a \texorpdfstring{$1\%$}{1\%} budget}
\label{sec:supp:s1_1pct}
\Cref{tab:s1_1pct} extends the cross-modality experiment of \mainref{sec:exp:modality_transfer}{Sec.~4.3} to a $1\%$ budget and includes a $k$-NN evaluation using pooled patch tokens. On reBEN-S1, \ours{} outperforms full-archive pretraining in every setting but one: under linear probing with both embedding sources and at both budgets, and under $k$-NN except when curating from Sentinel-2 at $1\%$, which falls slightly below the baseline. Flood segmentation on Sen1Floods11 remains largely unaffected, within $0.2$ mIoU of the baseline. The two embedding sources differ at the smaller budget: curation from Sentinel-1 embeddings further improves upon its $10\%$ result under linear probing, reaching $+1.6$ $\mu$F1, while curation from Sentinel-2 retains less than half of its initial gain, and $k$-NN favors Sentinel-1 curation at both budgets. Optical embeddings thus transfer to radar curation, but they are at least as effective as radar embeddings (\mainref{sec:exp:modality_transfer}{Sec.~4.3}) only under linear probing at $10\%$; embeddings from the target modality yield more consistent gains.
\fi

\begin{table*}[t]
\centering
\caption{\textbf{Curation of S2-NAIP.} GEO-Bench linear probing of MAE models pretrained on the full S2-NAIP archive or on $1\%$ subsets of it, presented as in \mainref{tab:geobench_curation}{Table~1}: the first row reports full-archive scores, and all other values are differences relative to it, shaded green for gains and red for losses. \emph{Random} draws observations uniformly from the archive, and \emph{Urban} retains the observations whose tile has the largest built-up fraction. Bold marks the best curated subset in each column.\ifnum\arxivversion=1\relax\ Means are computed over unrounded scores.\fi}
\label{tab:s2naip}
\ifnum\arxivversion=1 \scriptsize
\setlength{\tabcolsep}{2pt}
\renewcommand{\arraystretch}{1.03}
\renewcommand{\applycolorwidth}{1.9em}
\resizebox{\textwidth}{!}{%
\begin{tabular}{l l@{\;}c@{\;} @{\quad} c c c c c c @{\;\;\;\;} c c c c c c}
\toprule
\multicolumn{3}{c}{Curation} & \multicolumn{6}{c}{Classification} & \multicolumn{6}{c}{Segmentation} \\
\cmidrule(lr){1-3}\cmidrule(lr){4-9}\cmidrule(lr){10-15}
\makecell{Criterion/\\features} & \makecell{Sampling \\ strategy} & Kept
& \makecell{So2\\Sat} & \makecell{Big\\Earth\\Net} & \makecell{Forest\\Net} & \makecell{Euro\\SAT} & \makecell{PV4\\GER} & \bf Mean
& \makecell{Cash\\ew} & \makecell{Chesa\\peake} & \makecell{SA\\Crop} & \makecell{Neon\\Tree} & \makecell{PV4\\GER} & \bf Mean \\
\midrule
\multicolumn{2}{l}{\emph{Full dataset}} & 100\% & 46.0 & 61.4 & 51.2 & 92.1 & 94.4 & 69.0 & 30.1 & 63.6 & 29.4 & 53.0 & 79.5 & 51.1 \\
\midrule
\multicolumn{2}{l}{Random} & 1\%
& \applycolordeltaA{-0.6}
& \applycolordeltaB{-1.2}
& \applycolordeltaD{-0.1}
& \applycolordeltaE{+0.6}
& \applycolordeltaF{+0.2}
& \applycolordeltaG{-0.2}
& \applycolordeltaH{-0.5}
& \applycolordeltaI{0.0}
& \applycolordeltaJ{-0.7}
& \applycolordeltaK{-0.5}
& \applycolordeltaL{-1.9}
& \applycolordeltaM{-0.7} \\
Urban & top-\% & 1\%
& \applycolordeltaAb{+2.9}
& \applycolordeltaB{-1.3}
& \applycolordeltaD{+0.3}
& \applycolordeltaE{-0.9}
& \applycolordeltaF{-0.5}
& \applycolordeltaG{+0.1}
& \applycolordeltaH{-1.2}
& \applycolordeltaI{-8.2}
& \applycolordeltaJ{-1.2}
& \applycolordeltaK{-0.7}
& \applycolordeltaL{-1.5}
& \applycolordeltaM{-2.6} \\
UniverSat & SDD & 1\%
& \applycolordeltaA{+0.6}
& \applycolordeltaB{-1.1}
& \applycolordeltaDb{+0.5}
& \applycolordeltaEb{+1.3}
& \applycolordeltaFb{+0.7}
& \applycolordeltaGb{+0.4}
& \applycolordeltaH{-0.1}
& \applycolordeltaI{-1.7}
& \applycolordeltaJ{-2.0}
& \applycolordeltaK{-0.9}
& \applycolordeltaL{-0.1}
& \applycolordeltaM{-0.9} \\
UniverSat & \bf \ours{} & 1\%
& \applycolordeltaA{-0.9}
& \applycolordeltaBb{-0.4}
& \applycolordeltaD{+0.3}
& \applycolordeltaE{+0.4}
& \applycolordeltaF{+0.3}
& \applycolordeltaG{-0.1}
& \applycolordeltaHb{+0.3}
& \applycolordeltaIb{+0.2}
& \applycolordeltaJb{-0.4}
& \applycolordeltaKb{-0.4}
& \applycolordeltaLb{+1.2}
& \applycolordeltaMb{+0.2} \\
\bottomrule
\end{tabular}
}
\else\input{tables/s2naip}\fi{}
\end{table*}

\section{Curating a second archive: S2-NAIP}
\label{sec:supp:s2naip}
We apply \ours{} to a second archive whose structure differs from Major-TOM: S2-NAIP\suppcite{wolters2023zooming}, which observes small tiles of the contiguous United States repeatedly over a few months. Restricted to a single country, it is geographically far narrower than Major-TOM, but it contains over a hundred times more observations, so that a $1\%$ subset is larger than the whole of Major-TOM. It thus mainly tests whether \ours{} can filter archives of hundreds of millions of observations. \Cref{tab:s2naip} reports the results.

\paragraph{Archive.}
S2-NAIP covers $19.3$M tiles of $640\times640$\,m. For each tile, it provides $16$ to $32$ Sentinel-2 Level-1C images of $64\times64$ pixels at $10$\,m, acquired within a few months, along with other modalities that we do not use. We treat each Sentinel-2 image as an independent observation, which gives $513$M observations. A quarter of them duplicate a satellite pass already present for the same tile, most often as the same acquisition processed twice; we keep one observation per pass. Products generated with version 04.00 or later of ESA's processing baseline, \ie, since January 2022 or reprocessed, add an offset of $1{,}000$ to their digital numbers, which we subtract. We then keep the observations with no missing pixel and at most $5\%$ cloud cover, which leaves $302$M observations. All observations of one tile in a thousand are held out for validation, and curation selects from the others.

\begin{table}[t]
\centering
\caption{\textbf{$k$-NN classification on S2-NAIP.} Models of \cref{tab:s2naip}, presented as in that table. The image embedding, the mean- or max-pooled patch tokens, is selected together with $k$ and the distance on each dataset's validation set (\cref{sec:supp:knn}).\ifnum\arxivversion=1\relax\ Means are computed over unrounded scores.\fi}
\label{tab:s2naip_knn}
\ifnum\arxivversion=1 {\scriptsize
\setlength{\tabcolsep}{2pt}
\renewcommand{\applycolorwidth}{2.0em}
\resizebox{\linewidth}{!}{%
\begin{tabular}{l c @{\quad} c c c c c c}
\toprule
\multicolumn{2}{c}{Curation} & \multicolumn{6}{c}{Classification} \\
\cmidrule(lr){1-2}\cmidrule(lr){3-8}
\makecell{Sampling\\strategy} & Kept & \makecell{So2\\Sat} & \makecell{Big\\Earth\\Net} & \makecell{Forest\\Net} & \makecell{Euro\\SAT} & \makecell{PV4\\GER} & \bf Mean \\
\midrule
\emph{Full dataset} & 100\% & 37.4 & 59.0 & 35.9 & 85.8 & 91.6 & 61.9 \\
\midrule
Random & 1\% & \applycolordeltaA{-0.3} & \applycolordeltaB{+1.0} & \applycolordeltaD{+3.8} & \applycolordeltaE{-3.6} & \applycolordeltaF{0.0} & \applycolordeltaG{+0.2} \\
Urban & 1\% & \applycolordeltaAb{+1.4} & \applycolordeltaBb{+2.3} & \applycolordeltaD{+4.1} & \applycolordeltaE{-1.2} & \applycolordeltaF{-1.0} & \applycolordeltaG{+1.2} \\
SDD & 1\% & \applycolordeltaA{-0.6} & \applycolordeltaB{-0.4} & \applycolordeltaD{+3.3} & \applycolordeltaEb{+0.2} & \applycolordeltaFb{+0.1} & \applycolordeltaG{+0.6} \\
\bf \ours{} & 1\% & \applycolordeltaA{0.0} & \applycolordeltaB{+1.2} & \applycolordeltaDb{+6.1} & \applycolordeltaE{-0.2} & \applycolordeltaF{-0.3} & \applycolordeltaGb{+1.4} \\
\bottomrule
\end{tabular}
}}
\else\input{tables/s2naip_knn}\fi{}
\end{table}

\paragraph{Curation.}
We use the configuration of \cref{sec:supp:curation} with two adaptations: each observation is resampled to $32\times32$ pixels at $20$\,m and embedded with $160$\,m patches, \ie, $4\times4$ tokens, and $k$-means uses $96{,}367$ clusters. At $1\%$, SemDeDup and \ours{} share less than a third of their observations, as on Major-TOM. Although the archive contains up to $32$ observations of each tile, neither selection reduces it to one observation per place: both retain about two observations per selected tile, against about one for a random subset of the same size. Different dates of a tile are thus often distinct enough to be retained together: curation exploits the temporal diversity of this archive, which Major-TOM, observing each location once, does not offer. Both selections also shift the geographic distribution toward the central United States (UTM zones 14 and 15), which holds over half of either subset against $28\%$ of the archive. Curation remains inexpensive at this scale: extracting the global embeddings and patch-token statistics of the $302$M observations takes about $30$ H100 GPU-hours, and clustering and selection about six A100 GPU-hours.

\begin{table}[t]
\centering
\caption{\textbf{Curation configuration.} The pipeline of \mainref{sec:method}{Sec.~3}, run
once per archive. Clustering, the ordering by distance to centroid, $m_i$, and $G_i$ all
operate on the same global embeddings $f_i$; local diversity uses the patch tokens.}
\label{tab:hyperparams_curation}
\small{
\begin{tabular}{@{}l>{\raggedright\arraybackslash}p{0.58\linewidth}@{}}
\toprule
Stage / hyperparameter & Value \\
\midrule
\multicolumn{2}{l}{\emph{Embedding (\mainref{sec:method:diversity}{Sec.~3.1})}} \\
Scorer                & UniverSat, frozen \\
Bands                 & 10 S2 (B02--B08, B8A, B11, B12) \\
Input                 & whole observation, resampled to $534\times534$ at $20$\,m \\
Patch size            & $240$\,m, $44\times44$ tokens \\
Global embedding $f_i$ & mean over tokens, $D=768$, $\ell_2$-normalized \\
Local diversity   & per-feature token std $\sigma_{i,d}$, $\epsilon_d$ = median of $\sigma_{\cdot,d}$ \\\greyrule
\multicolumn{2}{l}{\emph{Clustering (\mainref{sec:method:diversity}{Sec.~3.1})}} \\
Algorithm             & $k$-means, fp16 assign / fp32 similarity \\
Clusters              & $5{,}786$ \\
Iterations            & 20 \\\greyrule
\multicolumn{2}{l}{\emph{Per-cluster allocation}} \\
Priority order        & decreasing distance to centroid \\
Similarity            & cosine on $f_i$ (\mainref{sec:method:prune}{Sec.~3.2}) \\
Budget $\to$ threshold & quantile of $m_i$, giving $B_c$ \\\greyrule
\multicolumn{2}{l}{\emph{Selection within clusters}} \\
Global diversity  & $G_i$, cosine on $f_i$ (\mainref{eq:global}{Eq.~(5)}) \\
Score             & $S_i$, equal weights (\mainref{eq:diversity}{Eq.~(6)}) \\
Standardization   & median, IQR$/1.349$ (same ranking as \mainref{eq:diversity}{Eq.~(6)}) \\
Seed                  & 42 \\
\bottomrule
\end{tabular}
}
\end{table}

\begin{table}[t]
\centering
\caption{\textbf{Linear-probing hyperparameters.} The same grid is used for every pretrained model; the configuration with the best validation score is evaluated on the test set. $K$ is the number of classes and $P$ the patch size in pixels.}
\label{tab:hyperparams_lp}
\footnotesize{
\setlength{\tabcolsep}{4pt}
\begin{tabular}{@{}l>{\raggedright\arraybackslash}p{0.34\linewidth}>{\raggedright\arraybackslash}p{0.34\linewidth}@{}}
\toprule
Hyperparameter & Classification & Segmentation \\
\midrule
Input features        & pooled patch tokens & patch tokens \\
Pooling               & \{mean, max\} & -- \\
Output                & $K$ logits & $P\times P\times K$ logits per patch \\
Optimizer             & AdamW & AdamW \\
Schedule              & cosine, no warmup & cosine, no warmup \\
Epochs                & 50 & 50 \\
Learning rate         & \multicolumn{2}{l}{$10$ values from $10^{-4}$ to $10^{-1}$ ($1$--$2$--$5$ grid)} \\
Weight decay          & $\{0, 10^{-4}, 10^{-3},$ $10^{-2}, 10^{-1}\}$ & $\{0, 10^{-2}, 10^{-1}\}$ \\
Selection             & \multicolumn{2}{l}{best validation score over grid and epochs} \\
Metric                & accuracy ($\mu$F1 for BigEarthNet) & mIoU \\
Seed                  & 32 & 32 \\
\bottomrule
\end{tabular}
}
\end{table}

\paragraph{Pretraining.}
We pretrain MAE with the hyperparameters of \cref{sec:supp:pretraining}, adapted to the smaller observations: the input is the whole $64\times64$ tile with $8\times8$ patches ($64$ tokens) and the ten bands used for curation, the $20$\,m bands being upsampled to $10$\,m; the global batch size is $4{,}096$ and the learning rate $5\times10^{-4}$. Every run takes the same $180$k steps, \ie, $737$M observations: $2.4$ passes over the full archive and about $240$ over a $1\%$ subset. GEO-Bench evaluation follows \mainref{sec:exp}{Table~1}, with inputs restricted to the same ten bands.

\paragraph{Results.}
On S2-NAIP, retaining just $1\%$ of the archive largely preserves downstream performance, even with random sampling. This is consistent with its substantial spatial and temporal redundancy, with small tiles and multiple acquisitions per location, and with few large homogeneous regions, unlike Major-TOM, leaving limited room for performance gains from curation when evaluating on monotemporal tasks. Nevertheless, \ours{} is the only method to preserve segmentation performance ($+0.2$ mIoU), while remaining comparable to full-data pretraining in classification ($-0.1$ points). Under $k$-NN evaluation (\cref{tab:s2naip_knn}), \ours{} performs best ($+1.4$ points), suggesting better class-consistent neighborhoods despite similar linear-probing performance. Overall, \ours{} reduces the retained training data by $100\times$ without meaningful loss of performance.

\section{Implementation details}
\label{sec:supp:setup}
\begin{table*}[t]
\centering
\caption{\textbf{Pretraining hyperparameters.} MAE is used in all experiments, except those with DINOv2 and LeJEPA (Tables~\mainrefnum{tab:ssl_transfer}{3}, \ref{tab:ssl_per_dataset}, and~\ref{tab:knn}). Within each objective, the hyperparameters are identical for all subsets, whatever their size; entries spanning several columns are shared by those objectives, and -- marks components an objective does not use.}
\label{tab:hyperparams}
\footnotesize
\setlength{\tabcolsep}{4pt}
\newcolumntype{Q}[1]{>{\centering\arraybackslash}p{#1}}
\newlength{\hpcolA}\setlength{\hpcolA}{0.2\textwidth}
\newlength{\hpcolB}\setlength{\hpcolB}{0.305\textwidth}
\newcommand{\hpspanBC}[1]{\multicolumn{2}{Q{\dimexpr 2\hpcolB+2\tabcolsep\relax}}{#1}}
\newcommand{\hpspanABC}[1]{\multicolumn{3}{Q{\dimexpr \hpcolA+2\hpcolB+4\tabcolsep\relax}}{#1}}
\begin{tabular}{@{}lQ{\hpcolA}Q{\hpcolB}Q{\hpcolB}@{}}
\toprule
& MAE & DINOv2 & LeJEPA \\
\midrule
\multicolumn{4}{@{}l}{\emph{Encoder}} \\
Architecture & \hpspanABC{ViT-Base: patch size $16$, embedding dimension $768$, depth $12$, $12$ heads, MLP ratio $4$, $4$ registers, drop-path $0.1$} \\
Image descriptor & -- & \hpspanBC{register $0$ (as CLS token)} \\\greyrule
\multicolumn{4}{@{}l}{\emph{Input and views}} \\
Channels & $12$ (S2) or $3$ (S1) & \hpspanBC{$12$ (S2)} \\
Views & one $256\times256$ crop & \hpspanBC{$2$ global crops of $224\times224$ px, area $[0.32,1.0]$; $8$ local crops of $96\times96$ px, area $[0.05,0.32]$; blur $p=0.5$, $\sigma\in[0.1,2.0]$} \\
Spectral band drop & $0.1$ (S2 only) & $0.25$, student views only & $0.25$, every view \\\greyrule
\multicolumn{4}{@{}l}{\emph{Objective}} \\
Masking & spatial, ratio $0.8$ & iBOT: probability $0.5$, ratio $[0.1,0.5]$, supervised fraction $0.25$ & -- \\
Heads & decoder, depth $6$ & DINO and iBOT (shared): $3$ layers, $2{,}048$ hidden, $256$ bottleneck, $65{,}536$ prototypes; last layer frozen for $1$ epoch & linear head $768\to512$; projector: $3$ layers, $2{,}048$ hidden, $128$ out \\
Loss & masked-patch reconstruction & DINO $1.0$, iBOT $1.0$, KoLeo $0.1$ & invariance $0.95$, SIGReg $0.05$ \\
Loss details & -- & student temperature $0.1$; teacher temperature $0.04\!\to\!0.07$ over $10$ epochs; center momentum $0.9$ & invariance to the mean of the global-view embeddings; SIGReg: $256$ random slices, Epps--Pulley, $17$ knots on $[0,3]$ \\
Teacher & -- & EMA, momentum $0.992\!\to\!1.0$ (cosine) & -- \\\greyrule
\multicolumn{4}{@{}l}{\emph{Optimization}} \\
Optimizer & \hpspanABC{AdamW, $\epsilon=10^{-5}$, weight decay $0.05$} \\
Learning rate & $2\times10^{-4}$ & $2\times10^{-4}$ & $5\times10^{-4}$ \\
Schedule & ReduceLROnPlateau on validation loss (factor $0.1$, patience $10$) & \hpspanBC{$10$ warmup epochs, then cosine decay to $1\%$} \\
Epochs & $90$ ($180$k steps) & \hpspanBC{$100$ ($200$k steps)} \\
Global batch size & $256$ & \hpspanBC{$128$} \\
Precision, seed & \hpspanABC{bfloat16, $32$} \\
\bottomrule
\end{tabular}

\end{table*}

\subsection{The Major-TOM archive}
\label{sec:supp:majortom}
We use Major-TOM Core-S2L2A~\citep{francis2024majortom}, a global archive of Sentinel-2 Level-2A (surface reflectance) products cut along the Major-TOM grid, whose cells measure approximately $10\times10$\,km. Each observation is a $1{,}068\times1{,}068$ pixel image at $10$\,m resolution with $12$ spectral bands (B01--B12 and B8A, without the cirrus band B10). Only B02, B03, B04, and B08 are acquired at $10$\,m; the other bands, acquired at $20$ or $60$\,m, are resampled to $10$\,m.

Major-TOM stores a single product per grid cell, selected for low cloud cover. Apart from $7{,}197$ entries that duplicate the same product and which we remove, the archive therefore contains $2{,}238{,}687$ observations, each of a different location: there are no repeated observations of the same place. The archive covers the Earth's landmasses and coastlines, and $32.6\%$ of the observations contain no land pixel according to ESA WorldCover~\citep{zanaga2022worldcover}, \ie, depict open water.

We split the archive into training ($98\%$), validation ($1\%$), and test ($1\%$) sets with a deterministic hash of each observation. Curation only selects from the training set, so the validation set used for learning-rate scheduling is identical for all runs.

\subsection{Curation}
\label{sec:supp:curation}
\Cref{tab:hyperparams_curation} reports the configuration of \ours{} (\mainref{sec:method}{Sec.~3}). Curation is run once per archive, independently of the models later pretrained on the selected subsets. The patch tokens are extracted at a $20$\,m input resolution, which is about five times faster than at $10$\,m and barely changes the result: on a set of land observations, the mean embeddings and the diversity scores computed at both resolutions are nearly identical (cosine similarity and Spearman correlation of $0.99$).

\subsection{Linear probing}
\label{sec:supp:probing}
\Cref{tab:hyperparams_lp} reports the hyperparameters of the linear probes (\mainref{sec:exp}{Sec.~4.1}). For classification, the probe is applied to the mean- or max-pooled patch tokens; for segmentation, it is applied to each patch token and predicts the classes of the $P\times P$ pixels of the patch.

\subsection{Pretraining}
\label{sec:supp:pretraining}
\Cref{tab:hyperparams} reports the pretraining hyperparameters. MAE is the objective of all experiments except those of Tables~\mainrefnum{tab:ssl_transfer}{3}, \ref{tab:ssl_per_dataset}, and~\ref{tab:knn}, which use DINOv2 and LeJEPA. All three objectives train the same ViT-Base encoder, and an epoch is a fixed number of $2{,}000$ optimizer steps, whatever the size of the subset; smaller subsets are therefore revisited more often.

\end{document}